\documentclass[journal,onecolumn]{IEEEtran}

\usepackage{lineno,hyperref}
\usepackage[cmex10]{amsmath}
\usepackage{times}
\usepackage{epsfig}
\usepackage{graphicx}
\usepackage{amssymb}
\usepackage{algorithmic}
\usepackage{algorithm}
\usepackage{color}
\usepackage{subfig}

\usepackage{tikz}
\usepackage{mathrsfs}
\usepackage{makecell}
\usepackage{bm}
\usepackage{multicol}
\usepackage{multirow}
\usepackage[section]{placeins}
\usepackage[section]{placeins}
\usepackage{url}
\usepackage{array}
\definecolor{fgreen}{RGB}{34,139,34}

\def\eg{\emph{e.g.}}
\def\ie{\emph{i.e.}}
\def\etal{\emph{et al. }}

\definecolor{myblue}{RGB}{20,50,200}
\definecolor{mygreen}{RGB}{34,139,34}
\modulolinenumbers[5]

\begin{document}

\title{Multiple Instance Hybrid Estimator for Hyperspectral Target Characterization and Sub-pixel Target Detection}

\author{ Changzhe Jiao, Chao Chen, Ronald G. McGarvey, Stephanie Bohlman, Licheng Jiao, and Alina Zare \thanks{This material is based upon work supported by the National Science Foundation under Grant No. IIS-1350078 - CAREER: Supervised Learning for Incomplete and Uncertain Data.} \thanks{Changzhe Jiao and LiCheng Jiao are with the Key Laboratory of Intelligent Perception and Image Understanding, Ministry of Education of China, School of Artificial Intelligence, Xidian University, Xi'an 710071, China; Chao Chen is with the Mathworks, Natick MA 01760, USA; Stephanie Bohlman is with the School of Forest Resources and Conservation, University of Florida, Gainesville, FL 32611 USA. Alina Zare is with the Department of Electrical and Computer Engineering, University of Florida, Gainesville, FL 32611 USA (correspondence e-mail: azare@ufl.edu, cjiao@mail.missouri.edu).   }
}

\maketitle
\begin{abstract}
	The Multiple Instance Hybrid Estimator for discriminative target characterization from imprecisely labeled hyperspectral data is presented. In many hyperspectral target detection problems, acquiring accurately labeled training data is difficult. Furthermore, each pixel containing target is likely to be a mixture of both target and non-target signatures (\ie, sub-pixel targets), making extracting a pure prototype signature for the target class from the data extremely difficult. The proposed approach addresses these problems by introducing a data mixing model and optimizing the response of the hybrid sub-pixel detector within a multiple instance learning framework. The proposed approach iterates between estimating a set of discriminative target and non-target signatures and solving a sparse unmixing problem.  After learning target signatures, a signature based detector can then be applied on test data. Both simulated and real hyperspectral target detection experiments show the proposed algorithm is effective at learning discriminative target signatures and achieves superior performance over state-of-the-art comparison algorithms. 
\end{abstract}


\section{Introduction}
\label{sec:intro}

Hyperspectral imaging spectrometers collect electromagnetic energy scattered in the scene across hundreds or thousands of spectral bands, capturing both the spatial and spectral information \cite{landgrebe2002hyperspectral}. The spectral information is a combination of the reflection and/or emission of sunlight across wavelength by objects on the ground, and contains the unique spectral characteristics of different materials \cite {keshava2002spectral, bioucas2012hyperspectral}. The wealth of spectral information in hyperspectral imagery enables the possibility to conduct sub-pixel analysis in application areas including target detection \cite{yuksel2015multiple, 4389068}, precision agriculture \cite{mahajan2014using, wang2012mixture}, biomedical applications \cite{pike2016minimum, pardo2017directional} and others \cite{eismann2009automated, lara2013monitoring}.

Hyperspectral target detection generally refers to the task of locating all instances of a target given a known spectral signature within a hyperspectral scene. The reasons many classification methods are not applicable to hyperspectral target detection tasks are threefold:

1. Precise training labels for sub-pixel targets are often difficult or infeasible to obtain. For example, the ground truth information is often obtained using a Global Positioning System (GPS) receiver placed at the target location. However, the accuracy of the GPS coordinates could drift for several meters depending on the accuracy of the GPS system. 


2. The number of training instances from the positive (target) class is often small compared to that of the negative training data such that training an effective classifier is difficult. A hyperspectral image with hundred of thousands of pixels may only have a few pixels or sub-pixel level target points.

3. Due to the relatively low spatial resolution of hyperspectral imagery and the diversity of natural scenes, many targets are mixed points (sub-pixel targets) and the amount of the target proportions are not known. 

\begin{figure}[htb]
	\centering
	\includegraphics[width=12cm]{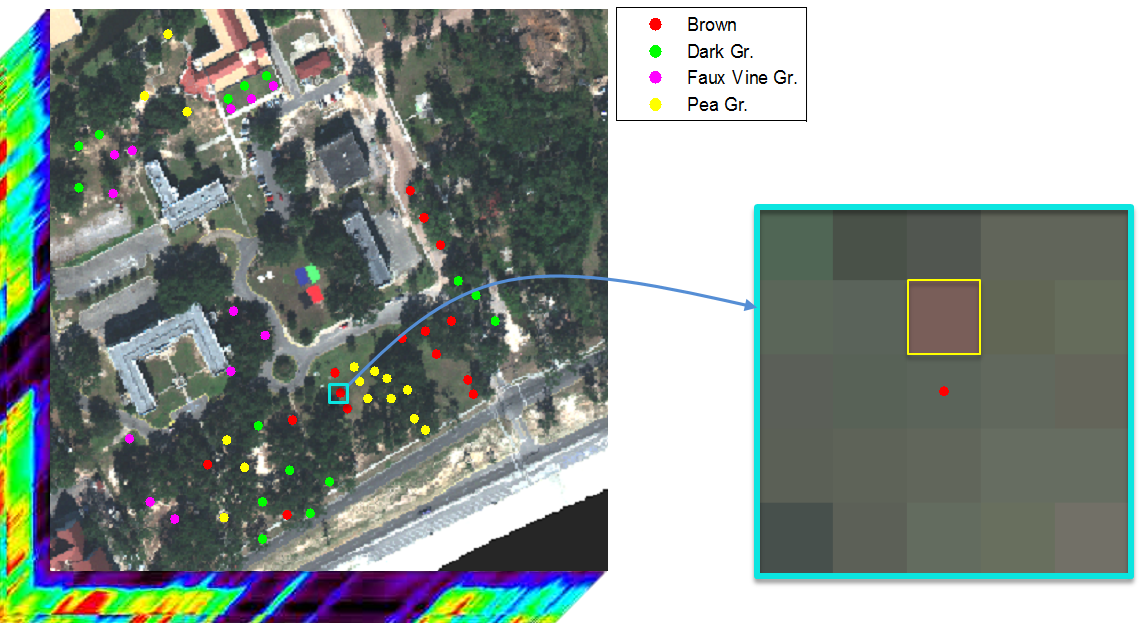}
	\caption{Illustration of inaccurate coordinates from GPS: one target denoted as brown by GPS has one pixel drift.}\label{fig:gulfport_tar_locs}
\end{figure}

For the above reasons, signature based hyperspectral target detection \cite{manolakis2002detection, nasrabadi2014hyperspectral} is generally applied over classification approaches. However, the performance of signature based detectors relies heavily on the quality of the target spectral signature used. Yet, obtaining an effective target signature can be a challenging problem. Sub-pixel targets are difficult to see in the imagery and the co-registration of the targets to GPS coordinates could drift across several pixels. 


As an example, the left image of Fig. \ref{fig:gulfport_tar_locs} shows the scattered target locations over the MUUFL Gulfport data set collected over the University of Southern Mississippi-Gulfpark Campus \cite{gader:2013}, where there are 4 types of targets throughout the scene: Brown (15 examples), Dark Green (15 examples), Faux Vineyard Green (12 examples) and Pea Green (15 examples). {The region highlighted by blue rectangle shown in the left image of Fig. \ref{fig:gulfport_tar_locs} contains one brown target. The image in the right of Fig. \ref{fig:gulfport_tar_locs} shows a zoomed in view of this region. From this zoomed image, we can clearly see that the GPS coordinate of this brown target (denoted as red dot) is shifted one pixel from the actual brown target location (denoted as yellow rectangle). This is a rare example where we can visually see the brown target. Most of the targets are difficult to distinguish visibly since they are highly mixed and sub-pixel.}

In this paper, we model the hyperspectral target estimation task as a multiple instance concept learning problem \cite{Dietterich:1997, Maron:1998} and present the multiple instance hybrid estimator (MI-HE) for characterization of a target signature from imprecisely labeled hyperspectral imagery. Here, \emph{concepts} refer to the generalized class prototypes in the feature space. In the case of hyperspectral image unmixing, a concept is the spectral signature of one assumed pure material in the scene, also known as an endmember. MI-HE explicitly addresses the aforementioned problems from the following aspects: 

1. \textbf{Uncertain labels}. MI-HE adopts the idea of ``bags'' from multiple instance learning (MIL), introduces the multiple instance data mixing model and proposes a multiple instance concept learning framework to address the uncertain labels in hyperspectral target detection, \eg, target locations coming from GPS or region of interest manually denoted by analysts. 

2. \textbf{Unbalanced number of target and non-target pixels in training data}. MI-HE addresses this problem by applying a signature-based detector only to the pixels from possible target regions denoted by a GPS in the training step. The non-target regions are only needed to refine the background concepts. 

3. \textbf{Mixed training data.} MI-HE models each pixel as a linear combination of target and/or non-target concepts, so the estimated target and background concepts applied to testing data can be used to perform sub-pixel detection.

Several previous methods for target characterization have been developed in the literature. The FUnctions of Multiple Instance (FUMI) algorithms \cite{Zare:2015fumi, jiao2016ICPR} learn representative concept from reconstruction error of the uncertainly labeled data. Compared with FUMI, MI-HE learns a discriminative target concept that maximizes the detection response of possible target regions so the estimated signatures are more discriminative for target detection. Furthermore, the FUMI algorithms do not exploit the complete label information from the training data, \ie, the FUMI algorithms combine all possible target regions together into one target bag and thus discard the information that each target region contains at least one target instance. Thus is not applicable to some target detection problems, \eg, negative bags cannot provide the entire background information. On the contrary, MI-HE adopts a generalized mean model to differentiate each individual target region. Compared to the discriminative target characterization algorithms, MI-ACE and MI-SMF \cite{zare2016miace} (which maximize the matched filter response in an MIL framework and estimates only one target signature),  MI-HE has the potential to learn multiple signatures to address signature variability.

Expanding upon our work in \cite{jiao2017MIHE}, the MI-HE algorithm and experiments presented in this paper maintain the following improvements and advantages in comparison to our prior work: (1) introducing a discriminative concept learning term; (2) improved gradient descent optimization using Armijo's rule; (3) comprehensive experiments on multiple concepts learning (both simulated and realistic) and more types of targets for Gulfport data; (4) comprehensive comparison with state-of-the-art MIL algorithms; (5) analysis of MI-HE's robustness to parameter setting.

\section{Related Methods}
Multiple instance learning (MIL) was first investigated by Dietterich \etal \cite{Dietterich:1997} in the 1990s for the prediction of drug activity (musk activity). In MIL, training data is partitioned into sets of labeled ``bags'' (instead of being individually labeled)  in which a bag is defined to be a multi-set of data points. A positive bag must contain at least one true positive (target) data point and negative bags are composed entirely of negative data.  Thus, data point-specific training labels are unavailable.  Given training data in this form, the majority of MIL methods  either: (1) learn target concepts for describing the target class; or (2) train a classifier that can distinguish between individual target and non-target data points and/or bags.

Since the introduction of the MIL framework \cite{Dietterich:1997}, many methods have been proposed and developed in the literature. The majority of MIL approaches focus on learning a classification decision boundary to distinguish between positive and negative instances/bags from the ambiguously labeled data. The mi-SVM \cite{andrews2002support} models the MIL problem as a generalized mixed integer formulation of support vector machine and was solved iteratively between training a regular SVM and heuristic reassignment of the training labels. The Multiple-Instance Learning via Embedded Instance Selection (MILES) \cite{chen2006miles} relaxes the constraint in MIL that negative bags are composed of all negative instances and allows target concept to be related to negative bags for a more general application in computer vision. MILES proposes to first embed each bag to a target concept based feature space, where the set of candidate target concept comes from the union of all bags. Then a 1-norm SVM \cite{zhu20041} is trained on the feature vectors extracted from each bag; finally, an instance selection is performed based on the SVM decision value to realize instance-level classification. 


Although the above mentioned approaches are effective for training classifiers given imprecise labels, they generally do not provide an intuitive description or \textit{representative concept} that characterizes the salient and discriminative features of the target class.  The few existing approaches that estimate a target concept include Diverse Density (DD) \cite {Maron:1998} that estimates a concept by minimizing its distance to at least one instance from each positive bag and maximizing its distance from all instances in negative bags. The Expectation-Maximization (EM) version of diverse density (EM-DD) \cite{Zhang:2002} iteratively estimates which instances in each positive bag belong to the target class and then only uses those points from the positive bags to estimate a target concept that maximizes the diverse density.  $e$FUMI \cite{Zare:2014whispers, Zare:2015fumi} treats each instance as a convex combination of positive and/or negative concepts and estimates the target and non-target concepts using an EM approach.  The MI-SMF and MI-ACE \cite{zare2016miace} maximize the response of SMF (spectral matched filter \cite{Theiler:2006,Nasrabadi:2008}) and ACE (adaptive cosine/coherent estimator \cite{Kraut:1999, kraut:2001}) respectively under a multiple instance learning framework and efficiently learn a discriminative target signature. However, most of these prototype-based methods only find a single target concept and are, thus, unable to account for large variation in the target class. 

\section{Multiple Instance Hybrid Estimator}
Let $\mathbf{X}=\left[\mathbf{x}_1,\cdots,\mathbf{x}_N\right]\in\mathbb{R}^{d\times N}$ be training data where $d$ is the dimensionality of an instance and $N$ is the total number of training instances. The data are grouped and arranged into $K$ \textit{bags},  $\mathbf{B} = \left\{ \mathbf{B}_1, \ldots, \mathbf{B}_K\right\}$, such that the first $K^+$ bags are positively labeled with associated binary bag-level label $L_i=1, i=1,\cdots, K^+$, and the rest $K^-$ bags are negatively labeled with associated binary bag-level label $L_i=0, i=K^++1,\cdots, K$. $N^+$ and $N^-$ are the total number of positive instances and negative instances, as indicated below: 
\begin{equation}
N=N^++N^-=\sum_{i=1}^{K^+}N_i+\sum_{i=K^++1}^{K}N_i,
\label{eq:N_decompose}
\end{equation}
where $N_i$ is the number of instances in bag $\mathbf{B}_i$. $\mathbf{x}_{ij} \in \mathbf{B}_i$ denotes the $j^{th}$ instance in bag $\mathbf{B}_i$ with instance-level label $l_{ij}\in\left\{ 0, 1\right\}$.

Since the MIL problem states that there must be at least one positive instance in each positive bag and each negative bag must consist of only negative instances, we can approximate the probability of an individual bag to the instances in each bag, as shown in Eq. \eqref{eq:MI-HE_inst_likelihood}. Specifically, the probability for a positive bag to be positive is substituted by the instance in this bag with highest ``positiveness'' and the probability for a negative bag to be negative is represented by the joint probability of all instances in this bag to be negative. 

\begin{equation}
J_1=\prod_{i=1}^{K^+} \max_{\mathbf{x}_{ij}\in \mathbf{B}_i}\Pr(l_{ij}=+|\mathbf{B}_i) \prod_{i=K^++1}^{K}\prod_{j=1}^{N_i}\Pr(l_{ij}=-|\mathbf{B}_{i}).
\label{eq:MI-HE_inst_likelihood}
\end{equation}

Eq. \eqref{eq:MI-HE_inst_likelihood} contains a $\max$ operation that is difficult to optimize numerically. Some algorithms in the literature \cite{maron1998multiple, Zhang:2002} adopt a noisy-OR model instead of using $\max$. However, experimental results show that the noisy-OR model is highly non-smooth and needs to be repeated with many different initializations (typically using every positive training instance) to avoid local optima.  {In order to find some alternatives for $\max$, a lot of effort has been spent. Tu \etal \cite{babenko2008simultaneous} summarized four methods for transferring the max operation to differentiable a `softmax'.  The generalized mean model \cite{bullen1988means} is one such `softmax' solution.  The generalized mean has been applied as an alternative to max in many MIL problems including cancer prediction \cite{xu2012multiple, quellec2016multiple}, image segmentation \cite{wu2014milcut, kraus2016classifying} and face recognition \cite{wohlhart2011multiple}.}  In our proposed approach, we adopt the generalized mean as an alternative to $\max$ operation in Eq. \eqref{eq:MI-HE_inst_likelihood}, as shown in Eq. \eqref{eq:MI-HE_gen_mean}.

\begin{footnotesize}
	\begin{equation}
	J_2=\prod_{i=1}^{K^+} \left(\frac{1}{N_i}\sum_{j=1}^{N_i}\Pr(l_{ij}=+|\mathbf{B}_i)^b\right)^{\frac{1}{b}}\prod_{i=K^++1}^{K}\prod_{j=1}^{N_i}\Pr(l_{ij}=-|\mathbf{B}_{i}),
	\label{eq:MI-HE_gen_mean}
	\end{equation}
\end{footnotesize}
where $b\in [-\infty, +\infty]$ is a real number controlling the function to approximately vary from $\min$ to $\max$.

Then taking the negative logarithm and scaling the second term of Eq. \eqref{eq:MI-HE_gen_mean} result in:
\begin{eqnarray}
-\ln J_2&=&-\sum_{i=1}^{K^+}\frac{1}{b}\ln \left(\frac{1}{N_i}\sum_{j=1}^{N_i}\Pr(l_{ij}=+|\mathbf{B}_i)^b\right)\nonumber\\ 
&-&\rho \sum_{i=K^++1}^{K}\sum_{j=1}^{N_i}\ln\Pr(l_{ij}=-|\mathbf{B}_{i}),
\label{eq:MI-HE_neg_log}
\end{eqnarray}
where the scaling factor $\rho$ is usually set to be smaller than one to control the influence of negative bags. 

Here each instance is modeled as a sparse linear combination of target and/or background concepts $\mathbf{D}$, $\mathbf{x}\approx\mathbf{D}\mathbf{a}$, where $\mathbf{D}=\begin{bmatrix}\mathbf{D}^+ & \mathbf{D}^-\end{bmatrix}\in\mathbb{R}^{d\times (T+M)}$, $\mathbf{D}^+ = \left[\mathbf{d}_{1},\cdots,\mathbf{d}_{T}\right]$ is the set of $T$ target  concepts and $\mathbf{D}^- = \left[\mathbf{d}_{T+1},\cdots,\mathbf{d}_{T+M}\right]$ is the set of $M$ background concepts, $\mathbf{a}$ is the sparse vector of  weights for instance $\mathbf{x}$. Solving $\mathbf{a}$ for data $\mathbf{x}$ given dictionary set $\mathbf{D}$ is modeled as a Lasso problem \cite{tibshirani1996regression, chen2001atomic} shown in Eq. \eqref{eq:MI-HE_lasso}:
\begin{equation}
{\mathbf{a}^*}=\arg\min\frac{1}{2}\|\mathbf{x}-\mathbf{D}\mathbf{a}\|^2_2+\lambda\|\mathbf{a}\|_1,
\label{eq:MI-HE_lasso}
\end{equation}
where $\lambda$ is a regularization constant to control the sparsity of $\mathbf{a}$. The solving of $l_1$ regularized least squares have been investigated extensively in the literature \cite{mallat1999wavelet,bach2012optimization, mairal2014sparse}. Here we adopt the iterative shrinkage-thresholding algorithm (ISTA) \cite{figueiredo2003algorithm, daubechies2003iterative} for solving the sparse codes  $\mathbf{a}$.
	
{Since in MIL the positive bags are mixture of both true positive points and false positive points, the hybrid detector \cite{broadwater2004hybrid, Broadwater:2007} was introduced to the proposed objective function \eqref{eq:MI-HE_neg_log} to determine if instances from positive bags are the true positive points. Specifically, define the following detection statistic as an approximation of $\Pr(l_{ij}=+|\mathbf{B}_i)$,}

\begin{equation}
\Lambda(\mathbf{x}_{ij},\mathbf{D}|\mathbf{B}_i,L_i=1)=\exp\left(-\beta\frac{\|\mathbf{x}_{ij}-\mathbf{D}\mathbf{a}_{ij}\|^2}{\|\mathbf{x}_{ij}-\mathbf{D}^-\mathbf{p}_{ij}\|^2}\right),
\label{eq:MI-HE_pos_inst_model}
\end{equation}
where $\beta$ is a scaling parameter, $\mathbf{a}_{ij}$ and $\mathbf{p}_{ij}$ are the sparse representation of $\mathbf{x}_{ij}$ given the entire concept set $\mathbf{D}$ and background concept set $\mathbf{D}^-$. Further, $\mathbf{a}_{ij}=[{\mathbf{a}}^+_{ij};\; {\mathbf{a}}^-_{ij}]$, where ${\mathbf{a}}^+_{ij}$ and ${\mathbf{a}}^-_{ij}$ are sub-vectors of $\mathbf{a}_{ij}$ corresponding to $\mathbf{D}^+$ and $\mathbf{D}^-$, respectively. Since $\mathbf{D}$ is a super set of $\mathbf{D}^-$, theoretically the reconstruction error of $\mathbf{x}_{ij}$ using $\mathbf{D}$ (the numerator) should not be greater than that using $\mathbf{D}^-$ (the denominator). The residual vectors,
\begin{eqnarray}
\left\{ \begin{array}{l}
\mathbf{r}_{ij}=(\mathbf{x}_{ij}-\mathbf{D}\mathbf{a}_{ij}) \\
\mathbf{q}_{ij}=(\mathbf{x}_{ij}-\mathbf{D}^-\mathbf{p}_{ij})
\label{eqn:residual_vects} 
\end{array}\right.,
\end{eqnarray}
are the reconstruction residual vectors corresponding to $\mathbf{D}$ and $\mathbf{D}^-$, respectively.

{The hybrid detector $\Lambda(\mathbf{x}_{ij},\mathbf{D}|\mathbf{B}_i,L_i=1)$ defined in Eq. \eqref{eq:MI-HE_pos_inst_model} indicates that if a point $\mathbf{x}_{ij}\in \mathbf{B}_i,L_i=1$, is a false positive point ($l_{ij}=0$) in a positively labeled bag, it should be well represented by both the non-target concepts $\mathbf{D}^-$ and the entire concepts $\mathbf{D}$, respectively. Since this false positive point is a none-target point, the residual errors by both reconstruction methods should be of the same magnitude, \ie,  $\|\mathbf{r}_{ij}\|^2\approx \|\mathbf{q}_{ij}\|^2$. After being scaled by $\beta$, the detection statistic of the hybrid detector for this false positive point, $\Lambda(\mathbf{x}_{ij},\mathbf{D}|\mathbf{B}_i, L_i=1)=\exp\left(-\beta\frac{\|\mathbf{r}_{ij}\|^2}{\|\mathbf{q}_{ij}\|^2}\right)\rightarrow 0$. However, if $\mathbf{x}_{ij}\in \mathbf{B}_i, L_i=1$, is a true positive point ($l_{ij}=1$) in the positively labeled bag, it should not be well represented by only the non-target concepts, so the residual error approximated by the entire concepts, $\|\mathbf{r}_{ij}\|^2$, will be much smaller than that by the background concepts, $\|\mathbf{q}_{ij}\|^2$, thus the detection for this true positive point, $\Lambda(\mathbf{x}_{ij},\mathbf{D}|\mathbf{B}_i,L_i=1)=\exp\left(-\beta\frac{\|\mathbf{r}_{ij}\|^2}{\|\mathbf{q}_{ij}\|^2}\right)\rightarrow 1$. Here, the scaling factor $\beta$ for this case is no longer consequential since $\|\mathbf{r}_{ij}\|^2\ll\|\mathbf{q}_{ij}\|^2$.}

{For points from negative bags, the negative logarithm of $\Pr(l_{ij}=-|\mathbf{B}_{i}, L_i=0)$ is modeled as the least squares of $\mathbf{x}_{ij}$ shown in Eq. \eqref{eq:MI-HE_neg_inst_model}, \\
	\begin{equation}
	-\ln\Pr(l_{ij}=-|\mathbf{B}_{i}, L_i=0)=\|\mathbf{x}_{ij}-\mathbf{D}^-\mathbf{p}_{ij}\|^2,
	\label{eq:MI-HE_neg_inst_model}
	\end{equation}
}\\
where $\mathbf{p}_{ij}$ is the sparse representation of $\mathbf{x}_{ij}$ given $\mathbf{D}^-$ and is solved by Eq. \eqref{eq:MI-HE_lasso}. Here instead of applying the hybrid detector, we use least squares to represent the residual error of $\mathbf{x}_{ij}$. This indicates that the negative points should be fully represented by only the non-target concepts, $\mathbf{D}^-$. The intuitive understanding of this assumption is that minimizing the least squares of all of the negative points provides a good description of the background. Moreover, because there are typically many more negative points in negatively labeled bags than true positive points in positive bags, the target concept estimation may be biased if the hybrid detector was also applied to negative instances. 

{Thus far, we have defined the probability terms in objective function \eqref {eq:MI-HE_neg_log} which aims to learn a set of concepts that maximize the hybrid sub-pixel detector statistic of the positive bags and characterize the negative bags.} However, given the objective function so far, there is no guarantee that the estimated target concept captures only the distinct features of the positive class and is discriminative from the negative class. Inspired by the discriminative terms proposed by the Dictionary Learning with Structured Incoherence \cite{ramirez2010classification} and the Fisher Discrimination Dictionary Learning (FDDL) algorithm \cite{yang2011fisher, yang2014sparse}, we propose a cross incoherence term $Q(\mathcal{X}, \mathbf{D}^+, {\mathcal{A}})$ shown in Eq. \ref{eq:MI-HE_discr_term} to complete the objective, where $\mathcal{X}$ is the concatenation of all instances from negatively labeled bags, $\mathbf{D}^+$ is the target concept set which is the subset of $\mathbf{D}=[\mathbf{D}^+\; \mathbf{D}^-]$,  $\mathcal{A}=[{\mathcal{A}}^+ \; {\mathcal{A}}^-]$ is the sparse codes matrix of $\mathcal{X}$ with respect to the entire concepts $\mathbf{D}$. 

\begin{eqnarray}
Q(\mathcal{X}, \mathbf{D}^+, {\mathcal{A}})&=&\frac{\alpha}{2}\|Diag\left((\mathbf{D}^+{\mathcal{A^+}})^T\mathcal{X}\right)\|^2_2\nonumber \\
&=&\frac{\alpha}{2}\sum_{i=K^++1}^{K}\sum_{j=1}^{N_i}\left((\mathbf{D}^+{\mathbf{a}}_{ij}^+)^T\mathbf{x}_{ij}\right)^2
\label{eq:MI-HE_discr_term}
\end{eqnarray}

The understanding of the proposed cross incoherence term is presented by examining the reconstruction of the negative data set $\mathcal{X}$.  First of all,  $\mathcal{X}$ should be well represented by the non-target concept set $\mathbf{D}^-$, \ie,  $\mathcal{X}\approx \mathbf{D}^- \mathbf{P}$. This is fulfilled by inclusion of the term in Eq. \eqref{eq:MI-HE_neg_inst_model}. Second, since $\mathbf{D}=[\mathbf{D}^+\; \mathbf{D}^-]$ is a superset of $\mathbf{D}^-$, the reconstruction error of $\mathcal{X}$ by the entire concept set $\mathbf{D}$ is also small, \ie, $\mathcal{X}\approx \mathbf{D}^+ {\mathcal{A}}^{+}+\mathbf{D}^- {\mathcal{A}}^{-}={\mathbf{R}}^{+}+{\mathbf{R}}^{-}$. In order to have a target concept $\mathbf{D}^+$ that is distinct from the negative data, it is expected that the reconstruction of $\mathcal{X}$ with respect to the target concept, ${\mathbf{R}}^{+}$, should either maintain small energy or else have a bad representation of $\mathcal{X}$, and thus Eq. \ref{eq:MI-HE_discr_term} is optimized. 

The final objective function is shown in Eq. \ref{eq:MI-HE_full_neg_log}, which contains three terms: generalized mean (GM) term (first), background data fidelity term (second) and the cross incoherence (discriminative) term (third):

\begin{eqnarray}
J_3&=&-\sum_{i=1}^{K^+}\frac{1}{b}\ln \left(\frac{1}{N_i}\sum_{j=1}^{N_i}\exp\left(-\beta\frac{\|\mathbf{x}_{ij}-\mathbf{D}\mathbf{a}_{ij}\|^2}{\|\mathbf{x}_{ij}-\mathbf{D}^-\mathbf{p}_{ij}\|^2}\right)^b\right)\nonumber\\
&&+\rho \sum_{i=K^++1}^{K}\sum_{j=1}^{N_i}\|\mathbf{x}_{ij}-\mathbf{D}^-\mathbf{p}_{ij}\|^2\nonumber\\
&&+\frac{\alpha}{2}\sum_{i=K^++1}^{K}\sum_{j=1}^{N_i}\left((\mathbf{D}^+{\mathbf{a}}^+_{ij})^T\mathbf{x}_{ij}\right)^2,
\label{eq:MI-HE_full_neg_log}
\end{eqnarray}

\section{Optimization}
The optimization of Eq. \eqref{eq:MI-HE_full_neg_log} can be decomposed into two sub-problems, updating the concepts $\mathbf{D}$ and the sparse representation $\mathbf{a}$ alternatively. The algorithm stops when the number of preset iterations is reached or the change between two iterations is smaller than a preset threshold. The method is summarized in Alg. \ref{alg:MI-HE} \cite{zare_MIHE_code:2018}. For readability, the derivation of update equations are described in \ref {MIHE_optimiz}.

\begin{algorithm}
	\caption{MI-HE algorithm}
	\algsetup{indent=4em}
	\begin{algorithmic}[1] 
		\REQUIRE  {MIL training bags $\mathbf{B} = \left\{ \mathbf{B}_1, \ldots, \mathbf{B}_K\right\}$, MI-HE parameters}
		\STATE Initialize $\mathbf{D}^0$, $iter = 0$
		\REPEAT
		\FOR{$t=1,\cdots,T$}
		\STATE Solve $\mathbf{a}_{ij}$,  $\mathbf{p}_{ij}$ according to \eqref{eq:alpha_update_final}, $\forall i\in\{1,\cdots,K\}, j\in\{1, \cdots, N_i\}$
		\STATE Update  $\mathbf{d}_t$ using gradient descent according to \eqref{eq:gradient_dt}
		\STATE $\mathbf{d}_t\gets\frac{1}{\|\mathbf{d}_t\|_2}\mathbf{d}_t$
		\ENDFOR
		\FOR{$k=T+1,\cdots,T+M$}
		\STATE Solve $\mathbf{a}_{ij}$,  $\mathbf{p}_{ij}$ according to \eqref{eq:alpha_update_final}, $\forall i\in\{1,\cdots,K\}, j\in\{1, \cdots, N_i\}$
		\STATE Update  $\mathbf{d}_k$ using gradient descent according to \eqref{eq:gradient_dk}
		\STATE $\mathbf{d}_k\gets\frac{1}{\|\mathbf{d}_k\|_2}\mathbf{d}_k$
		\ENDFOR
		\STATE $iter \gets iter + 1$
		\UNTIL{ Stopping criterion reached }
		\RETURN $\mathbf{D}$\\
	\end{algorithmic} 
	\label{alg:MI-HE}
\end{algorithm}

The initialization of target concepts in $\mathbf{D}$ is conducted by computing the mean of $T$ random subsets drawn from the union of all positive training bags. VCA (vertex component analysis \cite{nascimento:2005}) was applied to the union of all negative bags and the $M$ cluster centers (or vertices) were set as the initial background concepts.

\section{Experimental Results}
In the following, MI-HE is evaluated and compared to several MIL concept learning methods on simulated data and to a real hyperspectral target detection data set. The simulated data experiments are included to illustrate the properties of MI-HE and provide insight into how and when the methods are effective. 

{For the experiments conducted in this paper, the parameter settings of the proposed MI-HE and comparison algorithms were optimized using a grid search on the first task of each experiment and then applied to the remaining tasks. For example, for mi-SVM classifier on the Gulfport Brown target task, the $\gamma$ value of the RBF kernel was firstly varied from 0.5 to 5 at a step size of 0.5, and then a finer search around the current best value (with the highest AUC) at a step of 0.1 was performed. For algorithms with stochastic result, \eg, EM-DD, eFUMI, each parameter setting was run five times and the median performance was selected. Finally the optimal parameters that achieve the highest AUC for the brown target were selected and used for the other three target types. }

\subsection{Simulated Data}
As we discussed before, $e$FUMI combines all positive bags as one big positive bag and all negative bags as one big negative bag and learns target concept from the big positive bag that is different from the negative bag. So if the negative bags maintain incomplete knowledge of the background, \eg, some non-target concept appears only in the subset of positive bags, $e$FUMI will perform poorly. However, MI-HE which maintains bag structure will be able to distinguish the target. 

\begin{figure}[htb]
	\centering
	\includegraphics[width=10cm]{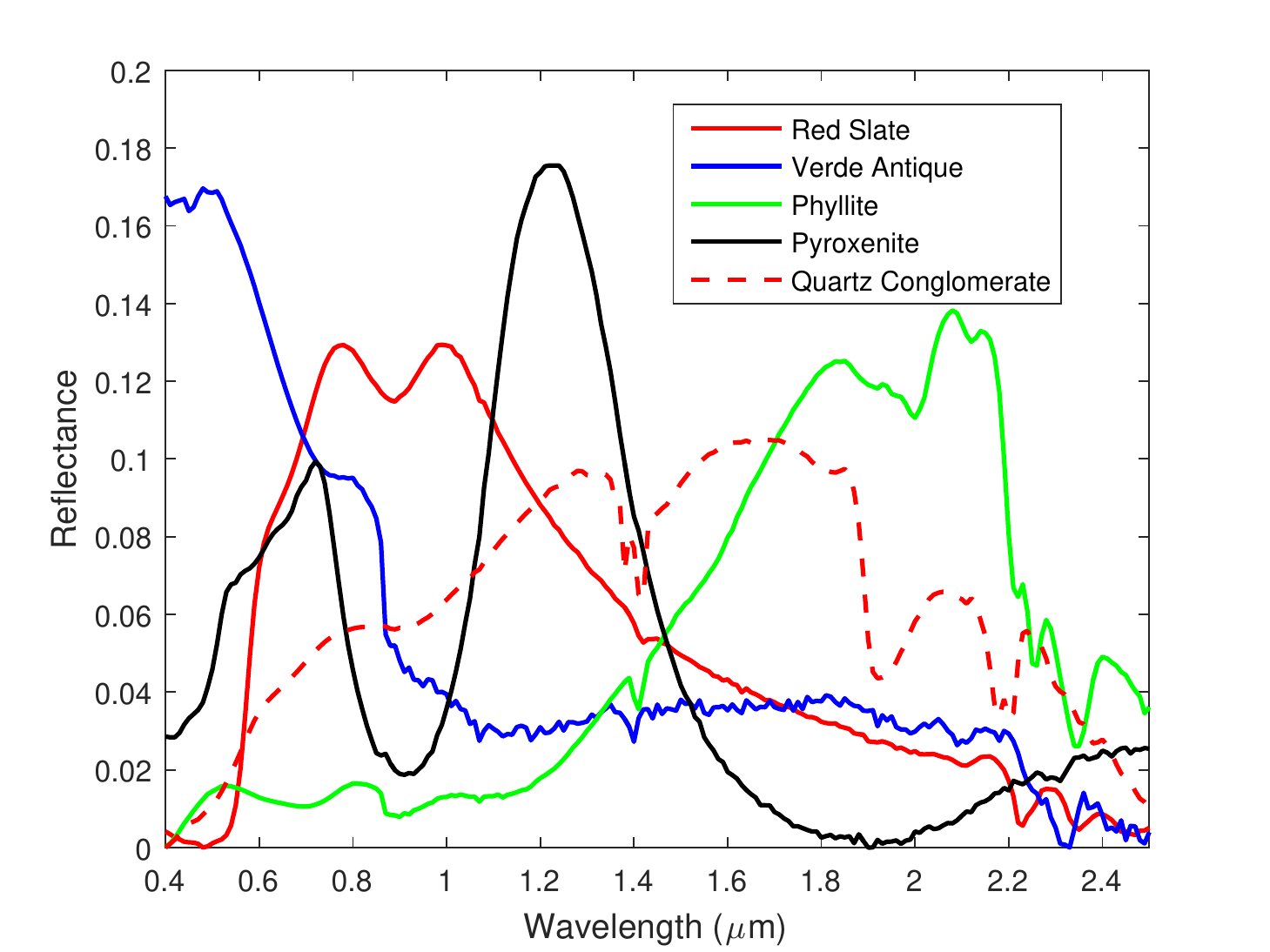}
	\caption{Signatures from ASTER library used to generate simulated data \label{fig:constituent_endmembers_extra1} } 
\end{figure}

Given this hypothesis, simulated data was generated from four spectra selected from the ASTER spectral library \cite{aster:2009}. Specifically, the Red Slate, Verde Antique, Phyllite and Pyroxenite spectra from the rock class with 211 bands and wavelengths ranging from $0.4 \mu$m to $2.5 \mu$m (as shown in Fig. \ref{fig:constituent_endmembers_extra1} in solid lines) were used as endmembers to generate hyperspectral data. Red Slate was labeled as the target endmember.

\begin{table} 
	\begin{footnotesize}
		\begin{center}
			\caption{List of Constituent Endmembers for Synthetic Data with Incomplete Background Knowledge}\label{tab:toydata_endmember_list}
			\begin{tabular}{|c|c|c|c|}
				\hline
				Bag No. 	&  Bag Label  &Target Endmember& Background Endmember \\
				\hline\hline
				\multirow{2}{*}{1-5} &    \multirow{2}{*}{$+$}         &\multirow{2}{*}{Red Slate}  & Verde Antique, Phyllite, \\& & & Pyroxenite    \\\hline
				6-10&          $+$     &{Red Slate}     &     Phyllite, Pyroxenite                       \\\hline
				11-15&      $+$       &{Red Slate}    &      Pyroxenite                    \\\hline
				16-20&       $-$     & N/A &      Phyllite, Pyroxenite                 \\\hline
			\end{tabular}
		\end{center}
	\end{footnotesize}
\end{table}

Four sets of highly-mixed noisy data with varied mean target proportion value ($\boldsymbol{\alpha}_{t\_mean}$) were generated, a detailed generation process can be found in \cite{Zare:2015fumi}. Specifically, this synthetic data has 15 positive and 5 negative bags with each bag having 500 points. If it is a positively labeled bag, there are 200 highly-mixed target points containing mean target (Red Slate) proportion from to 0.1 to 0.7 respectively to vary the level of target presence from weak to high. Gaussian white noise was added so that signal-to-noise ratio of the data was set to $20 dB$. To highlight the ability of MI-HE to leverage individual  bag-level labels, we use different subsets of background endmembers to build synthetic data as shown in Tab. \ref{tab:toydata_endmember_list}. Tab. \ref{tab:toydata_endmember_list} shows that the negatively labeled bags only contain 2 negative endmembers and there exists one confusing background endmember in the first 5 positive bags which is Verde Antique. It is expected that the proposed MI-HE will be able to learn the target signature correctly and $e$FUMI will confuse both Red Slate and Verde Antique as target signatures.

The parameter settings of MI-HE for this experiment are $T = 1, M = 9, \rho = 0.8, b = 5, \beta=5$ and $\lambda=1\times 10^{-3}$. MI-HE was compared to state-of-the-art MIL algorithms $e$FUMI \cite{Zare:2014whispers, Zare:2015fumi}, MI-SMF and MI-ACE \cite{zare2016miace}, DMIL \cite{shrivastava2015gen,shrivastava2014dictionary}, EM-DD \cite{Zhang:2002} and mi-SVM \cite{andrews2002support}. The mi-SVM algorithm was added to these experiments to include a comparison MIL approach that does not rely on estimating a target signature.

\begin{figure}
	\begin{center}
		\subfloat[{Estimated target signatures for Red Slate and comparison with ground truth}]{   
			\includegraphics[width=9cm]{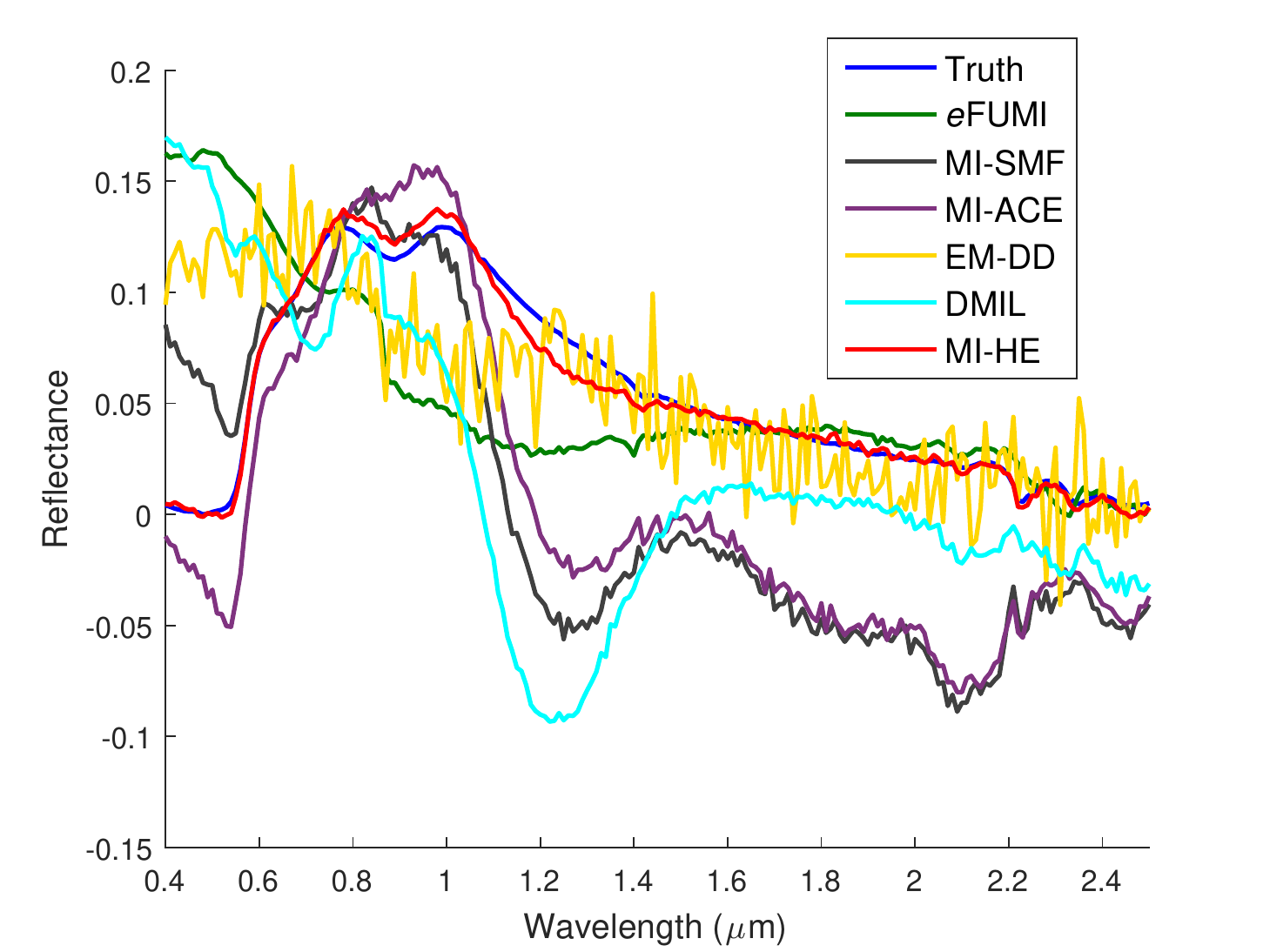} \label{fig:sig_plot_toydata_ptmean01}}\\
		\subfloat[{ROC curves cross validated on test data}]{   
			\includegraphics[width=9cm]{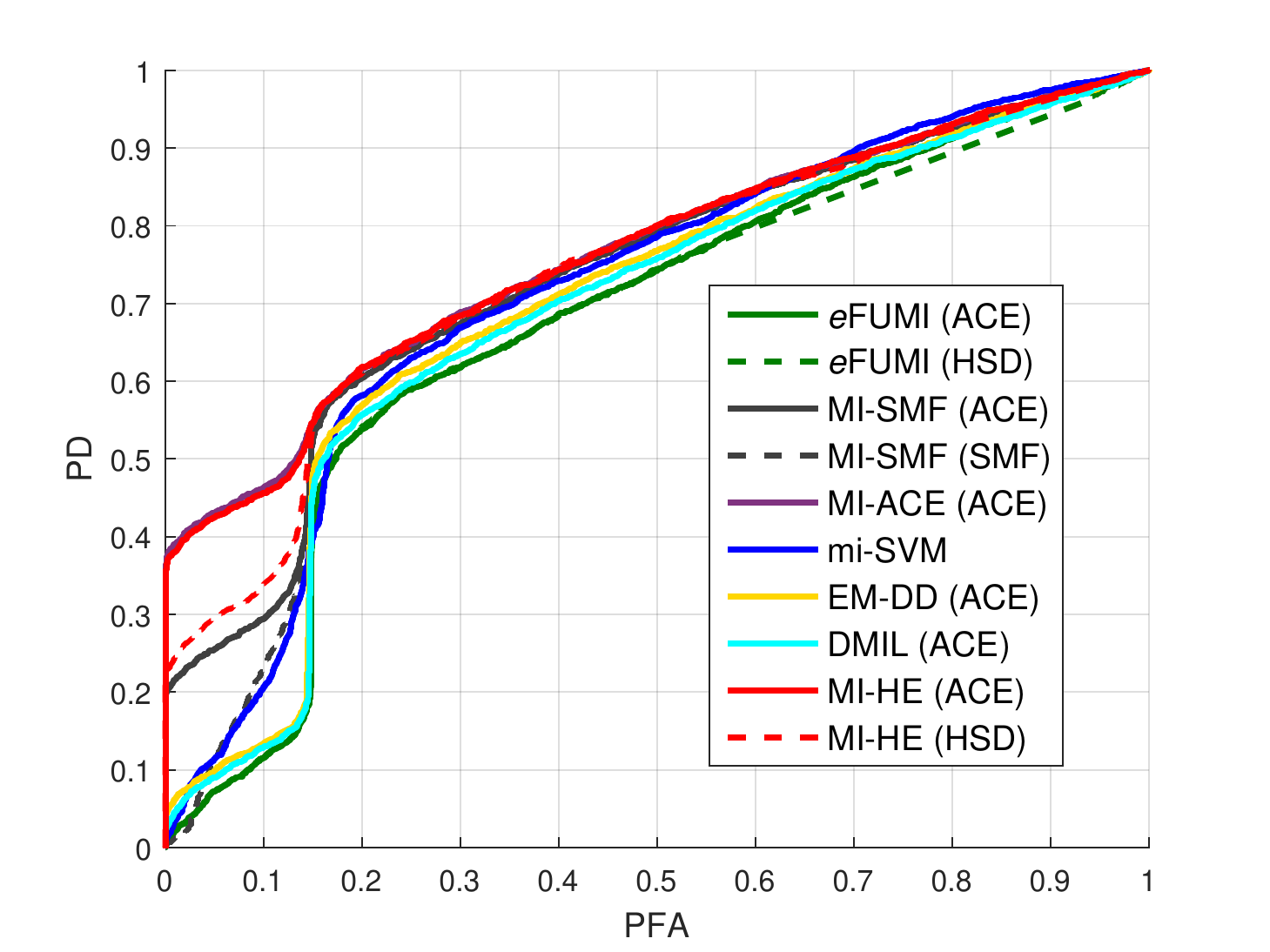} \label{fig:rocs_toydata_ptmean01}}
		\caption{{MI-HE and comparisons on synthetic data with incomplete background knowledge, $\boldsymbol{\alpha}_{t\_mean}=0.1$. MI-SMF and MI-ACE are not expected to recover the true signature.}}\label{fig:MIHE_tyodata_ptmean_01}
	\end{center}
	\vspace{-3mm}
\end{figure}

Fig. \ref{fig:sig_plot_toydata_ptmean01} shows the estimated target signature from data with 0.1 mean target proportion value. It clearly shows that the proposed MI-HE is able to correctly distinguish Red Slate as target concept from the incomplete background knowledge. Also, the other comparison algorithms can also estimate a target concept close to the groundtruth Red Slate spectrum. However, $e$FUMI is always confused with the other non-target endmember, Verde Antique, that exists in some positive bags but is excluded from the background bags. 

For simulated detection analysis, estimated target concepts from the training data were then applied to the test data generated separately following the same generating procedure. The detection was performed using the HSD or ACE detection statistic. For MI-HE and $e$FUMI, both methods were applied since those two algorithms can come out a set of background concept from training simultaneously; for MI-SMF, both SMF and ACE were applied since MI-SMF's objective is maximizing the multiple instance spectral matched filter; for the rest multiple instance target concept learning algorithms, MI-ACE, EM-EE, DMIL, only ACE was applied. For the testing procedure of mi-SVM, a regular SVM testing process was performed using LIBSVM \cite{chang2011libsvm}, and the decision values (signed distances to hyperplane) of test data determined from trained SVM model were taken as the confidence values. For the signature based detectors, the background data mean and covariance were estimated from the negative instances of the training data. 

For quantitative evaluation, Fig. \ref{fig:rocs_toydata_ptmean01} shows the  receiver operating characteristic (ROC) curves using estimated target signature, where it can be seen that the $e$FUMI is confused with the testing Verde Antique data  at very low PFA (probability of false alarms) rate. Tab. \ref{tab:AUC_toydata_overlapbags} shows the area under the curve (AUC) of proposed MI-HE and comparison algorithms. The results reported are the median results over five runs of the algorithm on the same data.  From Tab. \ref{tab:AUC_toydata_overlapbags}, it can be seen that for MI-HE and MI-ACE, the best performance on detection was achieved using ACE detector, {which is quite close to the performance of using the ground truth target signature (denoted as values with stars).} The reason that MI-HE's detection using HSD detector is a little worse is that HSD relies on knowing the complete background concept to properly represent each non-target testing data, the missing non-target concept (Verde Antique) makes the non-target testing data containing Verde Antique maintain a relatively large reconstruction error, and thus large detection statistic.

\begin{table} 
	\begin{center}
		\caption{{Area under the ROC curves for MI-HE and comparison algorithms on Simulated Hyperspectral Data with Incomplete Background Knowledge.  Best results shown in bold, second best results underlined, and ground truth shown with an asterisk.}}\label{tab:AUC_toydata_overlapbags}
		\begin{tabular}{|c|c|c|c|c|}
			\hline
			\multirow{2}{*}{Algorithm} 	&  \multicolumn{4}{c|}{$\boldsymbol{\alpha}_{t\_mean}$} \\
			\cline{2-5}&{0.1}&{0.3}&{0.5}&{0.7}\\
			\hline\hline
			{MI-HE (HSD)}   &    {0.743}                    &  \underline{0.931}         &            0.975            &   0.995     \\\hline
			{MI-HE (ACE)}   &      \underline{0.763}        &      \textbf{0.952}          &      \textbf{0.992}       &   \textbf{0.999}    \\\hline
			{$e$FUMI \cite{Zare:2015fumi} (ACE)} &      0.675                    &       0.845                  &      0.978                &   \underline{0.998}    \\\hline
			{$e$FUMI \cite{Zare:2015fumi} (HSD)}  &      0.671                   &        0.564                 &      0.978                &   \underline{0.998}     \\\hline
			{MI-SMF \cite{zare2016miace} (SMF)}  &      0.719                    &        0.923                 &      0.972                &   0.993     \\\hline
			{MI-SMF \cite{zare2016miace} (ACE)}  &     0.735                     &       \textbf{0.952}          &      \textbf{0.992}      &   \textbf{0.999}     \\\hline
			{MI-ACE \cite{zare2016miace} (ACE)}  &      \textbf{0.764}           &        \textbf{0.952}         &      \textbf{0.992}      &   \textbf{0.999}     \\\hline
			{mi-SVM} \cite{andrews2002support}       &     0.715                     &         0.815                 &      0.866               &   {0.900}    \\\hline
			{EM-DD \cite{Zhang:2002} (ACE)}   &      0.695                    &        0.918                  &      \underline{0.983}   &   \underline{0.998}    \\\hline
			{DMIL \cite{shrivastava2015gen,shrivastava2014dictionary} (ACE)}    &      0.687                    &        0.865                  &      {0.971}                &   {0.996}    \\\hline
			{{Ground Truth (ACE)}}    &      {0.765*}                    &        {0.953*}                 &      {0.992*}              &   {{0.999*}}    \\\hline
		\end{tabular}
	\end{center}
\end{table}

To formulate a multiple concept learning problem, another rock endmember, Quartz Conglomerate, from ASTER spectral library was used as the second target concept (as shown in Fig. \ref{fig:constituent_endmembers_extra1} in dashed line). Three sets of highly-mixed noisy data with varied mean target proportion value from [0.1, 0.1] to [0.3, 0.3] were generated. The synthetic data has 5 positive bags containing both target concept (Red Slate and Quartz Conglomerate)  and non-target concept (Verde Antique, Phyllite, Pyroxenite); 5 negative bags containing only the background concept (Verde Antique, Phyllite, Pyroxenite). The mean target proportion value was set to [0.1, 0.1], [0.2, 0.2] and [0.3, 0.3], respectively to vary the level of target presence from weak to high. The other settings for simulated data are the same as above mentioned. It is expected that the proposed MI-HE is able to learn multiple target concept at one time.  

\begin{figure}
	\begin{center}
		\subfloat[{Estimated target signatures for Red Slate and comparison with ground truth}]{   
			\includegraphics[width=10cm]{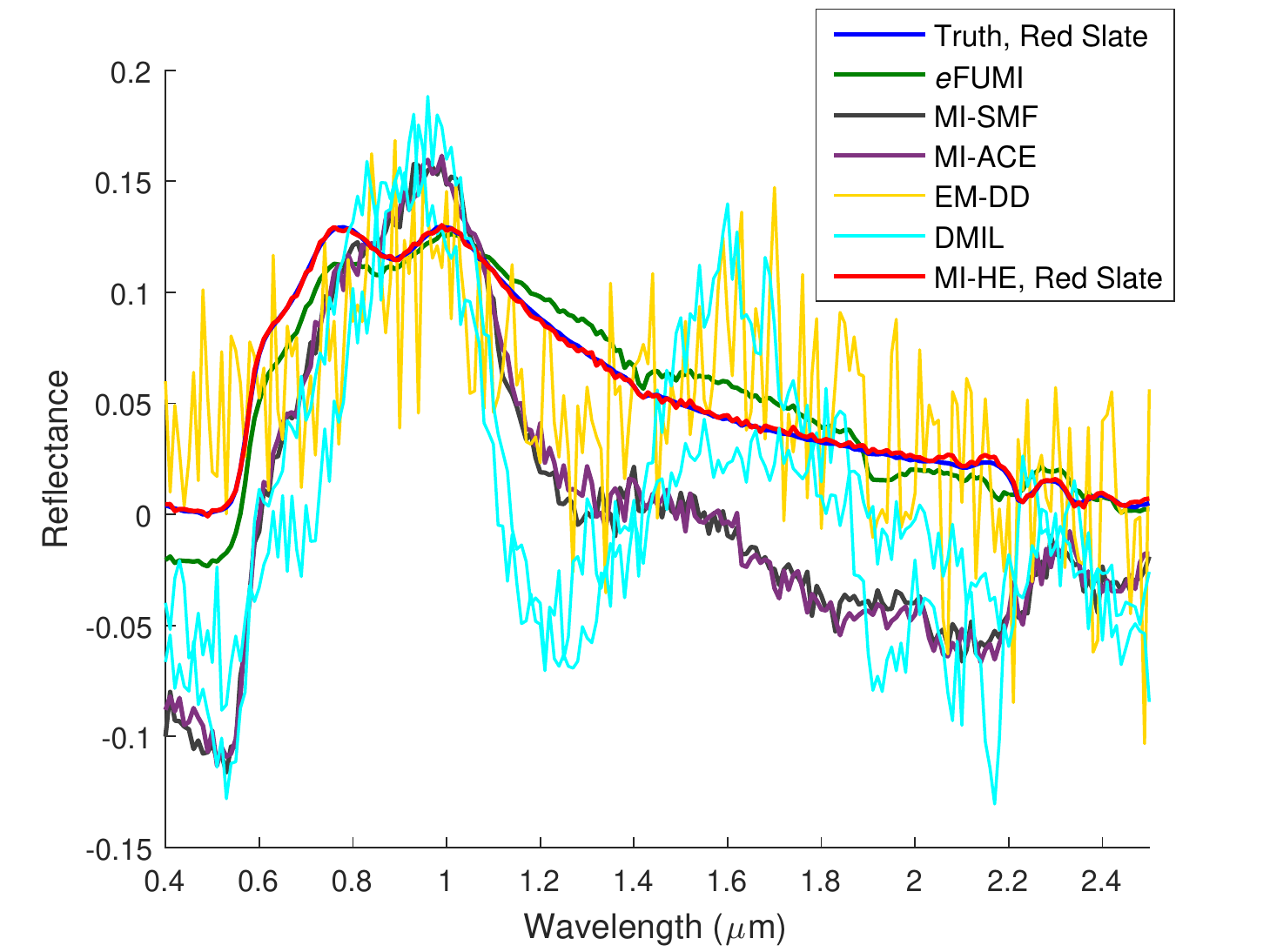} \label{fig:sig_plot_multitar_0101_RedSlate}}\\
		\subfloat[{Estimated target signatures for Quartz Conglomerate and comparison with ground truth}]{   
			\includegraphics[width=10cm]{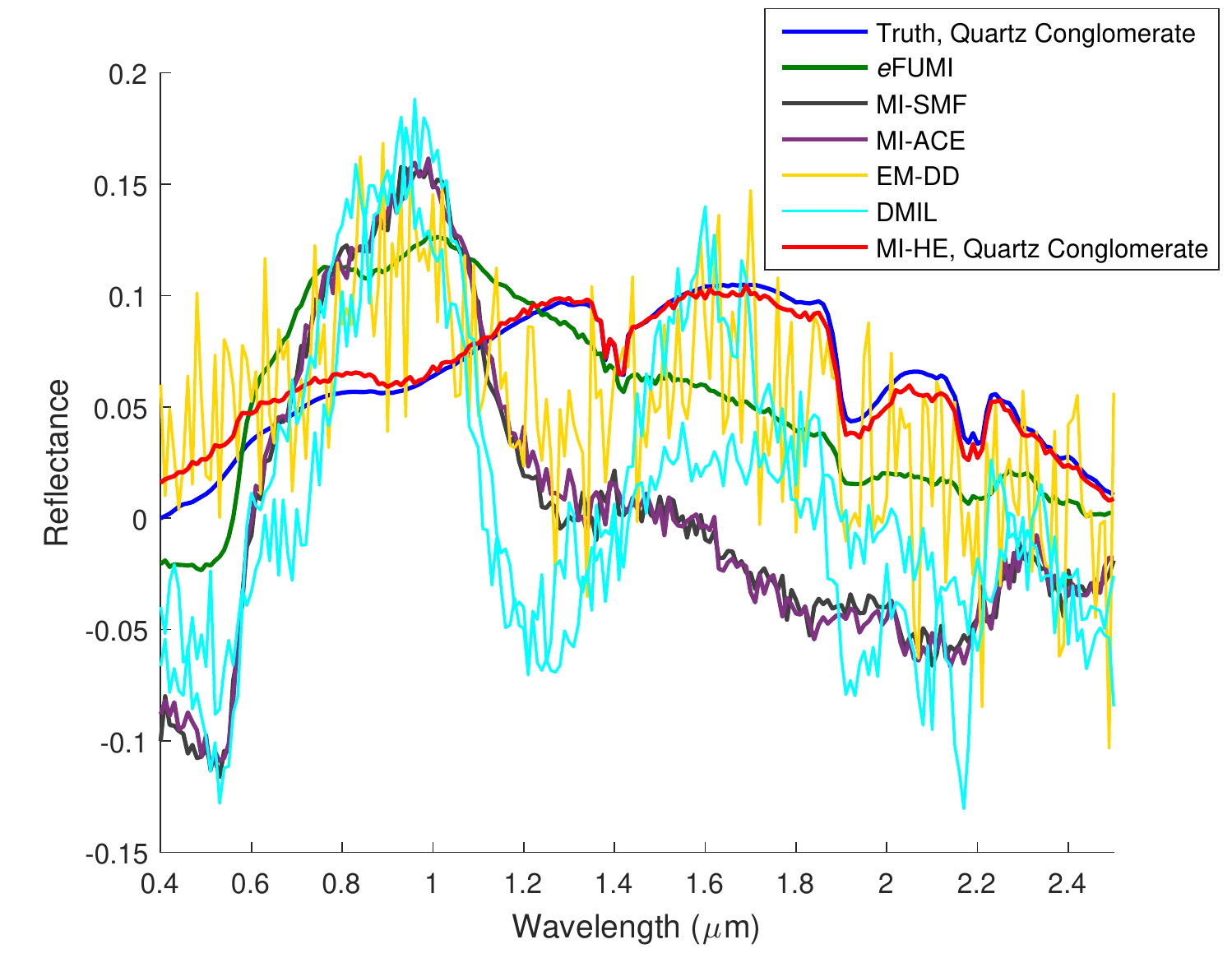} \label{fig:sig_plot_multitar_0101_QuartzConglomerate}}
		\caption{{Estimated target signatures by MI-HE and comparisons on synthetic data with multiple target concepts, $\boldsymbol{\alpha}_{t\_mean}=[0.1, 0.1]$. Not all comparisons algorithms are expected to recover true target signatures.}}\label{fig:estiamted_sigs_MIHE_multitar_0101}
	\end{center}
	\vspace{-5mm}
\end{figure}

\begin{figure}[htb]
	\centering
	\includegraphics[width=10cm]{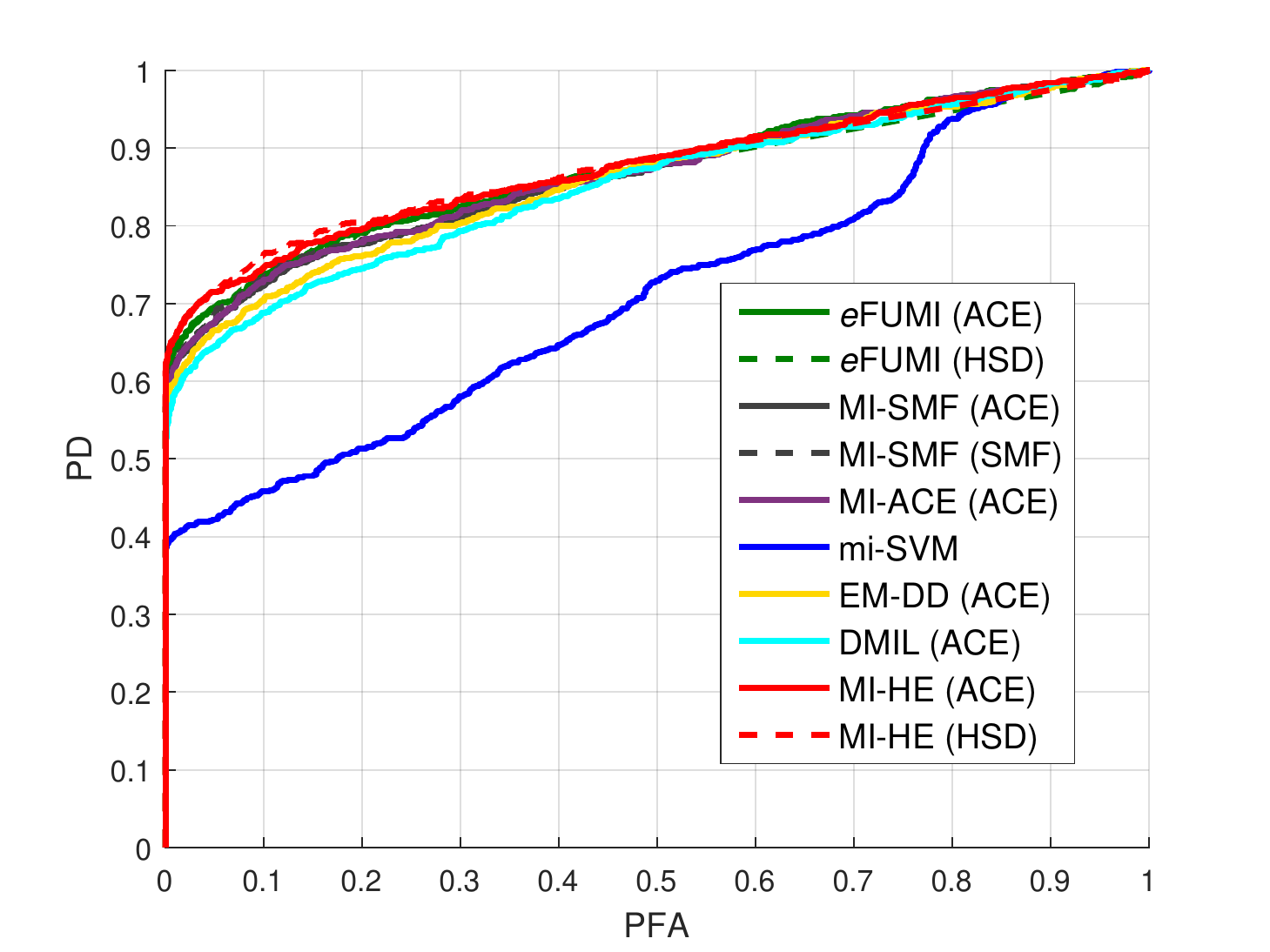}
	\caption{{ROCs of MI-HE and comparisons on synthetic data with multiple target concepts, $\boldsymbol{\alpha}_{t\_mean}=[0.1, 0.1]$.} \label{fig:rocs_MIHE_multitar_0101} } 
\end{figure}

\begin{table} 
	\begin{center}
		\caption{{Area under the ROC curves for MI-HE and comparison algorithms on Simulated Hyperspectral Data with Multiple Target Concepts.  Best results shown in bold, second best results underlined, and ground truth shown with an asterisk. }}\label{tab:AUC_toydata_multitar}
		\begin{tabular}{|c|c|c|c|}
			\hline
			\multirow{2}{*}{Algorithm} 	&  \multicolumn{3}{c|}{$\boldsymbol{\alpha}_{t\_mean}$} \\
			\cline{2-4}&{[0.1, 01]}&{[0.2, 0.2]}&{[0.3, 0.3]}\\
			\hline\hline
			{MI-HE (HSD)}   &    \textbf{0.875}          &      \textbf{0.982}             &      \textbf{0.998}            \\\hline
			{MI-HE (ACE)}   &      \textbf{0.875}        &      \textbf{0.982}             &      \textbf{0.998}             \\\hline
			{$e$FUMI \cite{Zare:2015fumi} (ACE)} &      \underline{0.872}       &   \underline{0.980}            &     \textbf{0.998}          \\\hline
			{$e$FUMI \cite{Zare:2015fumi} (HSD)}  &      0.865                   &        0.976                 &      \underline{0.997}         \\\hline
			{MI-SMF \cite{zare2016miace} (SMF)}  &      0.866                    &        0.977                 &      \underline{0.997}          \\\hline
			{MI-SMF \cite{zare2016miace} (ACE)}  &     0.865                     &       {0.976}                &      \underline{0.997}           \\\hline
			{MI-ACE \cite{zare2016miace} (ACE)}  &      {0.866}                 &        {0.976}                &      \underline{0.997}          \\\hline
			{mi-SVM} \cite{andrews2002support}        &     0.711                     &         0.890                 &      0.970                             \\\hline
			{EM-DD \cite{Zhang:2002}  (ACE)}   &      0.858                    &        0.979                  &      \textbf{0.998}           \\\hline
			{DMIL \cite{shrivastava2015gen,shrivastava2014dictionary} (ACE)}    &      0.850                    &        0.971                  &      {0.994}                           \\\hline
			{{Ground Truth (ACE)}}    &      {0.869*}                    &        {0.979*}                 &      {0.997*}               \\\hline
		\end{tabular}
	\end{center}
\end{table}

The parameter settings of MI-HE for this experiment are $T = 2, M = 9, \rho = 0.8, b = 5, \beta=5$ and $\lambda=1\times 10^{-3}$. Fig. \ref{fig:sig_plot_multitar_0101_RedSlate} and \ref{fig:sig_plot_multitar_0101_QuartzConglomerate} show the estimated target concepts by proposed MI-HE and comparisons, where we can see that the proposed MI-HE is able to accurately estimate multiple target concepts simultaneously. Compared with MI-HE, although DMIL is also a multiple concept learning algorithm, target concept as estimated by DMIL is noisy, and not a representative prototype of the target class. The remaining comparison algorithms are single target concept learning which are always confused by the multiple target concept problem.

Fig. \ref{fig:rocs_MIHE_multitar_0101} shows the ROCs using estimated target signature, and Tab. \ref{tab:AUC_toydata_multitar} shows the AUCs of proposed MI-HE and comparison algorithms. The results reported are the median results over five runs of the algorithm on the same data. Since ACE is a single signature based detector, given multiple target concepts, the maximum detection statistic across all estimated target concept was picked for each testing data. From Tab. \ref{tab:AUC_toydata_multitar}, it can be seen that the proposed MI-HE outperforms all the comparison single concept learning algorithms as well as the multiple MI concept learning algorithm DMIL. {The reason MI-HE even outperforms detection performance using ground truth signatures is that there also exists target points which are mixtures of two target concepts. The hybrid detector can directly model the mixture of target concepts outperforming the application of only one concept at a time.}

\subsection{MUUFL Gulfport Hyperspectral Data}

 \begin{figure}[tbh]
	\centering
	{\includegraphics[width=9cm]{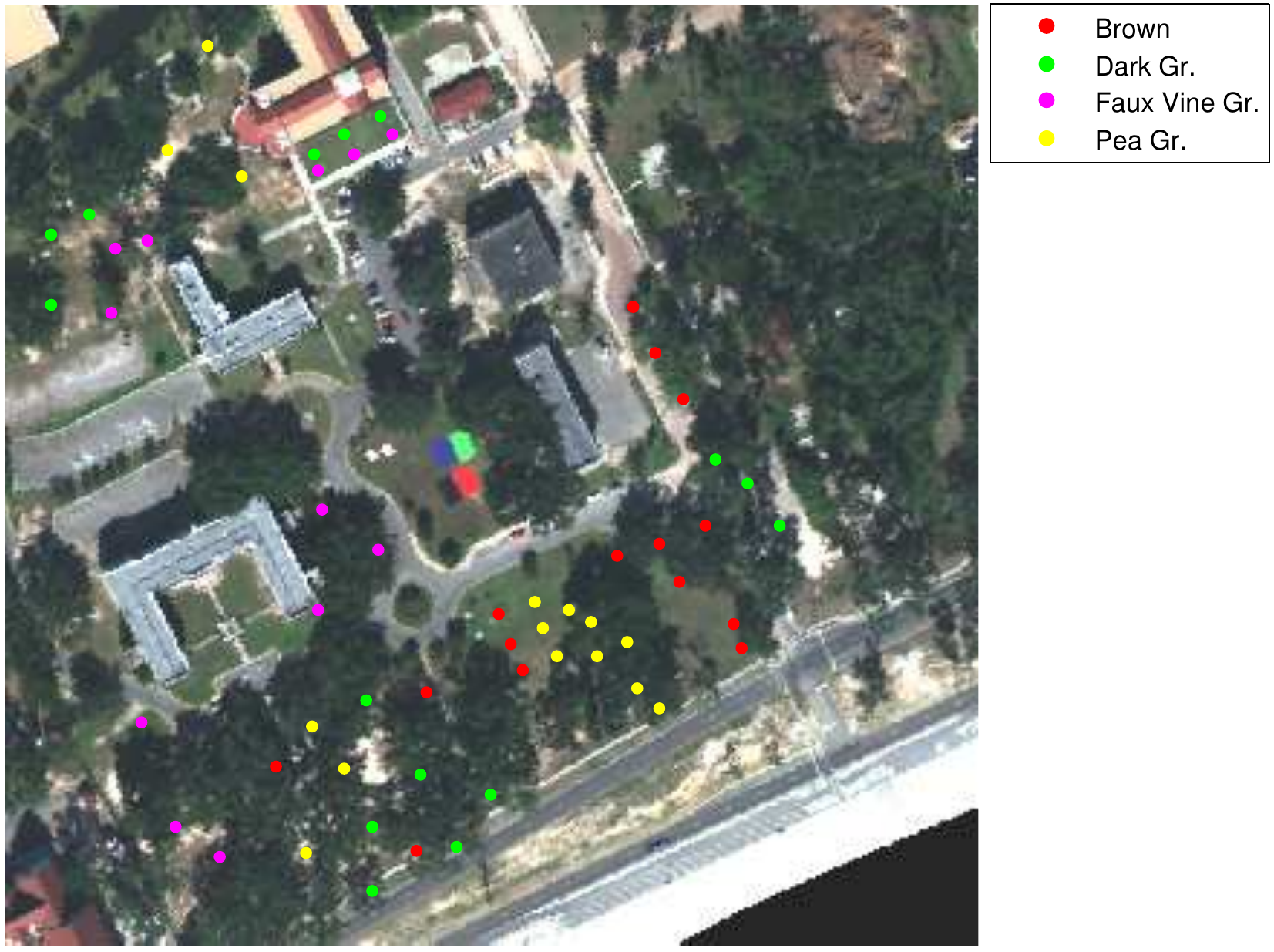}}
	\caption{MUUFL Gulfport data set RGB image and the 57 target locations}
	\label{fig:gulfport_rgb}
\end{figure}

For experiments on real hyperspectral target detection data, the MUUFL Gulfport hyperspectral data set collected over the University of Southern Mississippi-Gulfpark Campus was used.  This data set contains $325\times337$ pixels with 72 spectral bands corresponding to wavelengths from $367.7 nm$ to $1043.4 nm$ at a $9.5 - 9.6 nm$ spectral sampling interval with spatial resolution $1$ pixel$/m^2$ \cite{gader:2013}. The first four and last four bands were removed due to sensor noise. Two sets of this data (Gulfport Campus Flight 1 and Gulfport Campus Flight 3) were selected as cross-validated training and testing data for these two data sets have the same altitude and spatial resolution. Throughout the scene, there are 64 man-made targets in which 57 were considered in this experiment which are cloth panels of four different colors: Brown (15 examples), Dark Green (15 examples), Faux Vineyard Green (FVGr) (12 examples) and Pea Green (15 examples).  The spatial location of the targets are shown as scattered points over an RGB image of the scene in Fig. \ref{fig:gulfport_rgb}.  Some of the targets are in the open ground and some are occluded by the live oak trees.  Moreover, the targets also vary in size, for each target type, there are targets that are $0.25 m^2$, $1 m^2$ and $9 m^2$ in area, respectively, resulting a very challenging, highly mixed sub-pixel target detection problem.

\subsubsection{MUUFL Gulfport Hyperspectral Data, Individual Target Type Detection}
For this part of the experiments, each individual target type was treated as a target class, respectively. For example, when ``Brown'' is selected as target class, a $5\times5$ rectangular region corresponding to each of the 15 ground truth locations denoted by GPS was grouped into a positive bag to account for the drift coming from GPS. This size was chosen based on the accuracy of the GPS device used to record the ground truth locations. The remaining area that does not contain a brown target was grouped into a big negative bag. This constructs the detection problem for ``Brown'' target. Similarly, there are 15, 12, 15 positive labeled bags for Dark Green, Faux Vineyard Green and Pea Green, respectively. The parameter settings of MI-HE for this experiment are $T = 1, M = 9, \rho = 0.3, b = 5, \beta=1$ and $\lambda=5\times 10^{-3}$.

\begin{figure}
	\begin{center}
		\subfloat[{Estimated target signatures from flight 3 for Brown and comparison with ground truth}]{   
			\includegraphics[width=10cm]{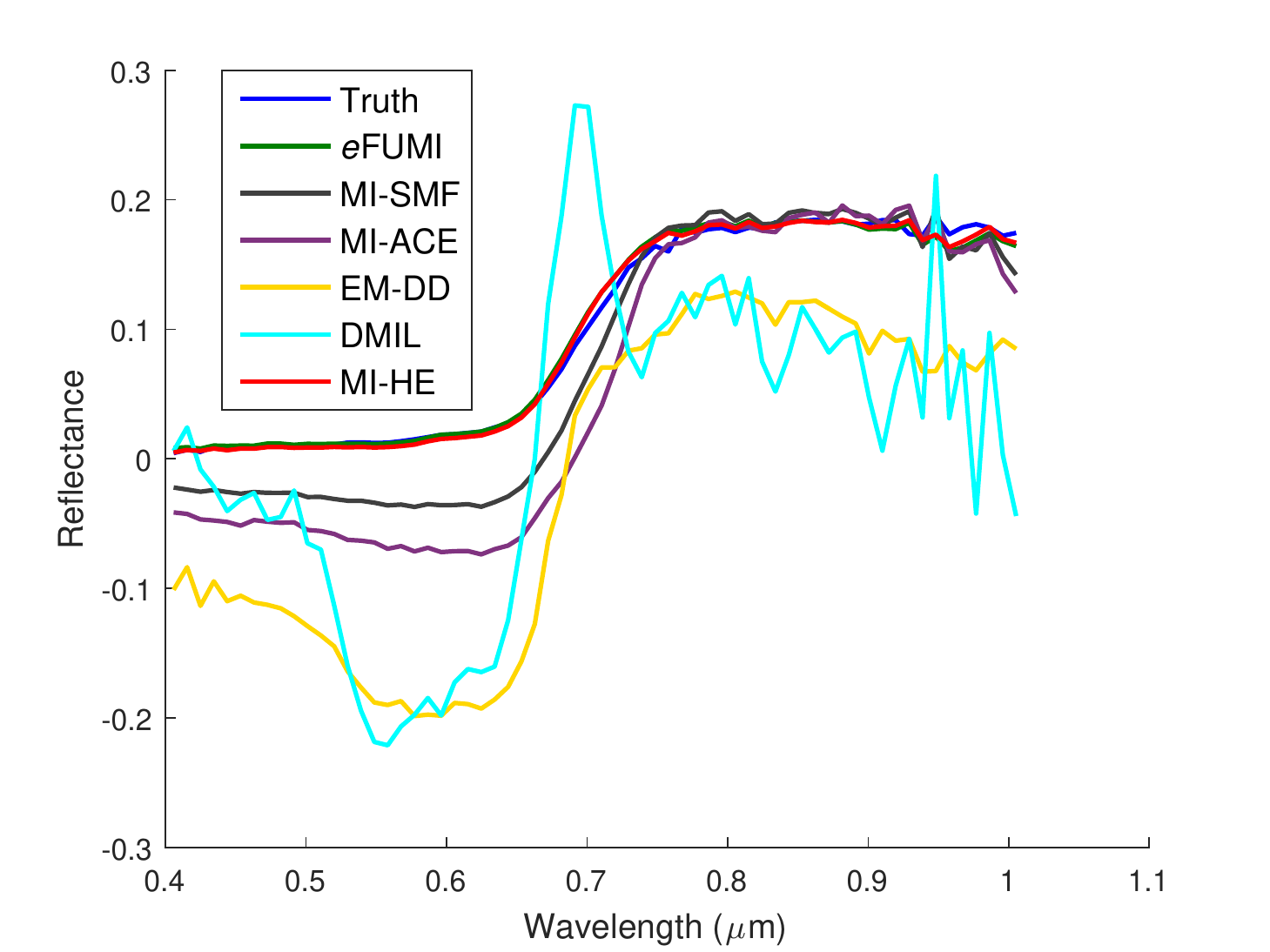} \label{fig:Gulfport_tar_sig_Train3Test1_Brown}}\\
		\subfloat[{ROC curves cross validated on flight 1}]{   
			\includegraphics[width=10cm]{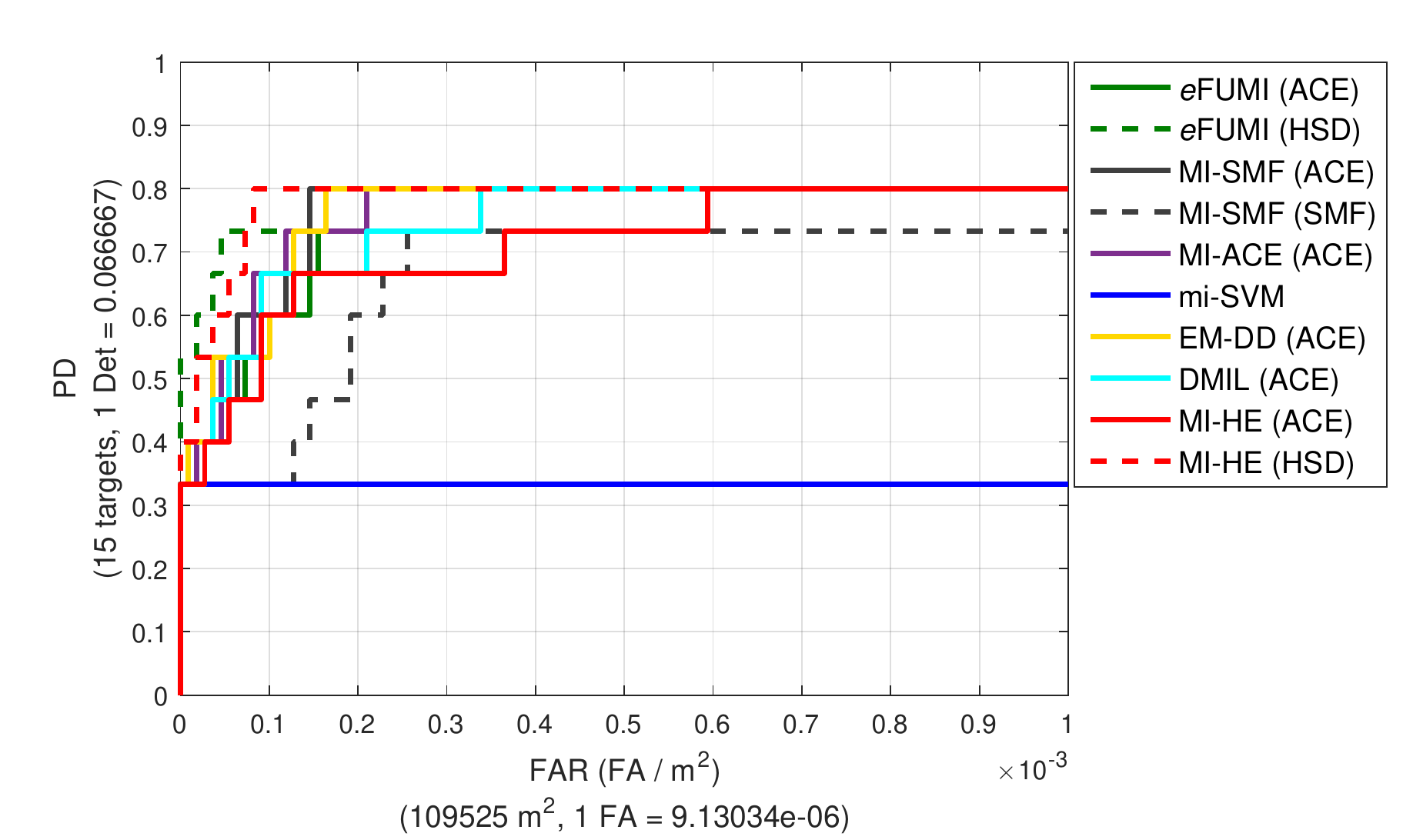} \label{fig:Gulfport_rocs_Train3Test1_Brown}}
		\caption{{MI-HE and comparisons on Gulfport Data Brown, training flight 3 testing flight 1}}\label{fig:MIHE_gulfport_train3test1_brown}
	\end{center}
\end{figure}

MI-HE and comparison algorithms were evaluated on this data using the Normalized Area Under the receiver operating characteristic Curve (NAUC) in which the area was normalized out to a false alarm rate (FAR) of $1\times 10^{-3}$ false alarms$/m^2$ \cite{glenn_gulfport:2013}. During detection on the test data, the background mean and covariance were estimated from the negative instances of the training data. The results reported are the median results over five runs of the algorithm on the same data.

Fig. \ref{fig:Gulfport_tar_sig_Train3Test1_Brown} shows the estimated target concept by proposed MI-HE and comparisons for Brown target type training on flight 3. We can see that the proposed MI-HE is able to recover the target concept quite close to groundtruth spectra manually selected from the scene. Fig. \ref{fig:Gulfport_rocs_Train3Test1_Brown} shows the detection ROCs given target spectra estimated on flight 3 and cross validated on flight 1. Tab. \ref{tab:gulfport_individual} shows the NAUCs for MI-HE and comparison algorithms cross validated on all four types of target, where it can be seen the proposed MI-HE generally outperforms the comparisons for most of the target types {and achieves close to the performance of using ground truth target signatures. Since MI-HE is a discriminative target concept learning framework that aims to distinguish one target instance from each positively labeled bag, MI-HE had a lower performance for the pea green target because of the relatively large occlusion of those targets causing difficulty in distinguishing pea green signature from each of the positive bag.}

\begin{table*}[!htb]
	\begin{scriptsize}
		\begin{center}
			\vspace{-4mm}\caption{{Area under the ROC curves for MI-HE and comparison algorithms on Gulfport Data with Individual Target Type.  Best results shown in bold, second best results underlined, and ground truth shown with an asterisk. }}  \label{tab:gulfport_individual}
			\begin{tabular}{|c|c|c|c|c|c|c|c|c|}
				\hline
				\multirow{2}{*}{Alg.} &  \multicolumn{4}{c|} {Train on Flight 1; Test on Flight 3 } &  \multicolumn{4}{c|} {Train on Flight 3; Test on Flight 1 } \\
				\cline{2-9}&           Brown     &    Dark Gr.           &              Faux Vine Gr.  &   Pea Gr.            &                Brown         &    Dark Gr.  & Faux Vine Gr.     &   Pea Gr.  \\\hline\hline
				\textbf{MI-HE (HSD)}      &   \textbf{0.499}    &    \textbf{0.453}     &        \textbf{0.655}       &   0.267              &         \textbf{0.781}       &   \textbf{0.532}     &       \textbf{0.655}       &    0.350              \\\hline
				\textbf{MI-HE (ACE)}     &           0.433     &            0.379      &    			  0.104        &   0.267              &                 0.710        &           0.360     &          	0.111       &    0.266     \\\hline
				\textbf{$e$FUMI \cite{Zare:2015fumi} (ACE)}  &           0.423     &            0.377      &     \underline{0.654}       &   0.267              &                 0.754        &  	     0.491     &       0.605       &    \underline{0.393}    \\\hline
				\textbf{$e$FUMI \cite{Zare:2015fumi} (HSD)}   &           0.444     & \underline{0.436}     &                0.653        &   0.267              &                 0.727        &\underline{0.509}     &       0.500       &    0.333     \\\hline
				\textbf{MI-SMF \cite{zare2016miace} (SMF)}   &           0.419     &            0.354      &   			  0.533        &   {0.266}  &      {0.657}       &   		 0.405     &       \underline{0.650}       &    0.384     \\\hline  
				\textbf{MI-SMF \cite{zare2016miace} (ACE)}   &           0.448     &            0.382      &   			  0.579        &   \underline{0.316}  &      \underline{0.760}       &   		 0.501     &       {0.613}       &    0.388     \\\hline  
				\textbf{MI-ACE \cite{zare2016miace} (ACE)}   &\underline{0.474}    &            0.390      &			      0.485        &   \textbf{0.333}     &      \underline{0.760}       &    		 0.483     &       0.593       &    0.380     \\\hline
				\textbf{mi-svm} \cite{andrews2002support}       &           0.206     &            0.195      &   			  0.412        &   0.265              &                 0.333        &  		 0.319     &       0.245       &    0.274      \\\hline
				\textbf{EM-DD \cite{Zhang:2002}  (ACE)}    &           0.411     &            0.381      &    		      0.486        &   0.279              &      \underline{0.760}       &   		 0.503     &       0.541       &    \textbf{0.416}     \\\hline
				\textbf{DMIL \cite{shrivastava2015gen,shrivastava2014dictionary}  (ACE)}     &           0.419     &            0.383      &    			  0.191        &   0.009              &                 0.743        &   		 0.310     &       0.081       &    0.083     \\\hline
				{\textbf{Ground Truth  (ACE)}}     &           {0.528*}     &            {0.429*}      &    			  {0.656* }       &   {0.267* }            &                {0.778*}        &   		{0.521*}     &       {0.663* }      &  { 0.399*}     \\\hline
			\end{tabular}
		\end{center}
	\end{scriptsize}
\end{table*}

\subsubsection{MUUFL Gulfport Hyperspectral Data, All Four Target Types Detection}

\begin{figure}[tbh!]
	\begin{center}
		\subfloat[Train on flight 1, detect on flight 3]{   
			\includegraphics[width=10cm]{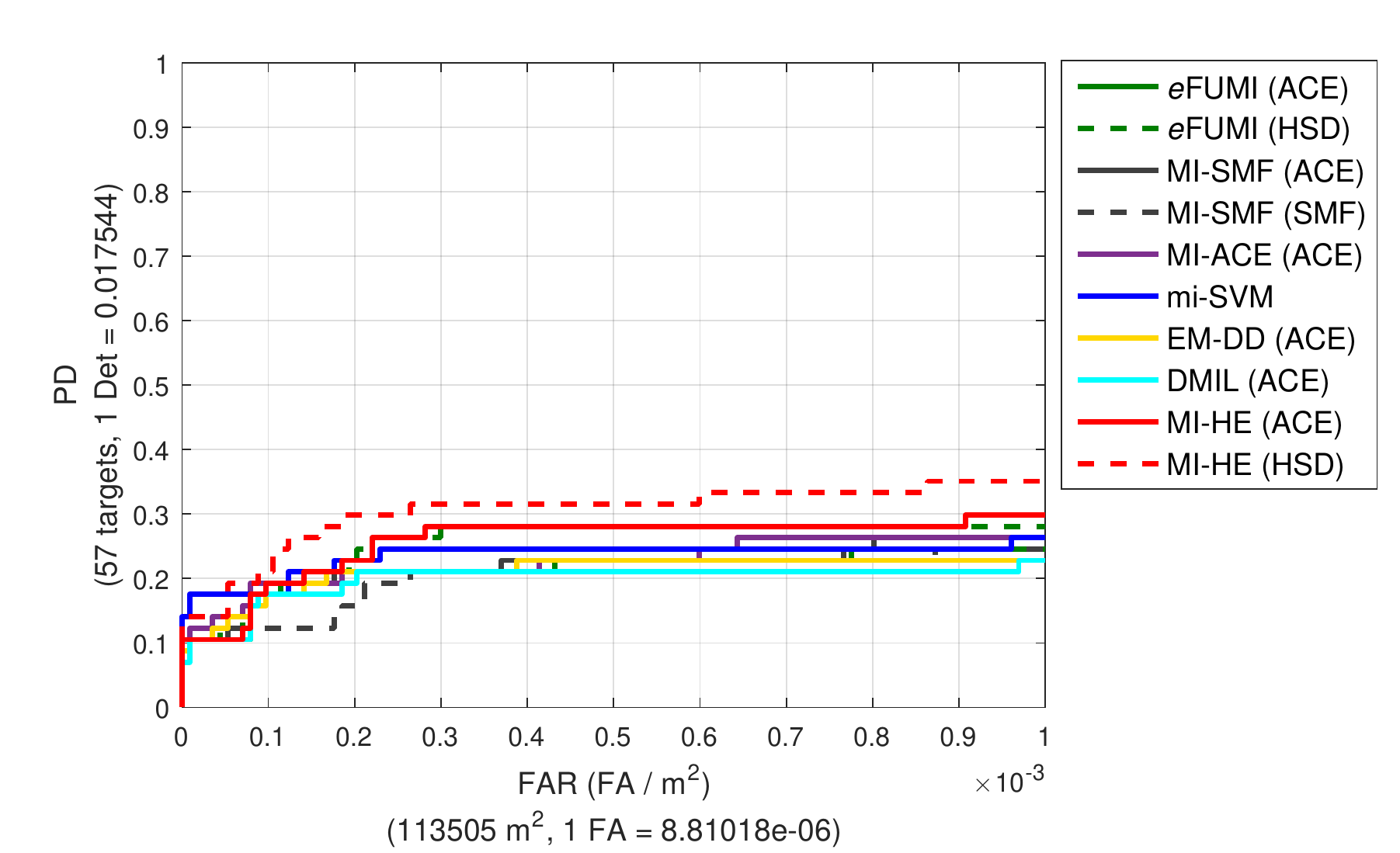} \label{fig:Gulfport_rocs_Train1Test3_All4Types}}\\
		\subfloat[Train on flight 3, detect on flight 1]{   
			\includegraphics[width=10cm]{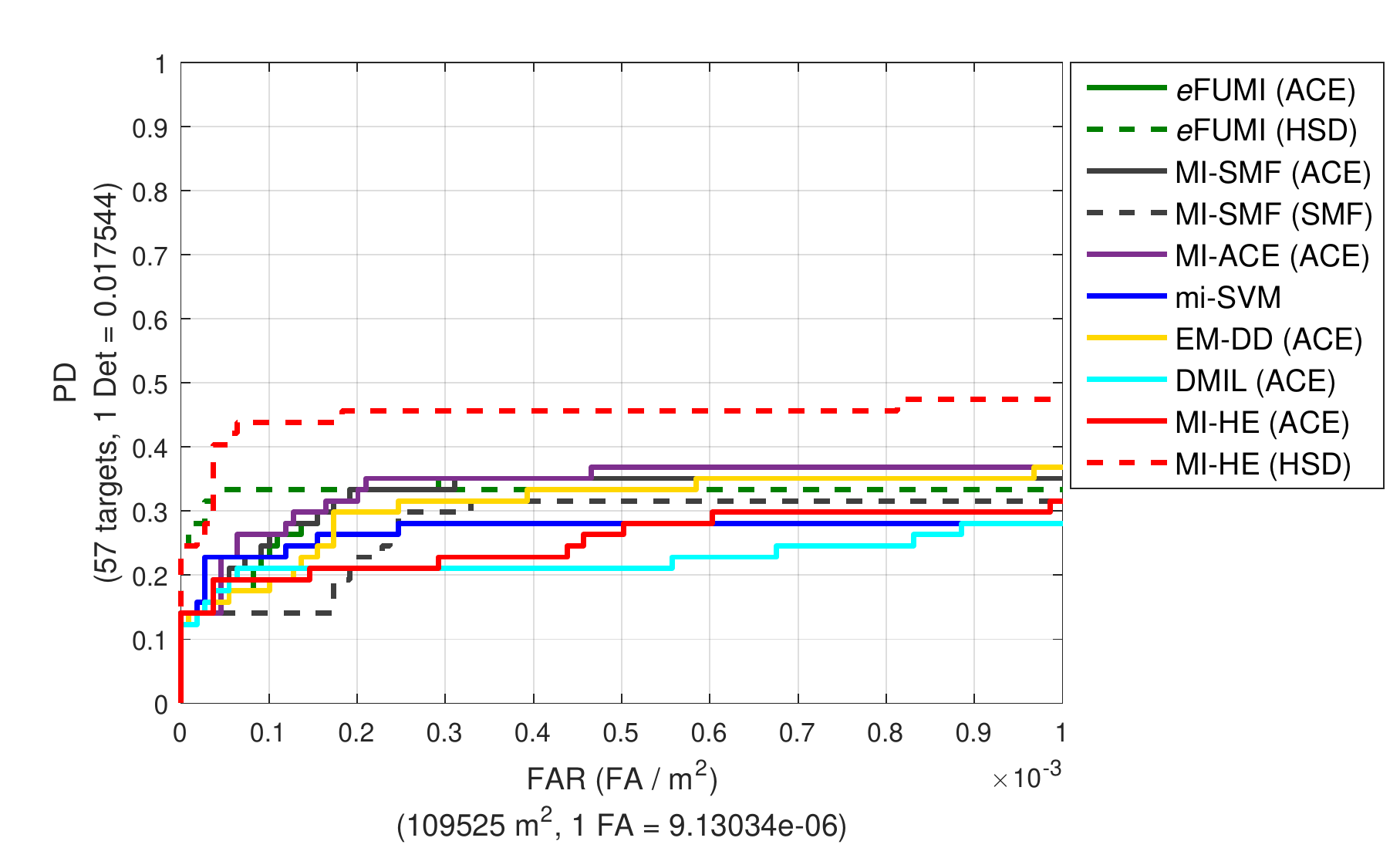} \label{fig:Gulfport_rocs_Train3Test1_All4Types}}
		\caption{ROCs of MI-HE and comparisons on Gulfport Data all types detection}\label{fig:MIHE_gulfport_rocs_all4types}
	\end{center}
\end{figure}

For training and detection for the four target types together, the positive bags were generated by grouping each of the $5\times5$ regions denoted by the groundtruth that it contains any of the four types of target. Thus, for each flight there are 57 target points and 57 positive bags were generated. The remaining area that does not contain any target was grouped into a big negative bag.  The parameter settings of MI-HE for this experiment are $T = 9, M = 11, \rho = 0.3, b = 5, \beta=1$ and $\lambda=5\times 10^{-3}$.

\begin{table*}[!htb]
	\begin{scriptsize}
		\begin{center}
			\vspace{-4mm}\caption{{Area under the ROC curves for MI-HE and comparison algorithms on Gulfport Data with All Four Target Types.  Best results shown in bold, second best results underlined, and ground truth shown with an asterisk. }}  \label{tab:gulfprot_all4type}
			\begin{tabular}{|c|c|c|c|c|c|}
				\hline
				{Alg.} &  {Tr. Fl. 1; Te. Fl. 3 } &  {Tr. Fl. 3; Te. Fl. 1 } &{Alg.} &  {Tr. Fl. 1; Te. Fl. 3} &  {Tr. Fl. 3; Te. Fl. 1} \\\hline
				\textbf{MI-HE (HSD)}      &      \textbf{0.304}         &   \textbf{0.449}           &    \textbf{MI-SMF \cite{zare2016miace}  (ACE)}    &   0.219       &    0.327    \\\hline
				\textbf{MI-HE (ACE)}     &     \underline{0.257}          &    {0.254}               &     \textbf{MI-SMF \cite{zare2016miace}  (SMF)}   &   0.198       &    0.277        \\\hline
				\textbf{$e$FUMI \cite{Zare:2015fumi}  (ACE)}  &      0.214                     &   \underline{0.325}      &       \textbf{mi-SVM} \cite{andrews2002support}      &   0.235       &    0.269        \\\hline
				\textbf{$e$FUMI \cite{Zare:2015fumi}  (HSD)}   &      0.256                      &   0.331                &    \textbf{EM-DD \cite{Zhang:2002} (ACE)}     &   0.211       &    0.310        \\\hline
				\textbf{MI-ACE \cite{zare2016miace}  (ACE)}    &      0.226                       &   0.340                 &   \textbf{DMIL \cite{shrivastava2015gen,shrivastava2014dictionary} (ACE)}       &   0.198       &    0.225        \\\hline  
				{\textbf{Ground Truth  (ACE)}}    &    { 0.330*}                     &   {0.490*}              &        &          &            \\\hline  
			\end{tabular}
		\end{center}
	\end{scriptsize}
\end{table*}

Fig. \ref{fig:Gulfport_rocs_Train1Test3_All4Types}  and \ref{fig:Gulfport_rocs_Train3Test1_All4Types} shows the detection ROCs given target spectra estimated one flight and cross validated on another flight, which show that the detection statistic by proposed MI-HE using HSD is significantly better than the comparison algorithms. Tab. \ref{tab:gulfprot_all4type} summarizes the NAUCs as a quantitative comparison.

\subsection{Tree Species Classification from NEON Data}

{In this section, we applied MI-HE to a multiple instance tree species classification problem using a subset of National Ecological Observatory Network (NEON) hyperspectral data \cite{NEON_data} collected at the Ordway-Swisher Biological Station (OSBS) in north-central Florida, United States. This data contains $1020\times1631$ pixels with $428$ bands corresponding to wavelengths from $380$ nm to $2510$ nm at a $5$ nm spectral sampling interval. The spatial resolution and collection altitude are $1$ pixel/$m^2$ and 1000 meters, respectively. }

		\begin{figure}
	\begin{center}
		\subfloat[Full view]{   
			\includegraphics[width=13.5cm]{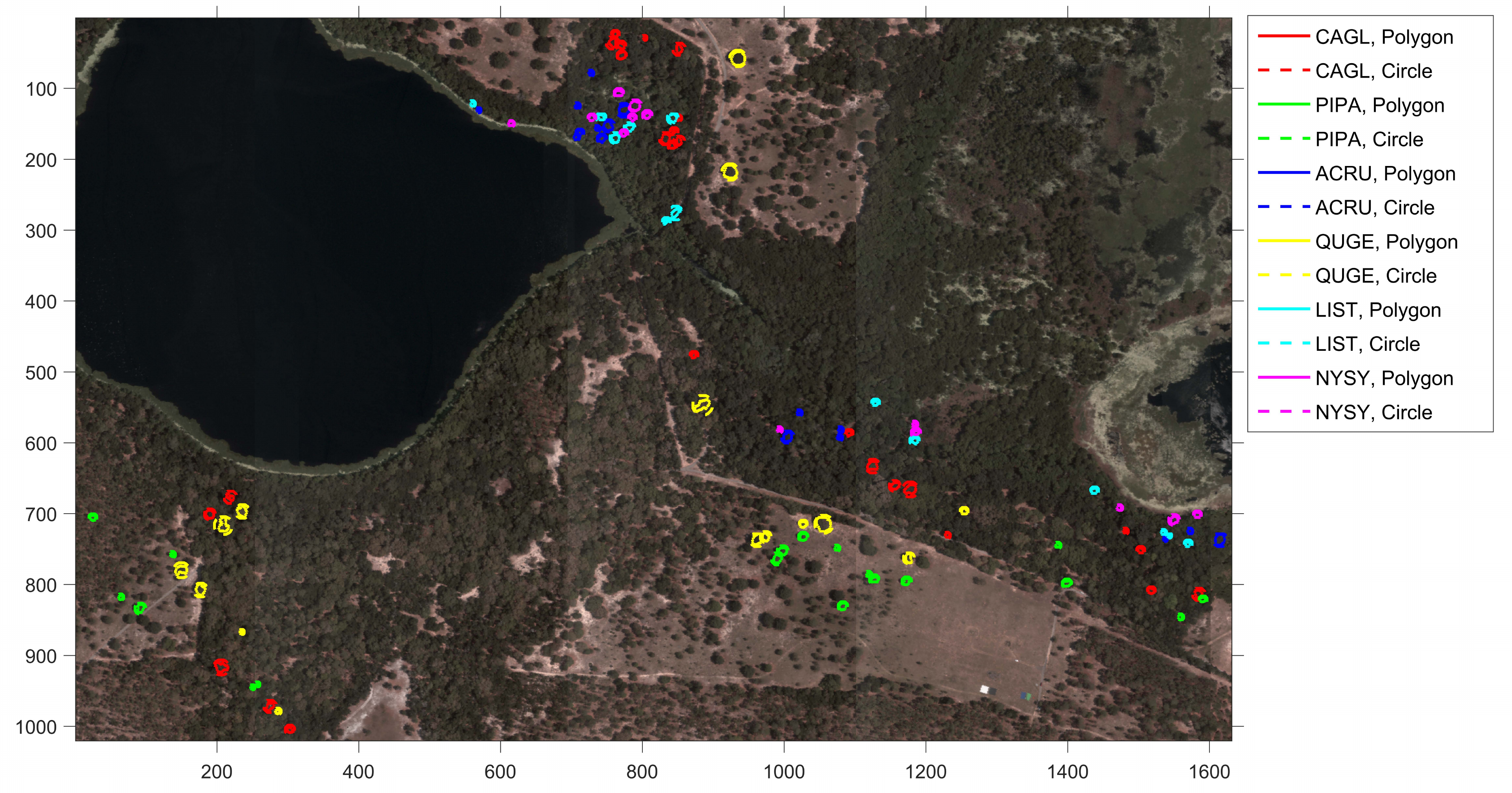} \label{fig:NEON_tree}}\\
		\subfloat[Zoomed view]{   
			\includegraphics[width=13.5cm]{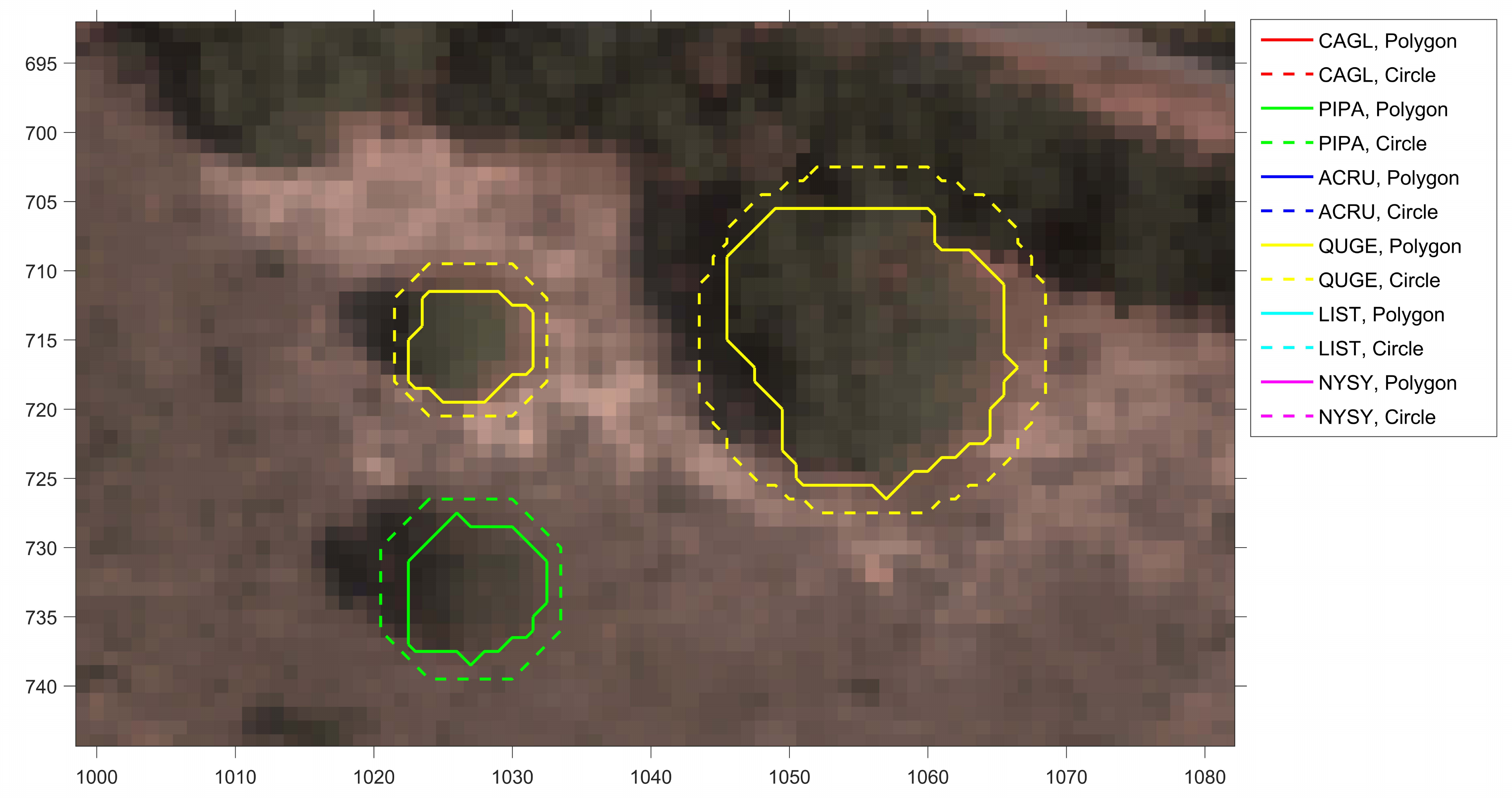} \label{fig:NEON_tree_zoomed}}
		\caption{RGB image of NEON OSBS with tree polygons}\label{fig:NEON_tree_img}
	\end{center}
\end{figure}

{Fig. \ref{fig:NEON_tree} shows the full ground view (RGB) of the hyperspectral data where there are six types of trees named CAGL (\textit{Carya glabra}), PIPA (\textit{Pinus palustris}), ACRU (\textit{Acer rubrum}), QUGE (\textit{Quercus geminata}), LIST (\textit{Liquidambar styraciflua}) and NYSY (\textit{Nyssa sylvatica}), denoted by polygons with different colors. The number of labeled trees for each type is: 26, 18, 16, 15, 13 and 13, respectively. Fig. \ref{fig:NEON_tree_zoomed} shows a zoomed in view of Fig. \ref{fig:NEON_tree}, where the solid line denotes the polygon boundaries of each tree’s crown. The polygon boundaries were mapped in the field on the image loaded on a GPS-connected tablet. These polygon contours create an accurately labeled training data set for tree species classification. Several existing methods \cite{graves2016tree, nia2015impact} train a SVM from crown polygons for tree species classification. However, obtaining accurate tree canopy polygons is time consuming, requiring a lot of effort in the field. Furthermore, the canopy labels are inherently inaccurate, \eg, the accuracy of GPS can drift several meters; the boundary of a tree crown can be ambiguous. }

{So here we model this task as a MIL problem and more general, inaccurate labels were generated by drawing a circle including each polygon shown as dashed line in Fig. \ref{fig:NEON_tree_zoomed}. The goal is to show the proposed MI-HE is able to do tree species classification well given the inaccurate circle labels. {One could imagine these ``circle labels'' being generated using the GPS coordinate of the center of a tree and an approximate tree radius, which would be much easier to collect than the precise polygon labels. The experiment was conducted by training on 70\% of the canopies, randomly selected, and testing on the remaining 30\%.} The testing step was conducted by scoring the testing data point by point, where the testing data were per-pixel labeled according to the polygon data. Hierarchical dimensionality  reduction was applied to reduce the dimensionality of the data to 124 \cite{zare_gulfport:2018}. Both the circle data and polygon data were used as training and compared with SVM. The parameter settings of MI-HE for this experiment are $T = 1, M = 8, \rho = 0.5, b = 5, \beta=1$ and $\lambda=1\times 10^{-3}$. The experiments were repeated for five times and the median performance (AUC) was shown.}

\begin{figure}
	\begin{center}
		\subfloat[CAGL]{   
			\includegraphics[width=7cm]{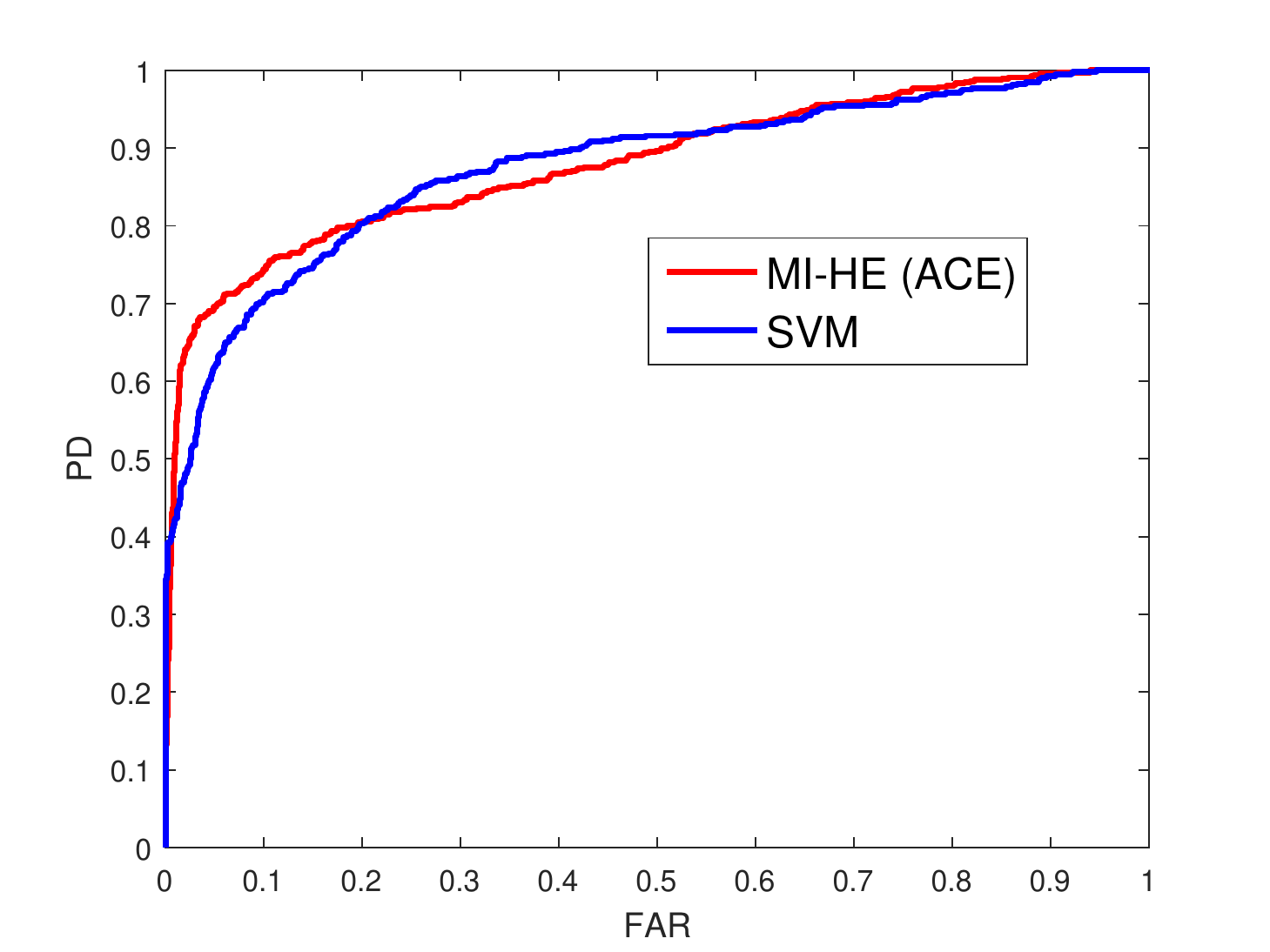} \label{fig:poly_CAGL}}
		\subfloat[PIPA]{   
			\includegraphics[width=7cm]{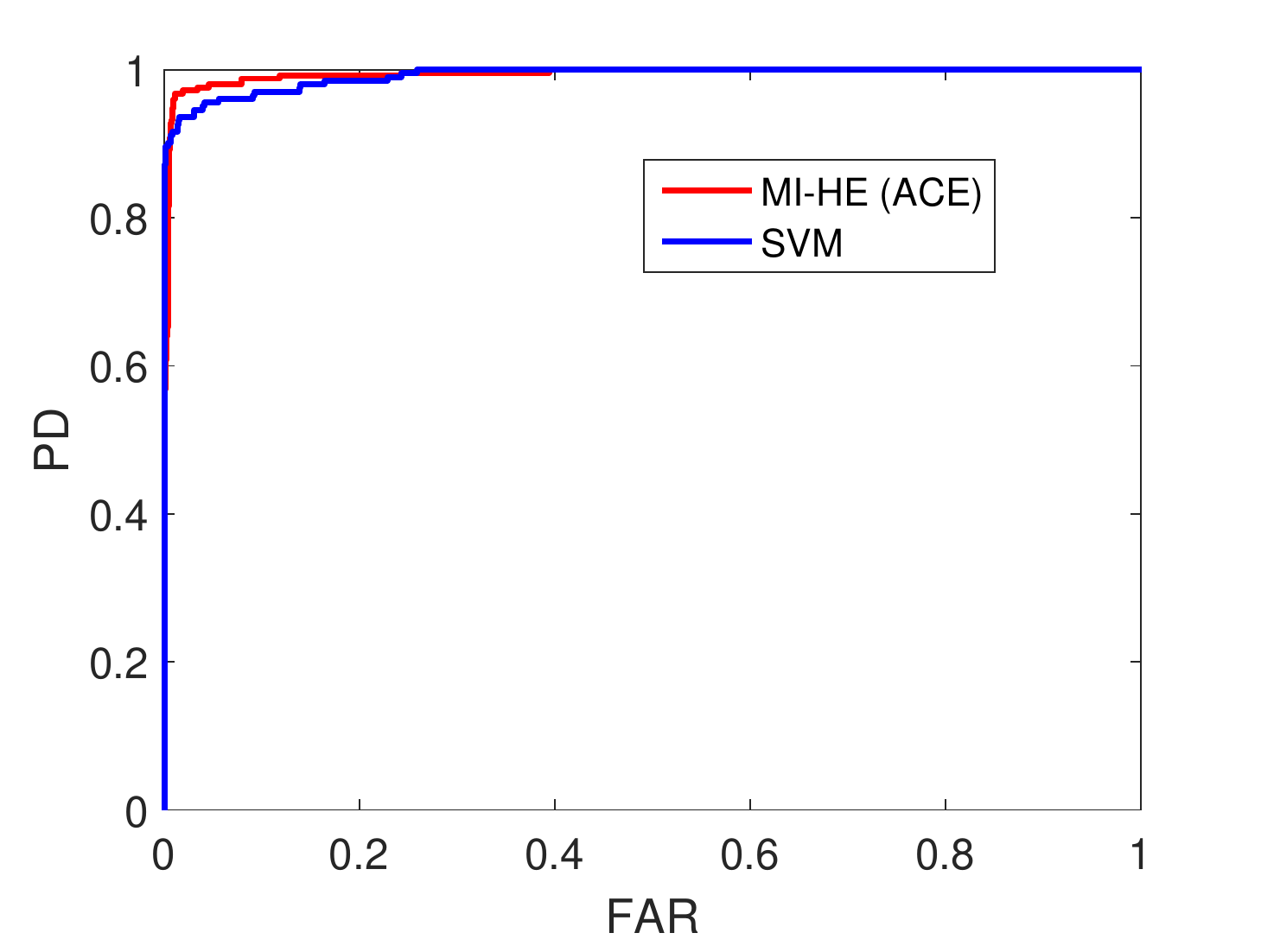} \label{fig:poly_PIPA}}\\
		\subfloat[ACRU]{   
			\includegraphics[width=7cm]{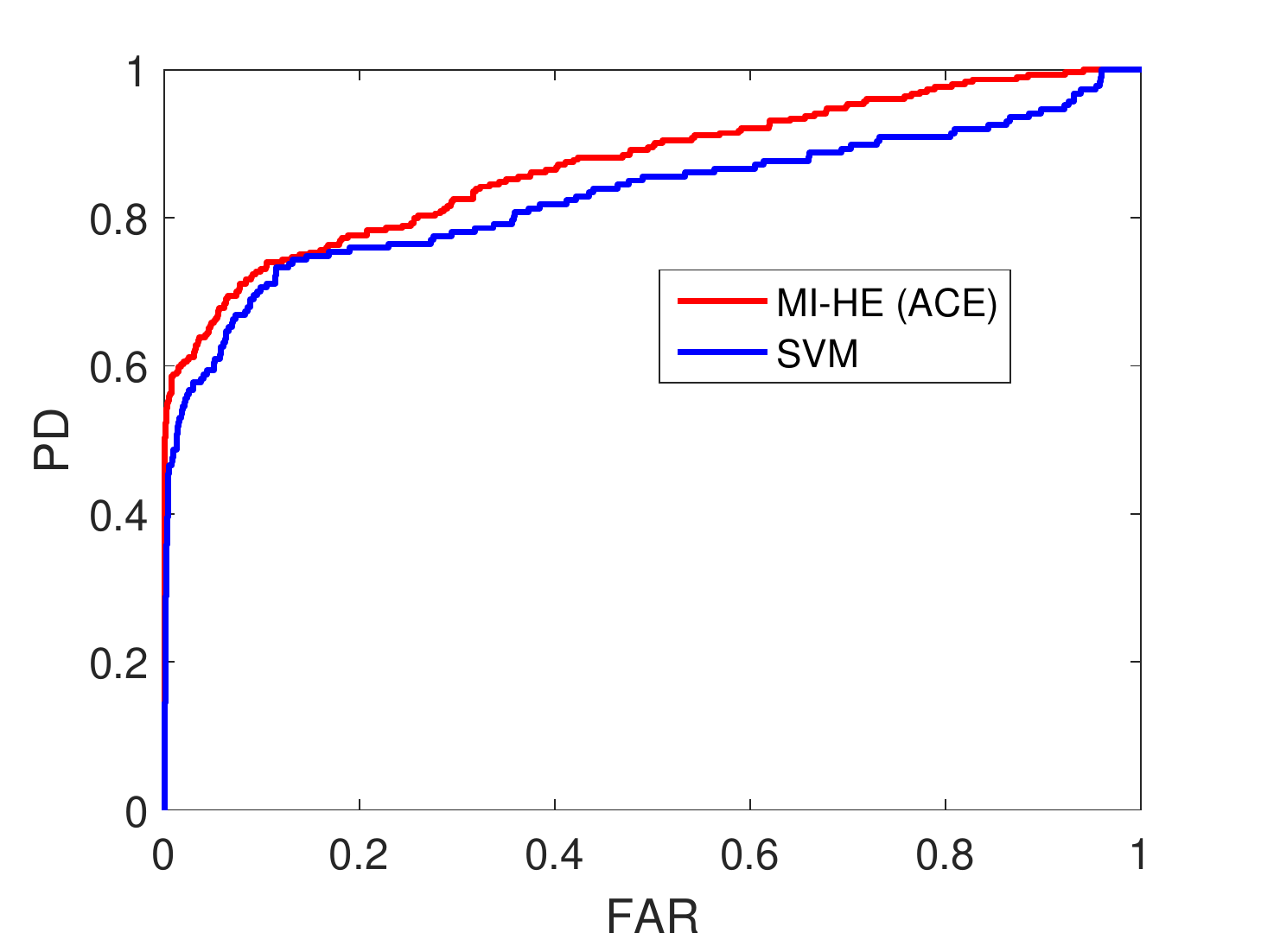} \label{fig:poly_ACRU}}
		\subfloat[QUGE]{   
			\includegraphics[width=7cm]{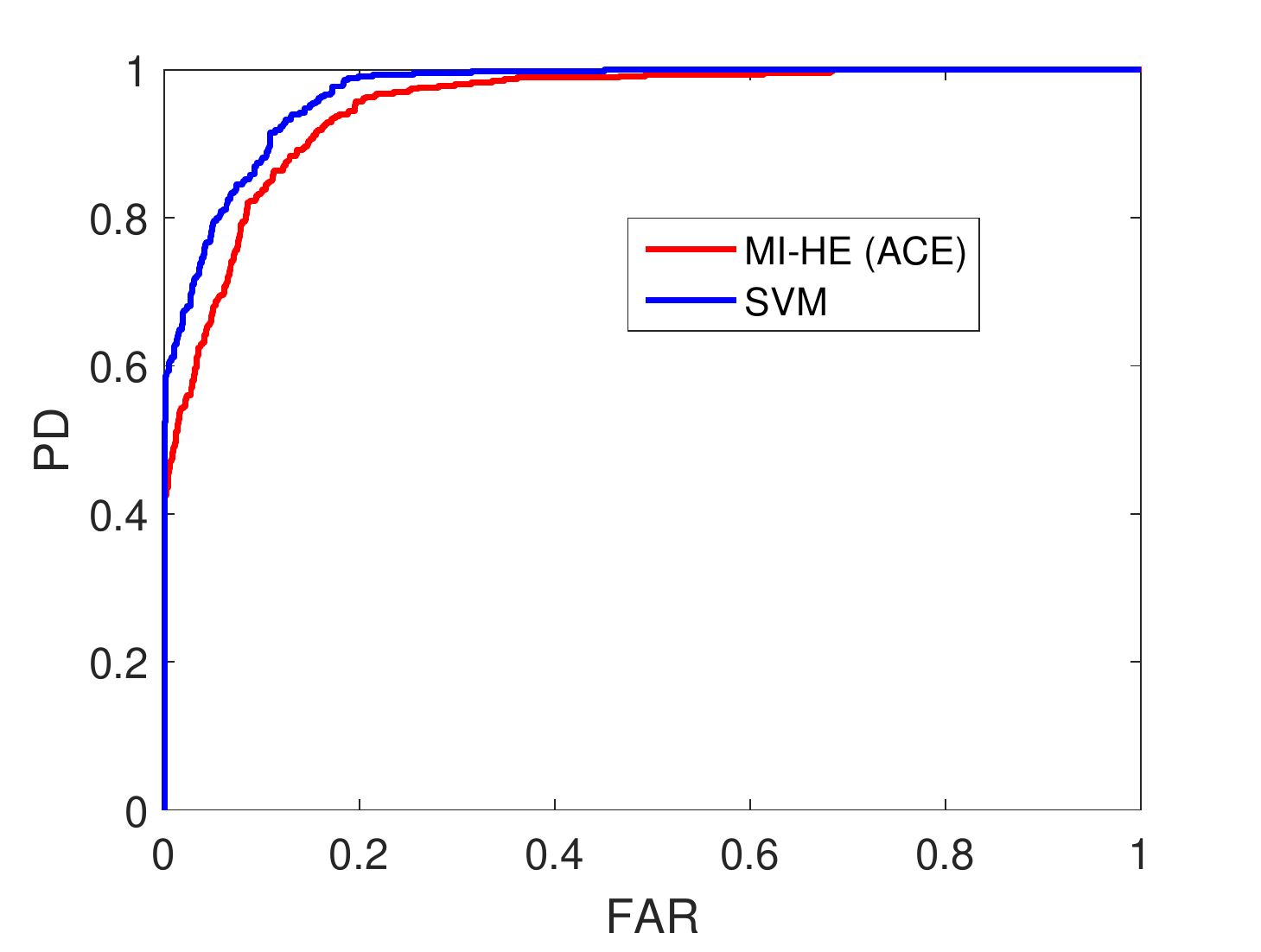} \label{fig:poly_QUGE}}\\
		\subfloat[LIST]{   
			\includegraphics[width=7cm]{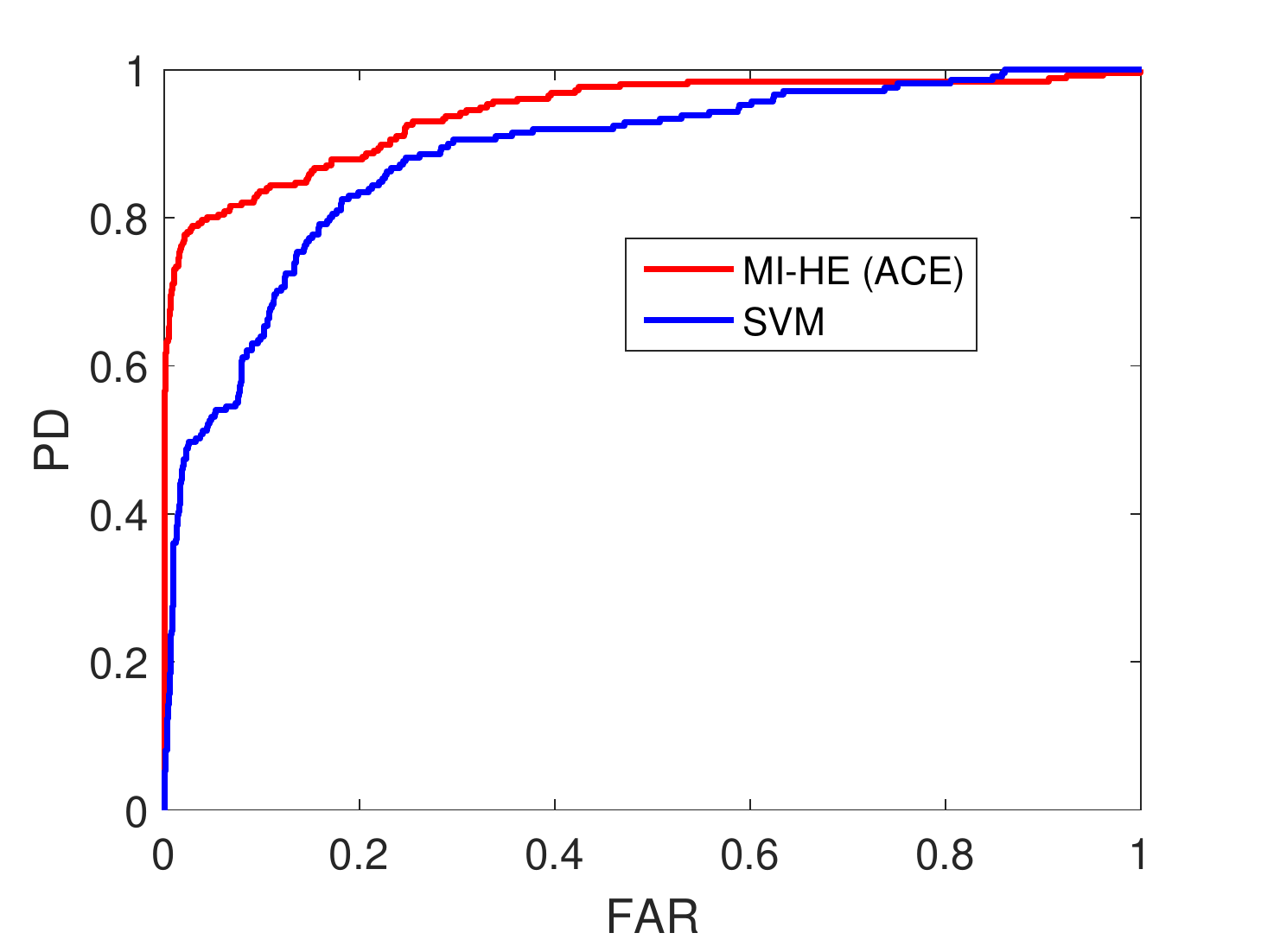} \label{fig:poly_LIST}}
		\subfloat[NYST]{   
			\includegraphics[width=7cm]{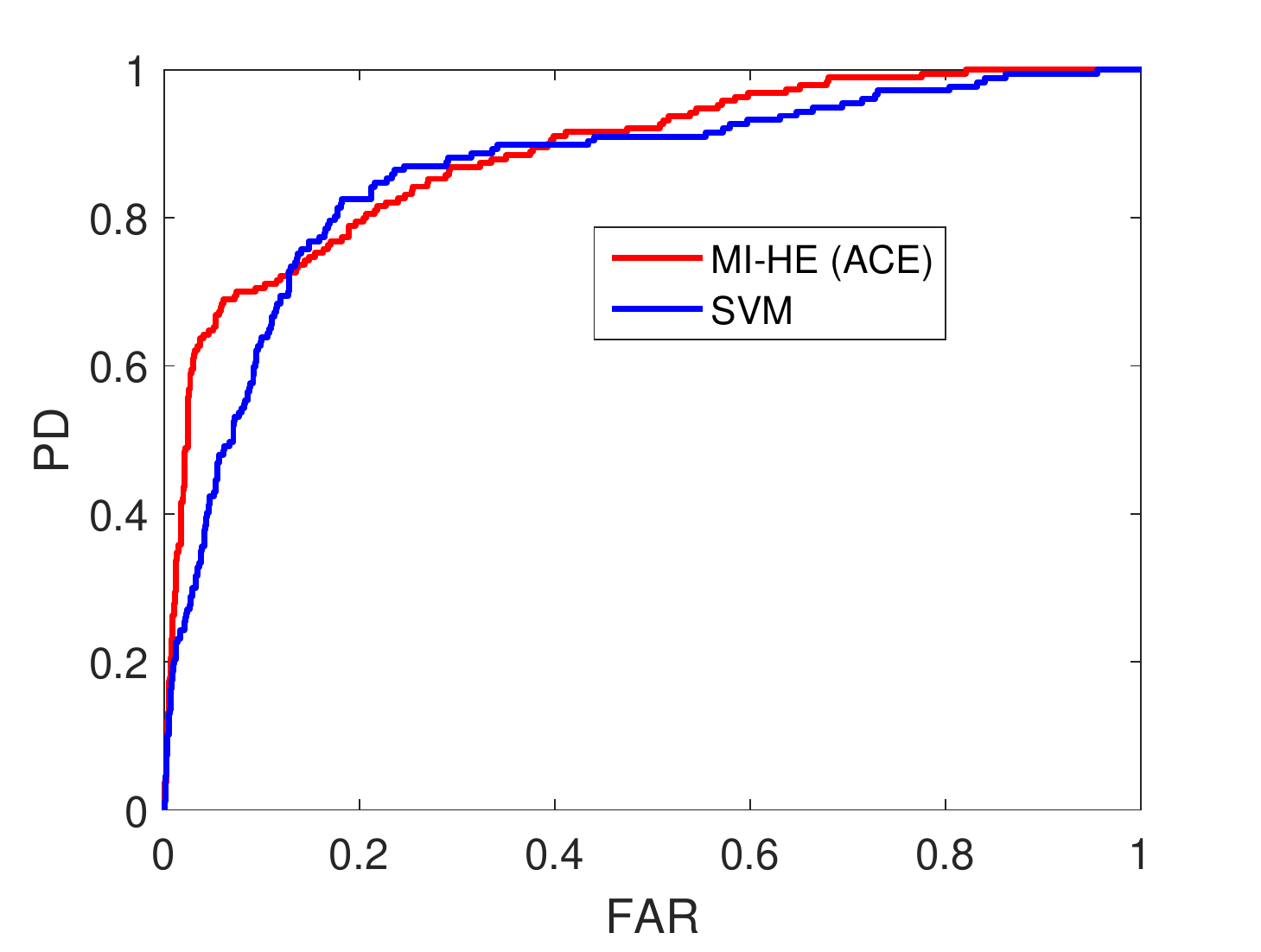} \label{fig:poly_NYSY}}
		\caption{ROC curves of MI-HE and SVM on polygon data }\label{fig:rocs_poly}
	\end{center}
\end{figure}

\begin{figure}
	\begin{center}
		\subfloat[CAGL]{   
			\includegraphics[width=7cm]{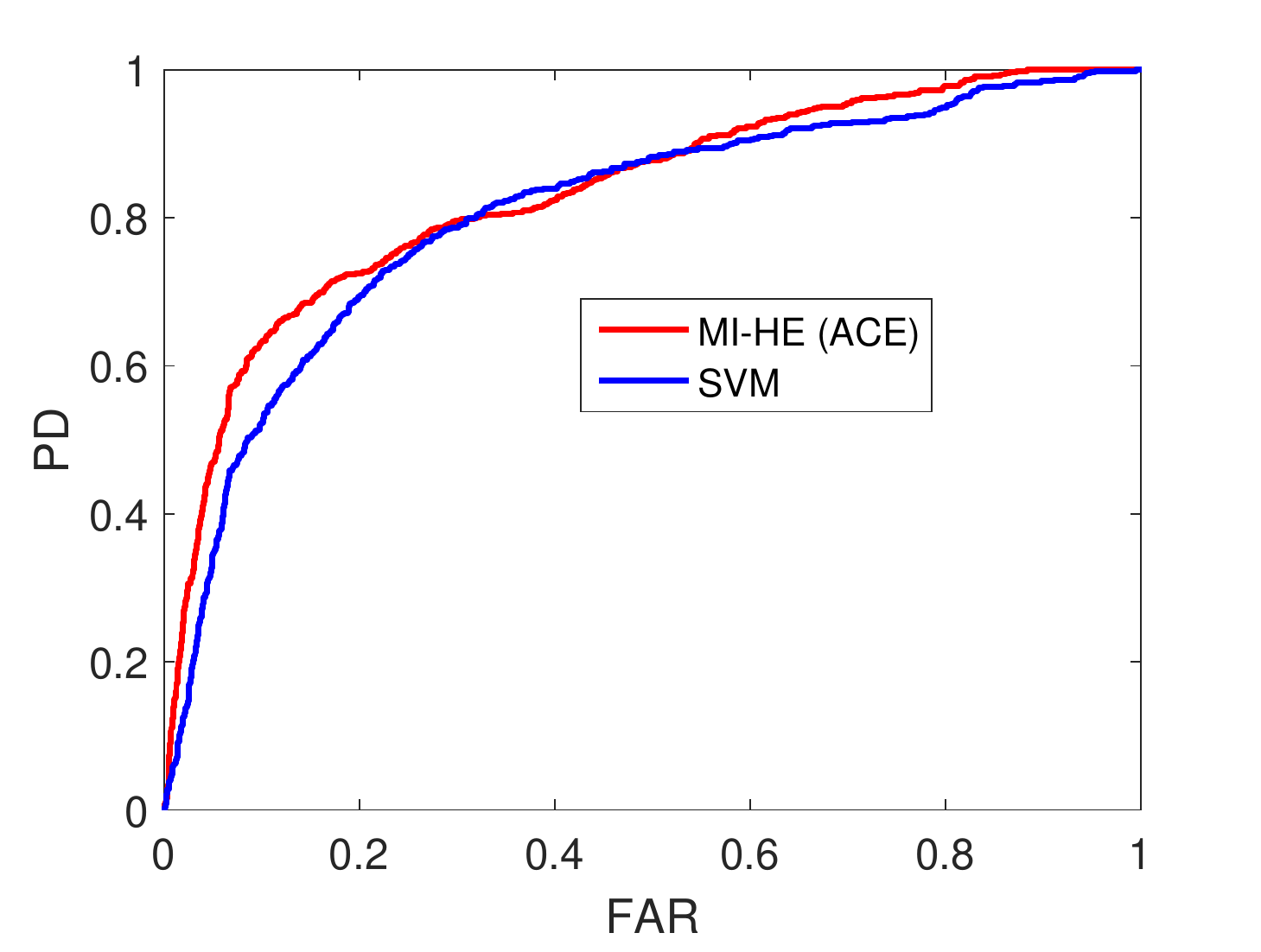} \label{fig:Circle_CAGL}}
		\subfloat[PIPA]{   
			\includegraphics[width=7cm]{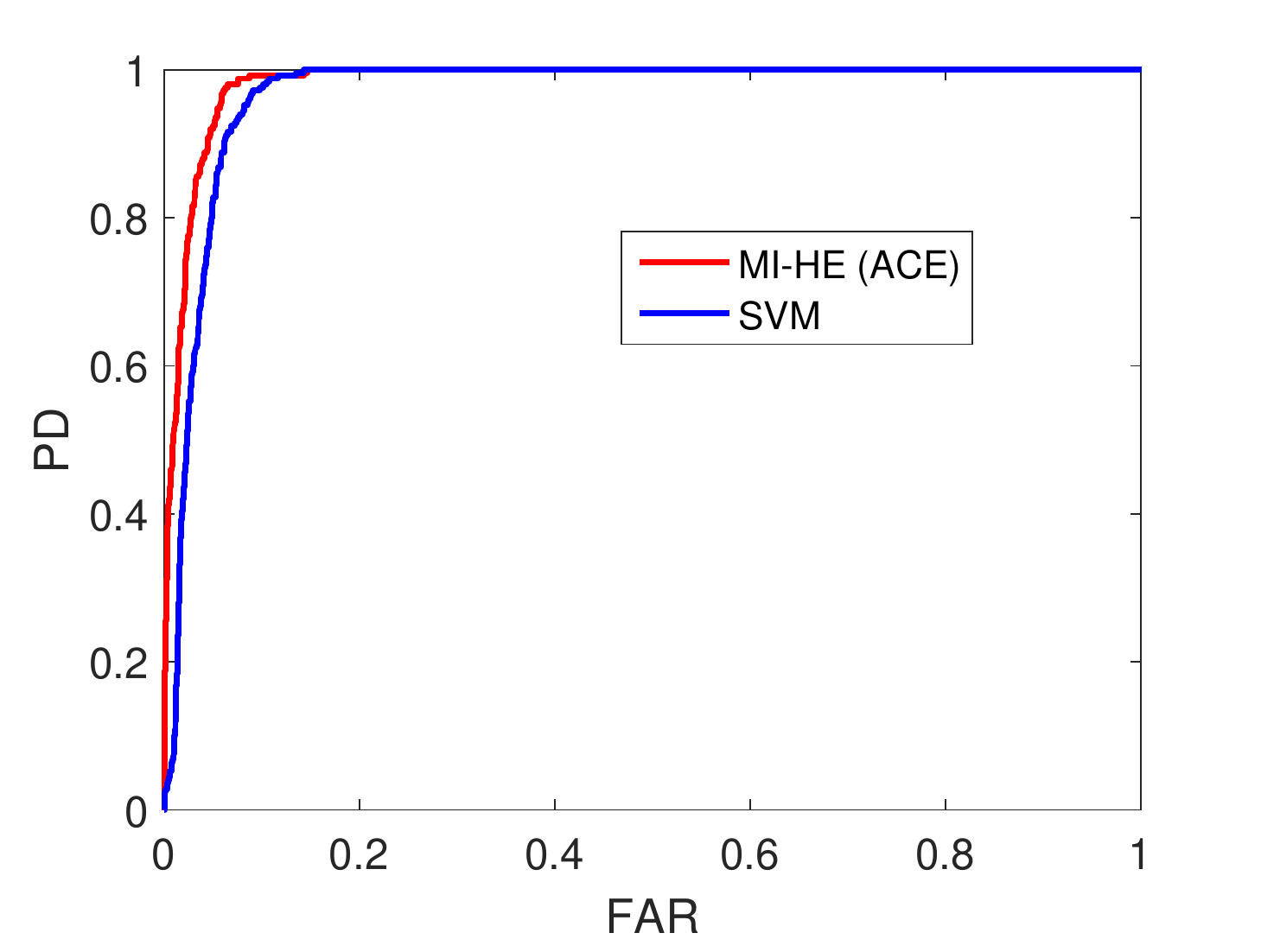} \label{fig:Circle_PIPA}}\\
		\subfloat[ACRU]{   
			\includegraphics[width=7cm]{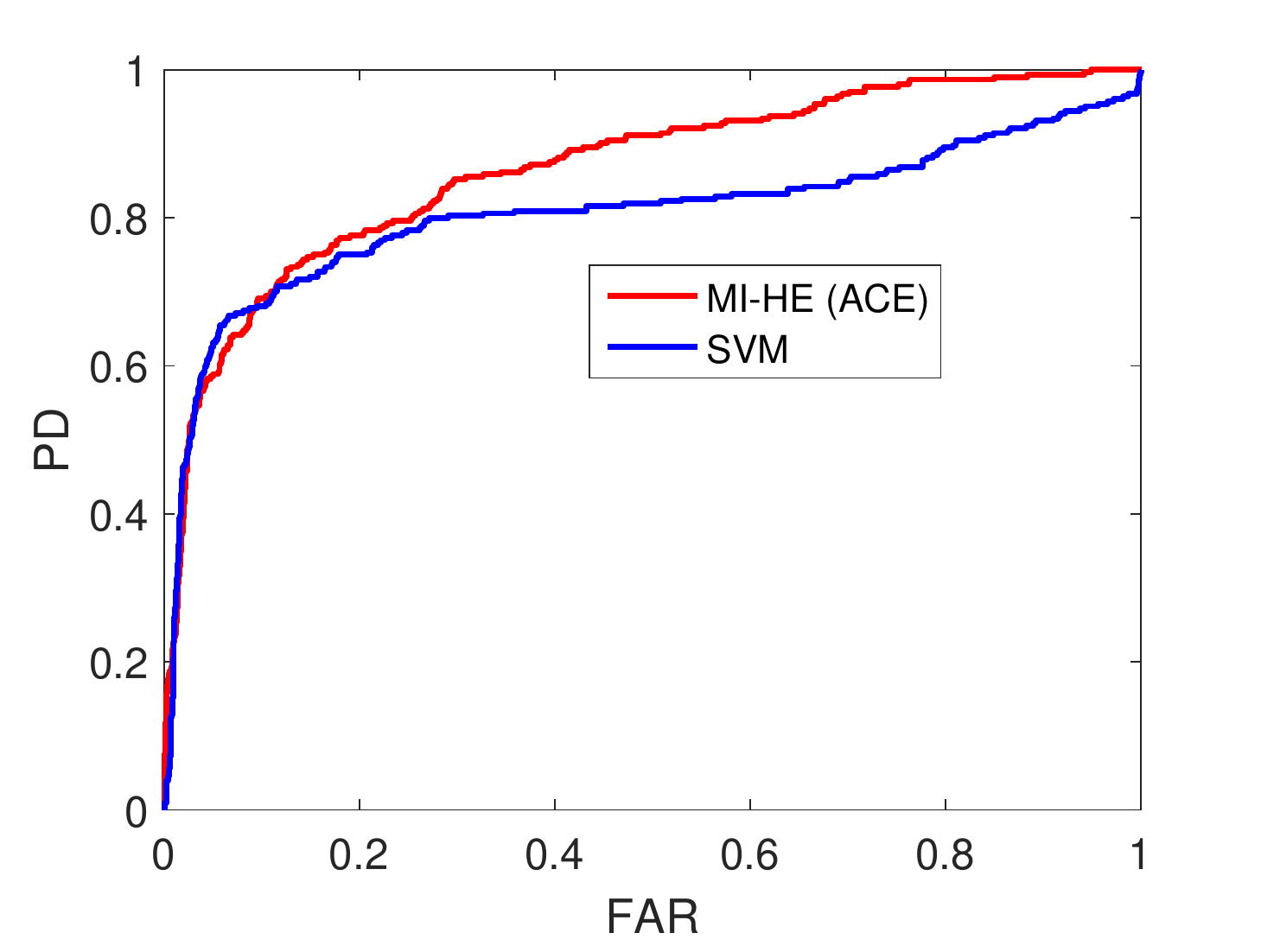} \label{fig:Circle_ACRU}}
		\subfloat[QUGE]{   
			\includegraphics[width=7cm]{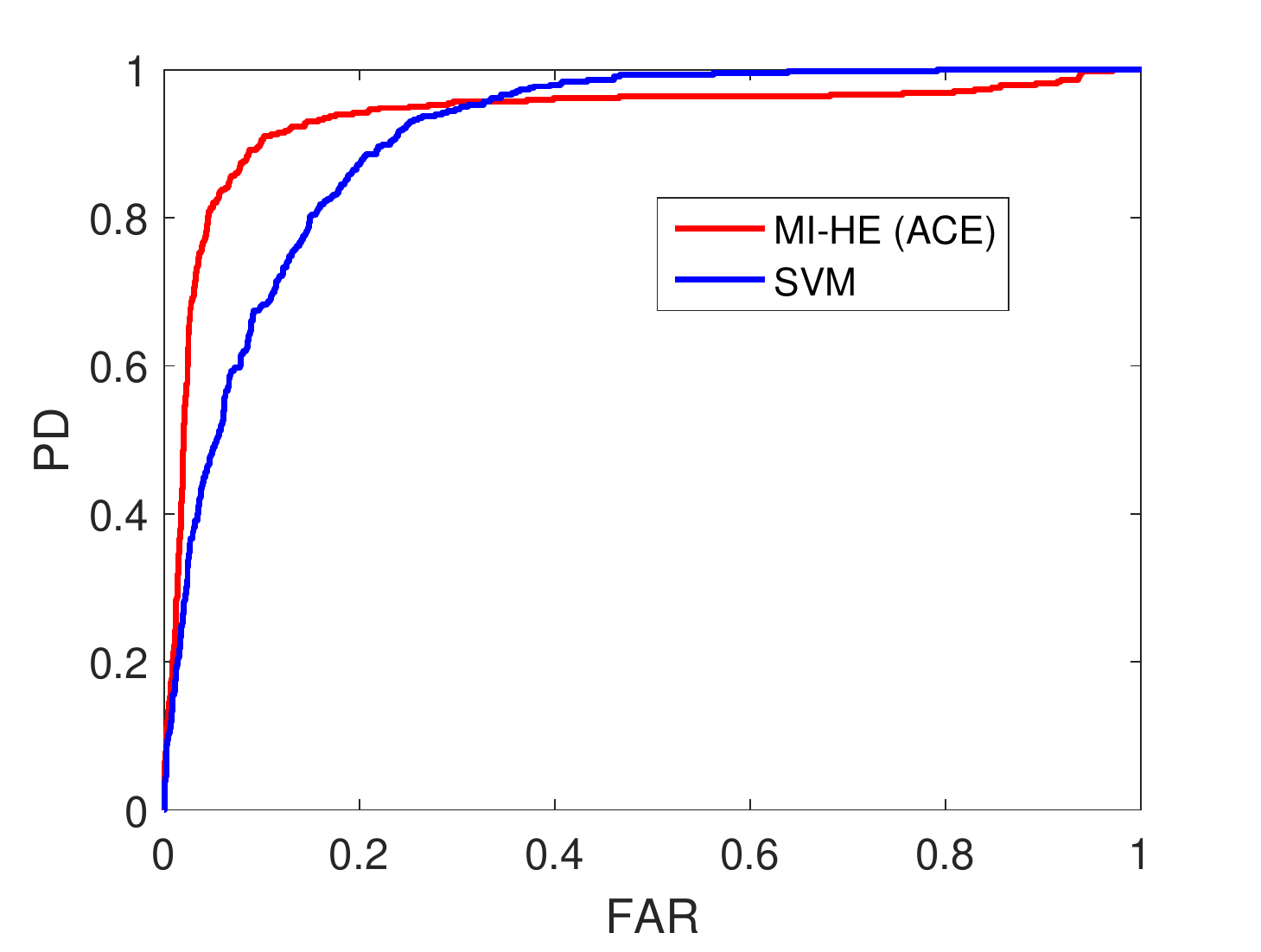} \label{fig:Circle_QUGE}}\\
		\subfloat[LIST]{   
			\includegraphics[width=7cm]{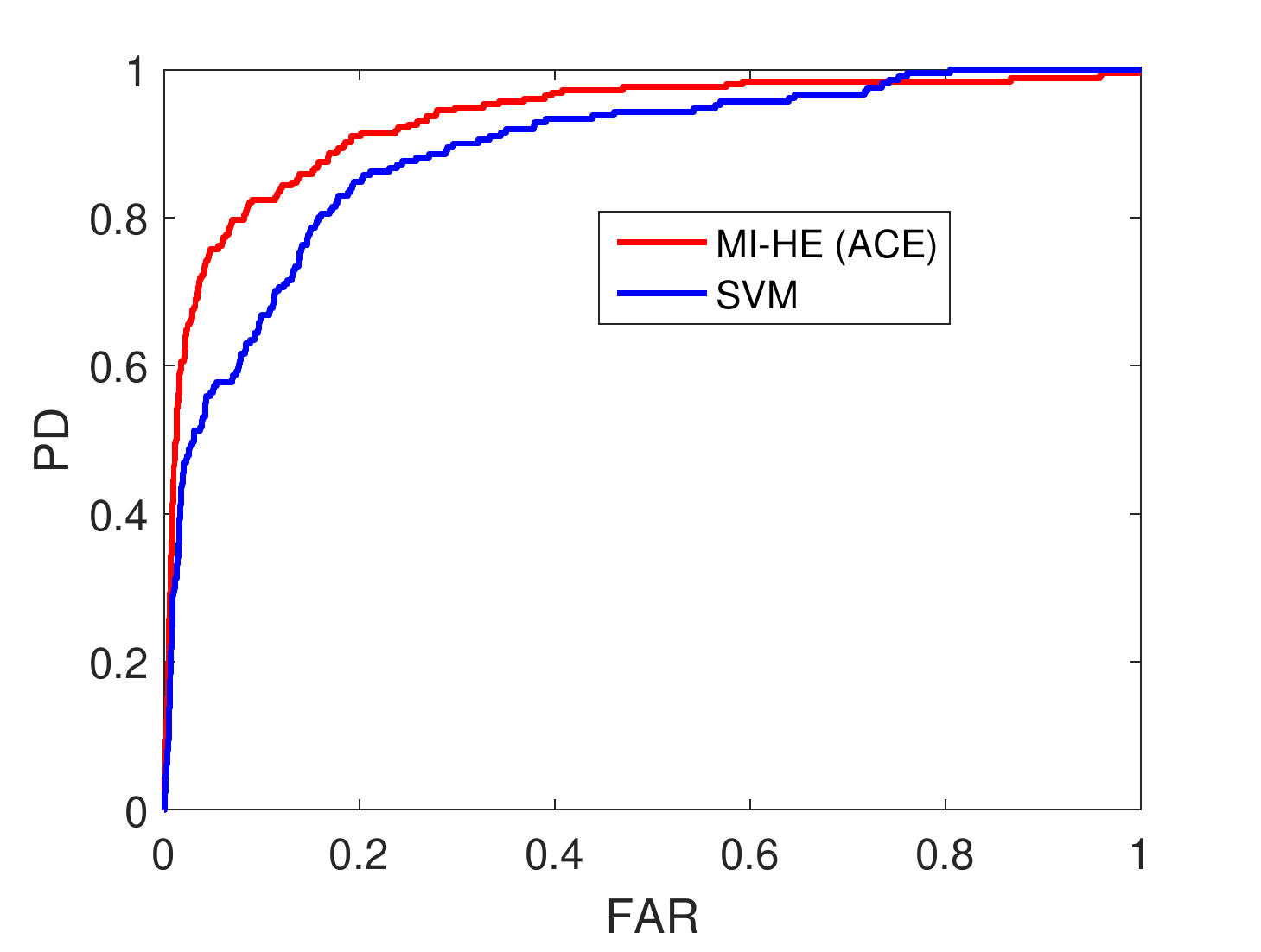} \label{fig:Circle_LIST}}
		\subfloat[NYST]{   
			\includegraphics[width=7cm]{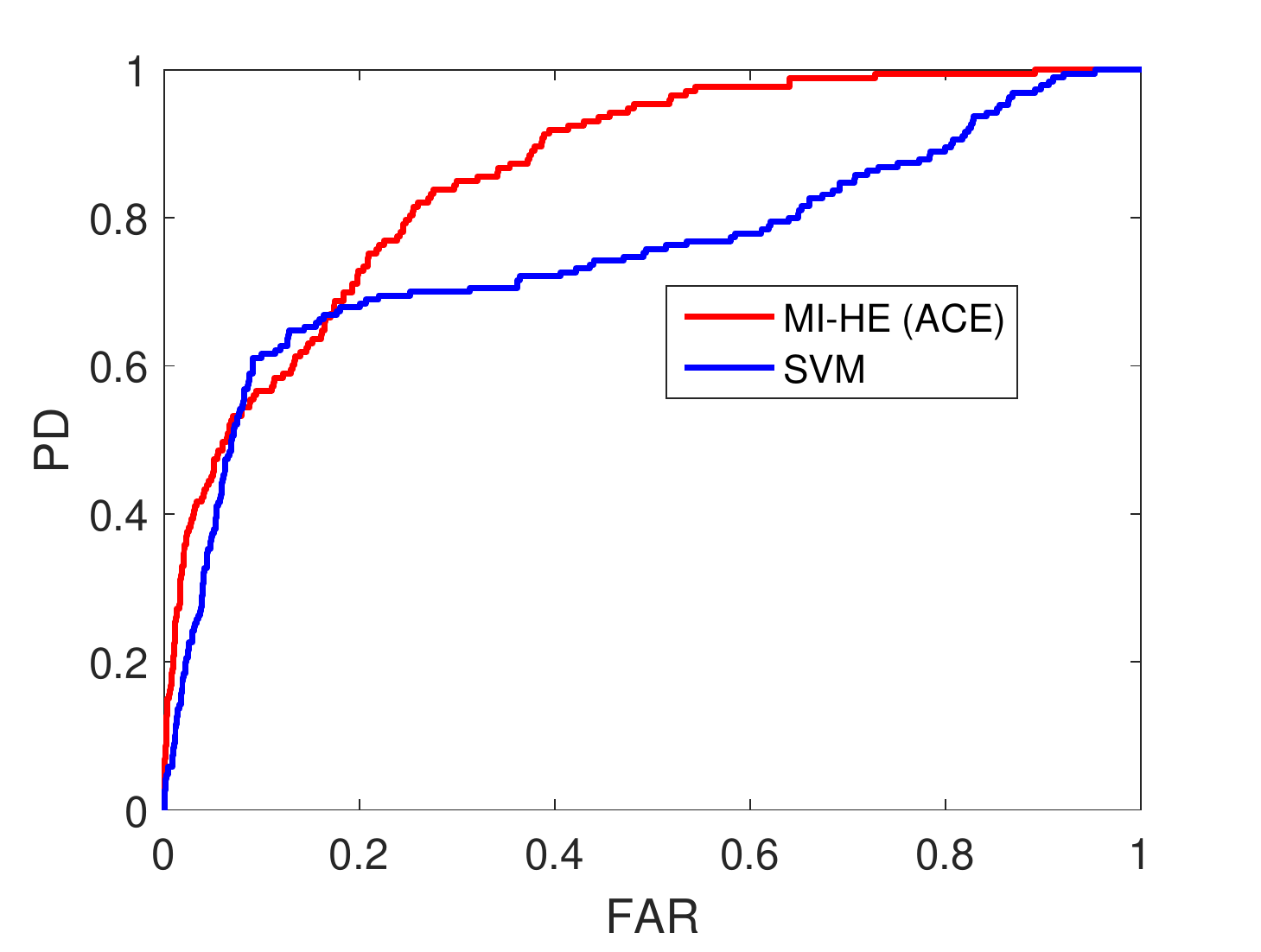} \label{fig:Circle_NYSY}}
		\caption{ROC curves of MI-HE and SVM on circle data }\label{fig:rocs_circle}
	\end{center}
\end{figure}	

{Fig. \ref{fig:rocs_poly} and \ref{fig:rocs_circle} show the ROC curves of MI-HE and SVM that train and test on polygon data and circle data, respectively. Tab. \ref{tab:NEON_tree_results} shows the detailed AUCs for each run. For training and testing on the polygon data, SVM outperforms MI-HE only on tree type QUGE and has close to but lower overall performance than MI-HE. However, for training and testing on the circle data, MI-HE outperforms SVM on each type of the classification. Furthermore, compared with the results from the polygon data, although MI-HE and SVM both provide decreased performance on the circle data, SVM suffers more from the label uncertainty given the circle data.}\\

\begin{table} [tbh!]
	\begin{center}
		\caption{Tree Species Classification Results (area under the ROC curves) by MI-HE and SVM, Best results shown in bold.}\label{tab:NEON_tree_results}
		\begin{tabular}{|c|c|c|c|c|}
			\hline
			\multirow{2}{*}{Species} 	&  \multicolumn{2}{c|}{Polygon Data} &\multicolumn{2}{c|}{Circle Data}\\
			\cline{2-5}&{MI-HE}&{SVM}&{MI-HE}&{SVM}\\
			\hline\hline
			{CAGL}   &    \textbf{0.880}          &      {0.876}                   &      \textbf{0.833}    & {0.807}\\\hline
			{PIPA}   &      \textbf{0.994}        &      {0.992}                  &      \textbf{0.984}          &    {0.969}\\\hline
			{ACRU} &      \textbf{0.873}         &         {0.829}                &     \textbf{0.867}           &0.806\\\hline
			{QUGE}  &      {0.950}               &       \textbf{0.970}           &      \textbf{0.936}        & 0.910\\\hline
			{LIST}  &      \textbf {0.942}       &       {0.884}                 &      \textbf{0.933}         & 0.890\\\hline
			{NYSY}  &     \textbf{0.887}         &       {0.862}                  &      \textbf{0.861}          & 0.755\\\hline
			{Average}  &     \textbf{0.921}      &       {0.902}                  &      \textbf{0.902}    &      0.856\\\hline
		\end{tabular}
	\end{center}
\end{table}

\section{Analysis of MI-HE Parameter Settings on Simulated Data}

In order to provide deeper insights into the sensitivity of MI-HE performance relative to variations in input parameters, we tested MI-HE on simulated hyperspectral data across a range of parameter values. Specifically, the varying parameters and ranges examined are shown in Tab. \ref{tab:test_para_range}. Red Slate was again selected as target endmember and the other three, Verde Antique, Phyllite and Pyroxenite were used as non-target endmembers as shown in Fig. \ref{fig:constituent_endmembers_extra1}. Tab. \ref{tab:toydata_test_para_bags_list} shows the bags labeling and constituent endmembers. The synthetic data has $K^+=5$ positive and $K^-=5$ negative bags with each bag containing $100$ points. If it is a positively labeled bag, there are $50$ highly-mixed target points containing mean target (Red Slate) proportion $\boldsymbol{\alpha}_{t\_mean}=0.1$ and are mixed with at least one randomly picked background endmember. Gaussian white noise was added so that the signal-to-noise ratio of the data was set to $20 dB$. The reference parameters for MI-HE were set to  $T = 1, M = 7, \rho = 0.8, b = 5, \beta=5$ and $\lambda=1\times 10^{-3}$.

\begin{table}[!htb]
	\begin{center}
		\caption{ Testing Ranges for Analysis of MI-HE Parameter Settings}  \label{tab:test_para_range}
		\begin{tabular}{|c|c|}
			\hline
			{Type} &  {Range}  \\\hline
			$M$     & $  1, 2, 3, 5, 7, 9, 11, 13, 15, 17, 19, 21  $      \\\hline
			$\beta$    & $  0.01, 0.02, 0.05, 0.1, 0.2, 0.5, 1, 2, 5, 10, 20, 50, 100  $        \\\hline
			\multirow{2}{*}{$\lambda$} & $  1\times10^{-4}, 2\times10^{-4}, 5\times10^{-4}, 1\times10^{-3}, 2\times10^{-3},$\\ &$5\times10^{-3}, 0.01, 0.02, 0.05, 0.1, 0.2, 0.5, 1   $   \\\hline
			$b$   & $ -10, -5, -2, -1, 1\times10^{-10}, 1, 2, 5, 10, 20, 50, 100 $ \\\hline
			
		\end{tabular}
	\end{center}
\end{table}

\begin{figure}
	\begin{center}
		\subfloat[M]{   
			\includegraphics[width=7cm]{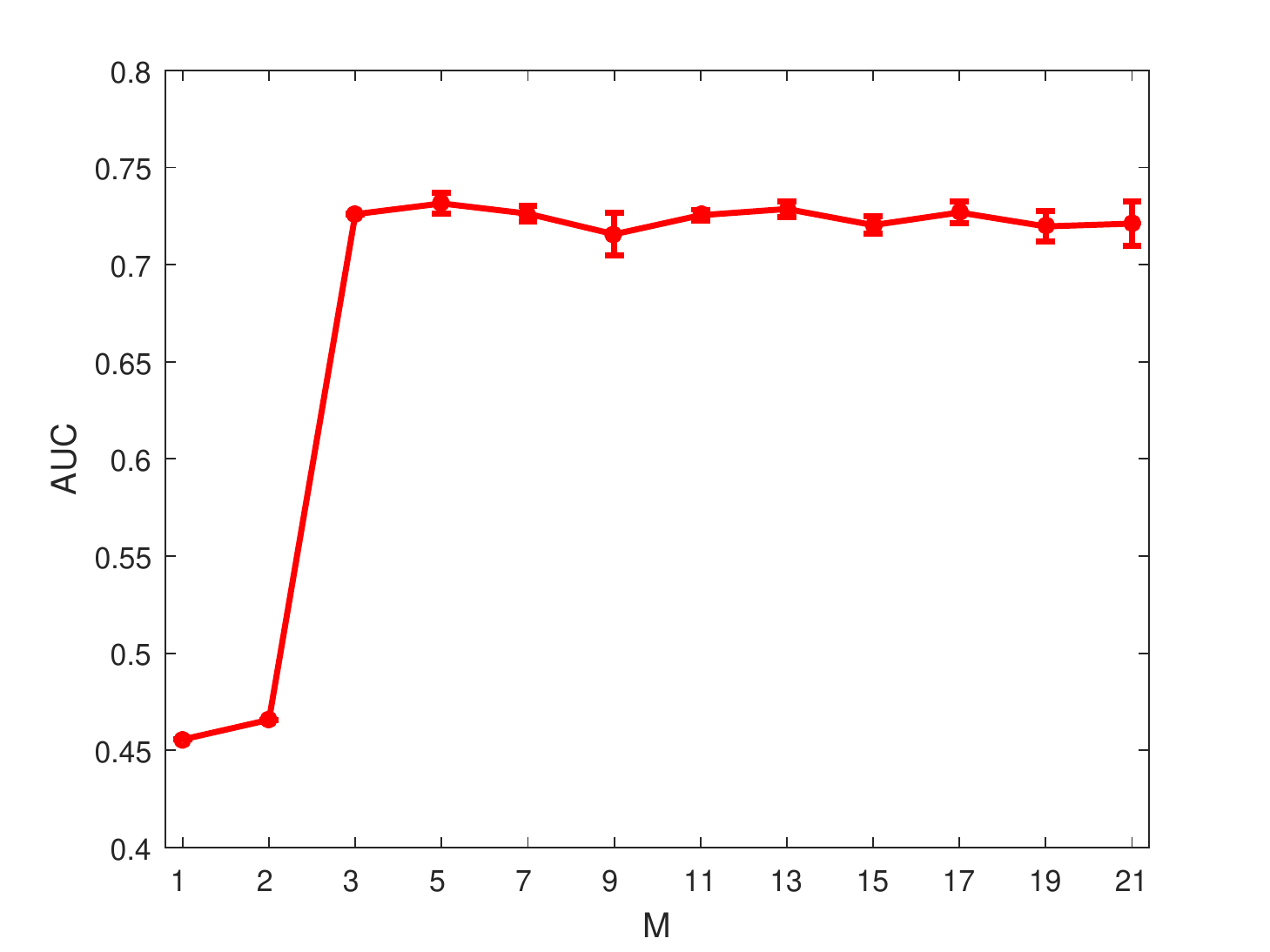} \label{fig:auc_test_M}}
		\subfloat[$\beta$]{   
			\includegraphics[width=7cm]{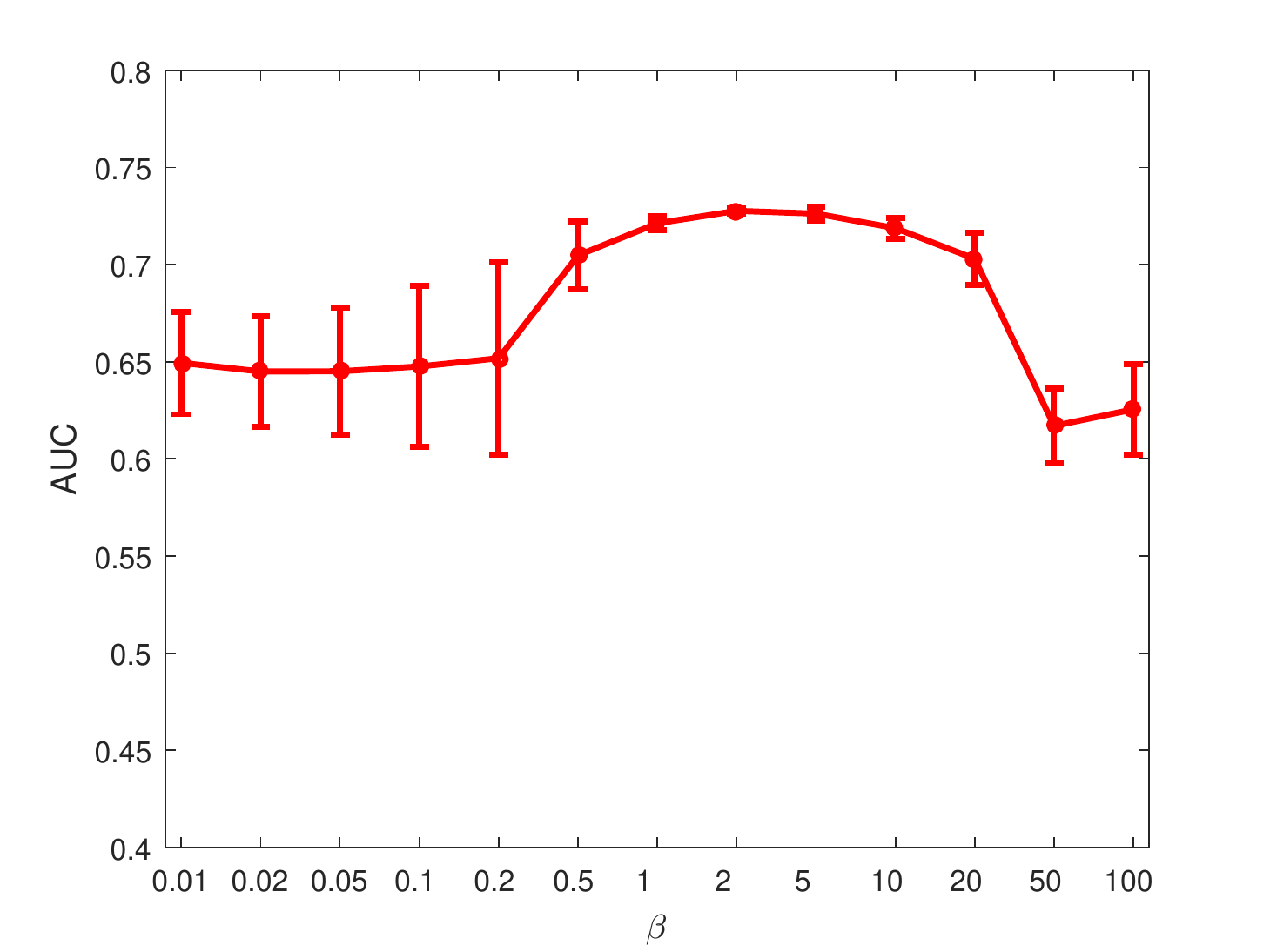} \label{fig:auc_test_beta}}\\
		\subfloat[$\lambda$]{   
			\includegraphics[width=7cm]{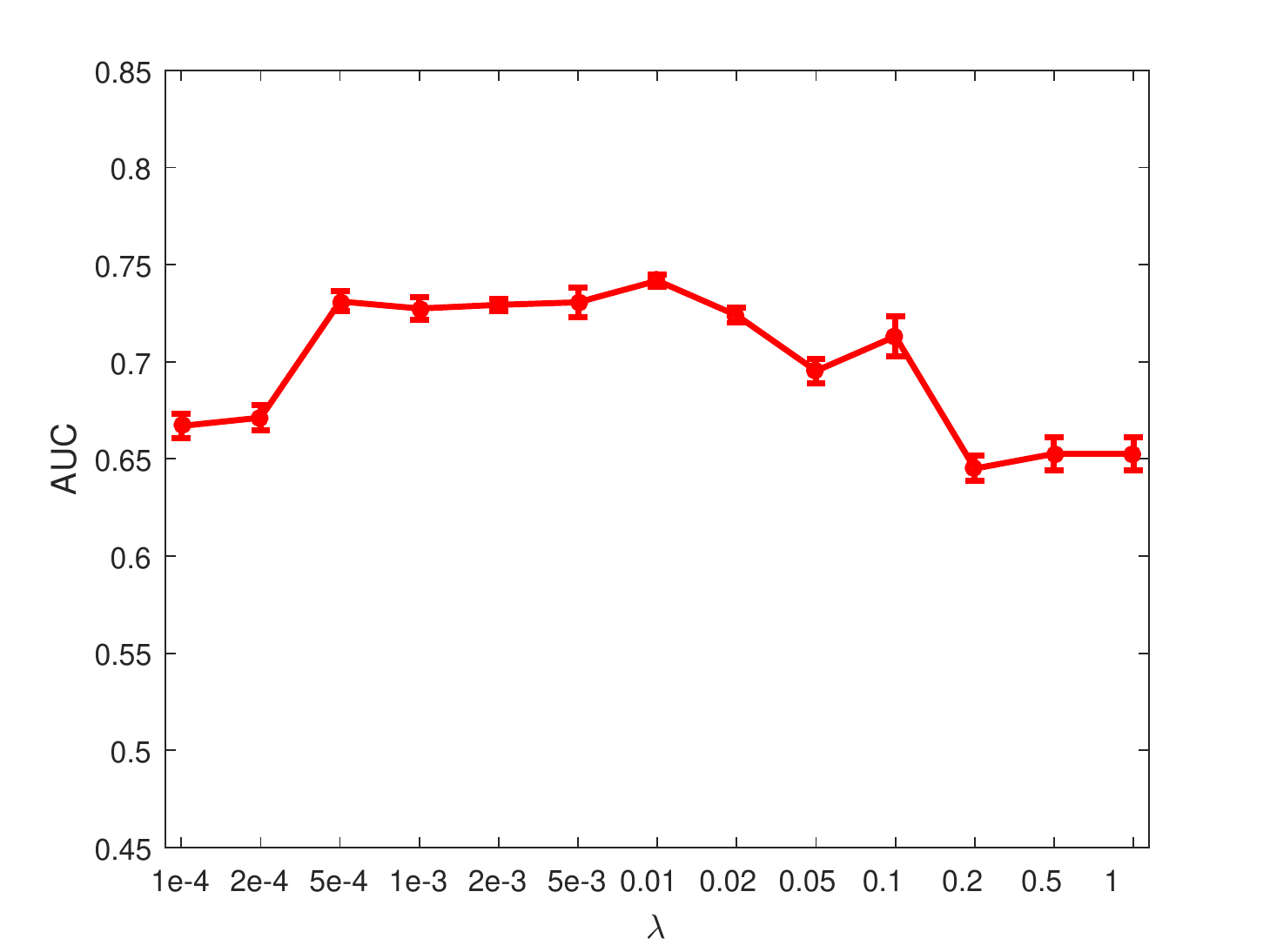} \label{fig:auc_test_lambda}}
		\subfloat[b]{   
			\includegraphics[width=7cm]{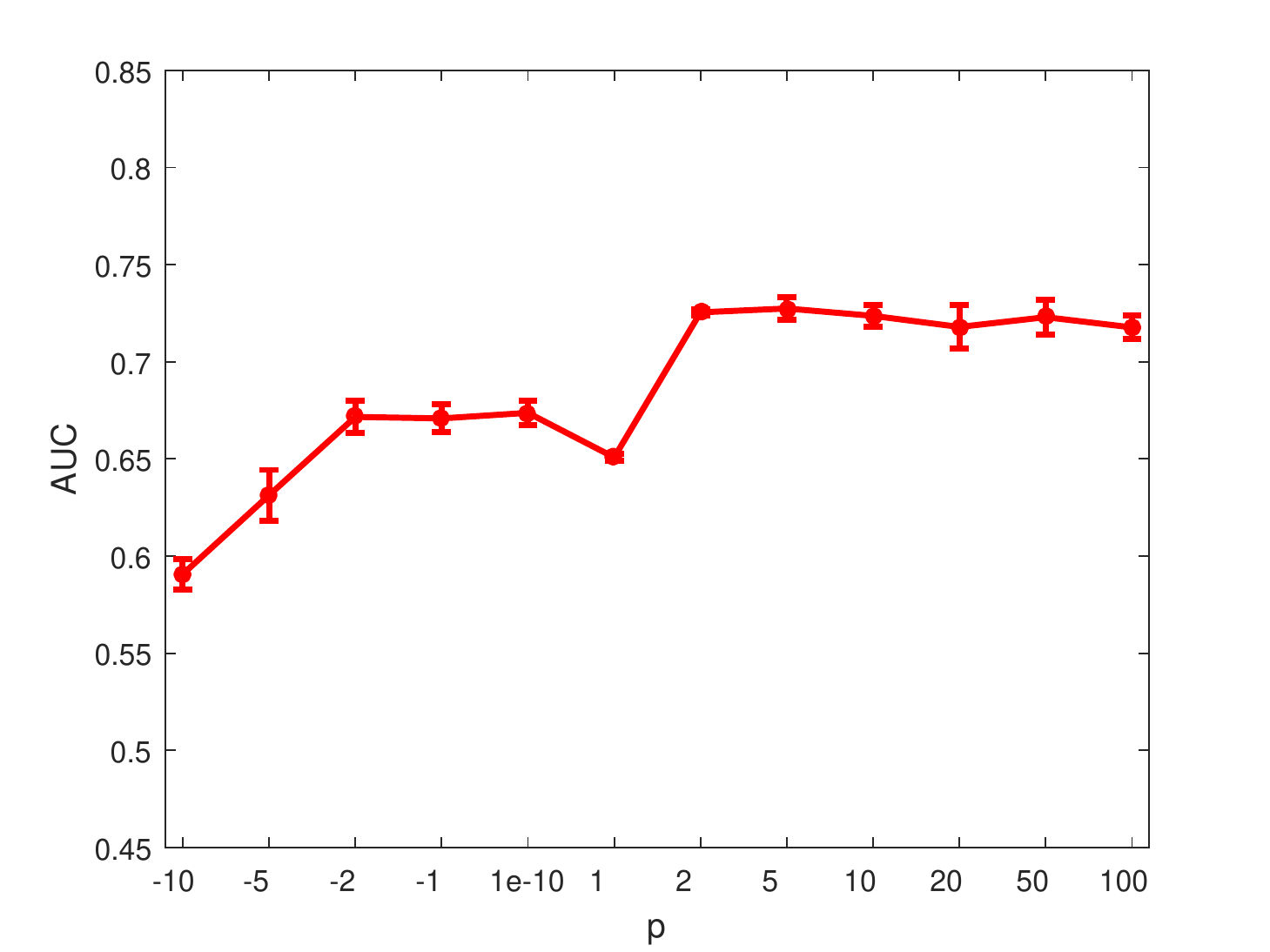} \label{fig:auc_test_p}}
		\caption{Detection statistics (AUCs) of MI-HE plots with different parameter settings}\label{fig:MIHE_AUCs_para_test}
	\end{center}
\end{figure}

Fig. \ref{fig:MIHE_AUCs_para_test} shows the detection performance (mean AUCs and variance over five runs) of MI-HE for the sensitivity analysis on the simulated data described above. Several interesting inferences can be drawn from Fig. \ref{fig:MIHE_AUCs_para_test} regarding how MI-HE responds to its parameter settings. For the setting of $M$, it can be seen from Fig. \ref{fig:auc_test_M} that MI-HE performs consistently when the number of background concepts is set to greater than or equal to 3, which is the true number of background concepts of the simulated data. Since MI-HE adopts a linear mixing model instead of convex mixing for the data mixture, the conclusion is MI-HE is not sensitive to the setting of $M$ given $M$ is not set smaller than the necessary constituent background concepts of input data.

\begin{table} 
	\begin{footnotesize}
		\begin{center}
			\caption{List of Constituent Endmembers for Synthetic Data}\label{tab:toydata_test_para_bags_list}
			\begin{tabular}{|c|c|c|}
				\hline
				Bag No. 	&  Bag Label  & Constituent Endmembers \\
				\hline\hline
				1-5 &     $+$     &    Red Slate,   Verde Antique, Phyllite, Pyroxenite    \\\hline
				6-10&          $-$        &   Verde Antique  Phyllite, Pyroxenite                       \\\hline
			\end{tabular}
		\end{center}
	\end{footnotesize}
\end{table}

For the setting of $\beta$, Fig. \ref{fig:auc_test_beta} shows MI-HE performs well with $\beta$ in the range [1, 10] on this data. Since $\beta$ is a scaling factor for the hybrid detector, $\Pr(\mathbf{x}_{ij}|\mathbf{D}, \mathbf{B}^+_i)=\exp\left(-\beta\frac{\|\mathbf{x}_{ij}-\mathbf{D}\boldsymbol{a}_{ij}^+\|^2}{\|\mathbf{x}_{ij}-\mathbf{D}^-\boldsymbol{p}_{ij}\|^2}\right)$,  the scaling effect controls the gradient of the objective function \eqref{eq:MI-HE_full_neg_log}. Since the range $\beta\in$ [1, 10] provides a moderate gradient for the exponential function, so it can be concluded that MI-HE requires $\beta$ to be set to a suitable range, \eg, [1, 10].

The setting of sparsity level, $\lambda$, results in the step length of soft shrinkage. This value should be related with the magnitude of the input data. For the simulated data tested here, the proportion values were generated from the Dirichlet distribution within range [0, 1]. As shown by Fig. \ref{fig:auc_test_lambda}, the appropriate range of $\lambda$ for this simulated data is $[5\times10^{-4}, 0.02]$ which is reasonable. However, in general, prior knowledge is needed for setting specific $\lambda$ for a real dataset. Currently we set $\lambda$ to be approximately the $1/1000$ of the $l_2$-norm mean of the training data. 

{The parameter $b$ controls the sharpness of the generalized mean model to realize the operations from minimum ($b\to-\infty$), average ($b\in(0, 1]$) to maximum ($b\to+\infty$). Since the proposed model aims to emphasize the most true positive instance from each positive bag and assumes the ``soft maximum'' operation for this generalized mean model, it is expected that the model will work well with $b$ greater than 1. The setting of this $b$ value was discussed in \cite{wu2014milcut, shrivastava2015gen}, where $b$ was set to $1.5$ \cite{wu2014milcut} and $10$ \cite{shrivastava2015gen} and observed to work well.} Fig. \ref{fig:auc_test_p} verifies this hypothesis showing that the algorithm works well for $b$ great than 1. For all experiments shown in this paper, the parameter $b$ was set to 5.

{In order to further evaluate the robustness of MI-HE to its parameters, a 4D parameters perturbation testing was conducted, in which the parameters were varied across all the possible combinations of the four parameters. The range is listed in Tab. \ref{tab:test_para_range_4D} which is a subset of Tab. \ref{tab:test_para_range}. The best and worst performance (median AUC over five runs) from this 4D loop are 0.742 and 0.695, respectively. The difference between the best and worst performance in percentage is $6.24\%$, which further validates MI-HE's stability on a large range of parameters.}

\begin{table}[!htb]
	\begin{center}
		\caption{Testing Ranges for MI-HE 4D Parameter Perturbation Testing}  \label{tab:test_para_range_4D}
		\begin{tabular}{|c|c|}
			\hline
			{Type} &  {Range}  \\\hline
			$M$     & $   3, 5, 7, 9$      \\\hline
			$\beta$    & $  1, 2, 5, 10 $        \\\hline
			$\lambda$ & $   1\times10^{-3}, 2\times10^{-3}, 5\times10^{-3}, 0.01 $   \\\hline
			$b$   & $  5, 10, 20, 50 $ \\\hline
			
		\end{tabular}
	\end{center}
\end{table}

{Although there are several parameters for MI-HE, these parameters come from the models assumed to underlie MI-HE. The sensitivity analysis of MI-HE parameters on a very noisy, highly mixed synthetic data provides a solid verification for the model stability. It can be concluded that the proposed MI-HE is generally robust to the perturbations of its parameters and there is a general range of model stability for the space of the MI-HE parameters.  Moreover, the above analysis provides a intuitive understanding and heuristic approach for setting the parameters of MI-HE to stable ranges.}

\section{Conclusion}

In this work the MI-HE target concept learning framework for MIL problems is proposed and investigated. MI-HE is able to learn multiple discriminative target concepts from ambiguously labeled data. After learning target concepts, target detection can be conducted by applying the estimated target concepts to any signature based detector. Comprehensive experiments show that the proposed MI-HE is effective in learning discriminative target concept and achieves  superior performance over comparison algorithms in several scenarios. {Although this work was inspired by sub-pixel hyperspectral target detection, the MI-HE algorithm is a general MIL concept learning algorithm that could be applied to problems with mixed and ambiguously labeled training data. }

\section*{Acknowledgements}

This material is based upon work supported by the National Science Foundation under Grant No. IIS-1350078 - CAREER: Supervised Learning for Incomplete and Uncertain Data. {The NEON crown polygon data was collected as part of National Science Foundation Dimensions of Biodiversity Grant No. 1442280. }

\numberwithin{equation}{section}
\begin{appendix}
	\section{}
\label{MIHE_optimiz}
Similar to the Dictionary Learning using Singular Value Decomposition (K-SVD) approach \cite{aharon2006ksvd}, the optimization of target and background concepts is performed by taking gradient descent with respect to one atom at a time and holding the rest fixed. Denote $f_{GM}$ as the generalized mean part of Eq. \eqref{eq:MI-HE_full_neg_log}:

\begin{footnotesize}
	\begin{eqnarray}
	f_{GM}&=&-\sum_{i=1}^{K^+}\frac{1}{b}\ln\left(\frac{1}{N_i}\sum_{j=1}^{N_i}\Lambda(\mathbf{x}_{ij},\mathbf{D}|\mathbf{B}_i)^b\right)\nonumber\\
	&=&-\sum_{i=1}^{K^+}\frac{1}{b}\ln\frac{1}{N_i}-\sum_{i=1}^{K^+}\frac{1}{b}\ln\left(\sum_{j=1}^{N_i}\Lambda(\mathbf{x}_{ij},\mathbf{D}|\mathbf{B}_i)^b\right)
	\label{eq:f_GM_ln}
	\end{eqnarray}  
\end{footnotesize}

Removing the constant part in $f_{GM}$ and taking the partial derivative with respect to $\mathbf{d}$ results in \eqref{eq:gradient_lnf_GM}:

\begin{footnotesize}
	\begin{eqnarray}
	\frac{\partial f_{GM}}{\partial\mathbf{d}}&=&-\sum_{i=1}^{K^+}\frac{1}{\sum_{j=1}^{N_i}\Lambda(\mathbf{x}_{ij},\mathbf{D}|\mathbf{B}_i)^b}\nonumber\\
	&&\cdot\left(\sum_{j=1}^{N_i}\Lambda(\mathbf{x}_{ij},\mathbf{D}|\mathbf{B}_i)^{b-1}\cdot\frac{\partial \Lambda(\mathbf{x}_{ij},\mathbf{D}|\mathbf{B}_i)}{\partial\mathbf{d}}\right),
	\label{eq:gradient_lnf_GM}
	\end{eqnarray}  
\end{footnotesize}
where $\mathbf{d}$ is a symbolic notation for any atom in $\mathbf{D}$.

Then take the partial derivative on the fidelity (second) term of the objective function Eq. \eqref{eq:MI-HE_full_neg_log} with respect to the background concept, $\mathbf{d}_k$:
\begin{small}
	\begin{eqnarray}
	&&\frac{\partial -\rho \sum_{i=K^++1}^{K}\sum_{j=1}^{N_i}\ln\Pr(l_{ij}=-|\mathbf{B}_{i})}{\partial \mathbf{d}_k}\nonumber\\
	&=&\frac{\partial\rho\sum_{i=K^++1}^{K}\sum_{j=1}^{N_i}\|\mathbf{x}_{ij}-\mathbf{D}^-\mathbf{p}_{ij}\|^2}{\partial \mathbf{d}_k}\nonumber\\
	&=&\rho\sum_{i=K^++1}^{K}\sum_{j=1}^{N_i}-2p_{ijk}(\mathbf{x}_{ij}-\mathbf{D}^-\mathbf{p}_{ij})\nonumber\\
	&=&\rho\sum_{i=K^++1}^{K}\sum_{j=1}^{N_i}-2p_{ijk}\mathbf{q}_{ij},
	\label{eq:gradient_fidelity}
	\end{eqnarray}  
\end{small}
where $p_{ijk}$ is the $k$th element in $\mathbf{p}_{ij}$ corresponding to $\mathbf{d}_k$.

The partial derivative of the cross incoherence (third) term corresponding to $\mathbf{d}_t$ is:
\begin{equation}
\frac{\partial Q(\mathcal{X}, \mathbf{D}^+, {\mathcal{A}})}{\partial \mathbf{d}_t}=\alpha\sum_{i=K^++1}^{K}\sum_{j=1}^{N_i}(\mathbf{D}^+{\mathbf{a}}_{ij}^+)^T\mathbf{x}_{ij}\cdot{a}_{ijt}^+\mathbf{x}_{ij}
\label{eq:gradient_discri}
\end{equation}

The partial derivatives of the negative objective function Eq. \eqref{eq:MI-HE_full_neg_log} with respect to $\mathbf{d}_t$ and $\mathbf{d}_k$ are shown in \eqref{eq:partial_dt} and  \eqref{eq:partial_dk}.

\begin{small}
	\begin{eqnarray}
	\frac{\partial J_3}{\partial \mathbf{d}_t}&=&-\sum_{i=1}^{K^+}\frac{1}{\sum_{j=1}^{N_i}\Lambda(\mathbf{x}_{ij},\mathbf{D}|\mathbf{B}_i)^b}\nonumber\\
	&&\cdot\left(\sum_{j=1}^{N_i}\Lambda(\mathbf{x}_{ij},\mathbf{D}|\mathbf{B}_i)^{b-1}\cdot\frac{\partial \Lambda(\mathbf{x}_{ij},\mathbf{D}|\mathbf{B}_i)}{\partial\mathbf{d}_t}\right)\nonumber\\
	&&+\alpha\sum_{i=K^++1}^{K}\sum_{j=1}^{N_i}(\mathbf{D}^+{\mathbf{a}}_{ij}^+)^T\mathbf{x}_{ij}\cdot{a}_{ijt}^+\mathbf{x}_{ij}
	\label{eq:partial_dt}
	\end{eqnarray}  
\end{small}

\begin{small}
	\begin{eqnarray}
	\frac{\partial J_3}{\partial \mathbf{d}_k}&=&-\sum_{i=1}^{K^+}\frac{1}{\sum_{j=1}^{N_i}\Lambda(\mathbf{x}_{ij},\mathbf{D}|\mathbf{B}_i)^b}\nonumber\\
	&&\cdot\left(\sum_{j=1}^{N_i}\Lambda(\mathbf{x}_{ij},\mathbf{D}|\mathbf{B}_i)^{b-1}\cdot\frac{\partial \Lambda(\mathbf{x}_{ij},\mathbf{D}|\mathbf{B}_i)}{\partial\mathbf{d}_k}\right) \nonumber\\ 
	&&+\rho\sum_{i=K^++1}^{K}\sum_{j=1}^{N_i}-2p_{ijk}\mathbf{q}_{ij}
	\label{eq:partial_dk}
	\end{eqnarray}  
\end{small}

The next step is taking the partial derivative of the hybrid detector in \eqref{eq:MI-HE_pos_inst_model} with respect to $\mathbf{d}_t$ and $\mathbf{d}_k$ shown as Eq. \eqref{eq:partial_HD_dt} and \eqref{eq:partial_HD_dk} respectively:
\begin{eqnarray} 
&&\frac{\partial\Lambda(\mathbf{x}_{ij},\mathbf{D}|\mathbf{B}_i)}{\partial\mathbf{d}_t}\nonumber\\
&=&\exp\left(-\beta\frac{\|\mathbf{x}_{ij}-\mathbf{D}\mathbf{a}_{ij}\|^2}{\|\mathbf{x}_{ij}-\mathbf{D}^-\mathbf{p}_{ij}\|^2}\right)\frac{\partial\left(-\beta\frac{\|\mathbf{x}_{ij}-\mathbf{D}\mathbf{a}_{ij}\|^2}{\|\mathbf{x}_{ij}-\mathbf{D}^-\mathbf{p}_{ij}\|^2}\right)}{\partial\mathbf{d}_t} \nonumber \\
&=&\Lambda(\mathbf{x}_{ij},\mathbf{D}|\mathbf{B}_i)\frac{2\beta a_{ijt}^+(\mathbf{x}_{ij}-\mathbf{D}\mathbf{a}_{ij})}{\|\mathbf{x}_{ij}-\mathbf{D}^-\mathbf{p}_{ij}\|^2}\nonumber \\
&=&\Lambda(\mathbf{x}_{ij},\mathbf{D}|\mathbf{B}_i)\frac{2\beta a_{ijt}^+\mathbf{r}_{ij}}{\|\mathbf{q}_{ij}\|^2}
\label{eq:partial_HD_dt}
\end{eqnarray}  
\begin{footnotesize}
	\begin{eqnarray} 
	&&\frac{\partial\Lambda(\mathbf{x}_{ij},\mathbf{D}|\mathbf{B}_i)}{\partial\mathbf{d}_k}\nonumber\\
	&=&\exp\left(-\beta\frac{\|\mathbf{x}_{ij}-\mathbf{D}\mathbf{a}_{ij}\|^2}{\|\mathbf{x}_{ij}-\mathbf{D}^-\mathbf{p}_{ij}\|^2}\right)\frac{\partial\left(-\beta\frac{\|\mathbf{x}_{ij}-\mathbf{D}\mathbf{a}_{ij}\|^2}{\|\mathbf{x}_{ij}-\mathbf{D}^-\mathbf{p}_{ij}\|^2}\right)}{\mathbf{d}_k} \nonumber \\
	&=&\Lambda(\mathbf{x}_{ij},\mathbf{D}|\mathbf{B}_i)\frac{2\beta a_{ijk}^-\mathbf{r}_{ij}\|\mathbf{q}_{ij}\|^2-2\beta p_{ijk}\|\mathbf{r}_{ij}\|^2\mathbf{q}_{ij}}{\|\mathbf{q}_{ij}\|^4}
	\label{eq:partial_HD_dk}
	\end{eqnarray}  
\end{footnotesize}

Substituting the gradient of hybrid detector with respect to $\mathbf{d}_t$ and $\mathbf{d}_k$ in Eq. \eqref{eq:partial_HD_dt} and \eqref{eq:partial_HD_dk} to Eq. \eqref{eq:partial_dt} and  \eqref{eq:partial_dk}, respectively, we can get the resultant gradient of the objective function \eqref{eq:MI-HE_full_neg_log} over $\mathbf{d}_t$ and $\mathbf{d}_k$:

\begin{eqnarray}
\triangle\mathbf{d}_t&=&-\sum_{i=1}^{K^+}\frac{1}{\sum_{j=1}^{N_i}\Lambda(\mathbf{x}_{ij},\mathbf{D}|\mathbf{B}_i)^b}\nonumber\\
&&\left(\sum_{j=1}^{N_i}\Lambda(\mathbf{x}_{ij},\mathbf{D}|\mathbf{B}_i)^{b}\cdot\frac{2\beta a_{ijt}^+\mathbf{r}_{ij}}{\|\mathbf{q}_{ij}\|^2}\right)\nonumber\\
&&+\alpha\sum_{i=K^++1}^{K}\sum_{j=1}^{N_i}(\mathbf{D}^+{\mathbf{a}}_{ij}^+)^T\mathbf{x}_{ij}\cdot{a}_{ijt}^+\mathbf{x}_{ij}
\label{eq:gradient_dt}
\end{eqnarray}

\begin{scriptsize}
	\begin{eqnarray}
	\triangle\mathbf{d}_k&=&-\sum_{i=1}^{K^+}\frac{1}{\sum_{j=1}^{N_i}\Lambda(\mathbf{x}_{ij},\mathbf{D}|\mathbf{B}_i)^b}\nonumber\\
	&&\left(\sum_{j=1}^{N_i}\Lambda(\mathbf{x}_{ij},\mathbf{D}|\mathbf{B}_i)^{b}\cdot2\beta\frac{a_{ijk}^-\mathbf{r}_{ij}\|\mathbf{q}_{ij}\|^2-p_{ijk}\|\mathbf{r}_{ij}\|^2\mathbf{q}_{ij}}{\|\mathbf{q}_{ij}\|^4}\right)\nonumber\\
	&&-\rho\sum_{i=K^++1}^{K}\sum_{j=1}^{N_i}2p_{ijk}\mathbf{q}_{ij}
	\label{eq:gradient_dk}
	\end{eqnarray}  
\end{scriptsize}

The optimization of sparse representation can be viewed as a $l_1$ regularized least squares problem, also known as lasso problem \cite{tibshirani1996regression, chen2001atomic}, denoted as $\mathcal{L}$. The lasso problem is shown in Eq. \eqref{eq:MI-HE_lasso}, where given concept (or dictionary) set $\mathbf{D}$ and preset {regularization constant} $\lambda$, $\mathbf{a}^*$ is the optimal sparse representation of the input data $\mathbf{x}$. Here we adopt the iterative shrinkage-thresholding algorithm (ISTA) \cite{figueiredo2003algorithm, daubechies2003iterative} for solving for the sparse codes  $\mathbf{a}$.

The gradient of \eqref{eq:MI-HE_lasso} with respect to $\mathbf{a}$ without considering the $l_1$ penalty term is: 
\begin{equation}
\frac{\partial \mathcal{L}}{\partial \mathbf{a}}=-\mathbf{D}^T\left(\mathbf{x} - \mathbf{D}\mathbf{a}\right).
\label{eq:gradient_Lasso}
\end{equation}

Then $\mathbf{a}$ at $q^{th}$ iteration can be updated using gradient descent shown in \eqref{eq:alpha_update}:
\begin{equation}
\mathbf{a}^{q}=\mathbf{a}^{q-1}-\delta	\frac{\partial \mathcal{L}}{\partial \mathbf{a}},
\label{eq:alpha_update}
\end{equation}
followed by a soft-thresholding step:
\begin{equation}
\mathbf{a}^{*}=S_{\lambda}\left(\mathbf{a}^{q}\right),
\label{eq:sf_alpha} 
\end{equation}
where $S_{\lambda}\text{: } \mathbb{R}^n\rightarrow\mathbb{R}^n$ is the \textit{soft-thresholding} operator defined by 
\begin{equation}
S_{\lambda}\left(\mathbf{a}[k]\right)=sign(\mathbf{a}[k])\max(|\mathbf{a}[k]|-\lambda,0), k=1,\cdots,n
\end{equation}

Following a similar proof to that in \cite{facchinei2007finite}, when the step length $\delta$ satisfies \eqref{eqn:MI-HE_eta}, the update of $\mathbf{a}$ using a gradient descent method with step length $\eta$ monotonically decreases the value of the objective function, where ${Eig}_{max}(\mathbf{\mathbf{D}^T\mathbf{D}})$ denotes the maximum eigenvalue of $\mathbf{D}^T\mathbf{D}$. For simplicity,  $\delta$ was set to ${Eig}_{max}\left( \mathbf{D}^T\mathbf{D}\right)$ for all input data: 

\begin{equation}
\delta\in\left(0, \frac{1}{{Eig}_{max}\left( \mathbf{D}^T\mathbf{D}\right)}\right)
\label{eqn:MI-HE_eta}
\end{equation}

Finally the resultant update equation for the sparse representation of instance $\mathbf{x}$ given concept set $\mathbf{D}$ is:
\begin{equation}
\mathbf{a}^*=S_{\lambda}\left(\mathbf{a}^{q}+\frac{1}{{Eig}_{max}\left(\mathbf{D}^{-T}\mathbf{D}\right)}\left(\mathbf{D}^{T}(\mathbf{x}-\mathbf{D}\mathbf{a}^{q})\right)\right)
\label{eq:alpha_update_final}
\end{equation}
\end{appendix}

{  \bibliographystyle{IEEEtran}
\bibliography{MIHE_bib}}

\begin{thebibliography}{10}
\providecommand{\url}[1]{#1}
\csname url@samestyle\endcsname
\providecommand{\newblock}{\relax}
\providecommand{\bibinfo}[2]{#2}
\providecommand{\BIBentrySTDinterwordspacing}{\spaceskip=0pt\relax}
\providecommand{\BIBentryALTinterwordstretchfactor}{4}
\providecommand{\BIBentryALTinterwordspacing}{\spaceskip=\fontdimen2\font plus
\BIBentryALTinterwordstretchfactor\fontdimen3\font minus
  \fontdimen4\font\relax}
\providecommand{\BIBforeignlanguage}[2]{{%
\expandafter\ifx\csname l@#1\endcsname\relax
\typeout{** WARNING: IEEEtran.bst: No hyphenation pattern has been}%
\typeout{** loaded for the language `#1'. Using the pattern for}%
\typeout{** the default language instead.}%
\else
\language=\csname l@#1\endcsname
\fi
#2}}
\providecommand{\BIBdecl}{\relax}
\BIBdecl

\bibitem{landgrebe2002hyperspectral}
D.~Landgrebe, ``Hyperspectral image data analysis,'' \emph{IEEE Signal Process.
  Mag.}, vol.~19, no.~1, pp. 17--28, 2002.

\bibitem{keshava2002spectral}
N.~Keshava and J.~F. Mustard, ``Spectral unmixing,'' \emph{IEEE Signal Process.
  Mag.}, vol.~19, no.~1, pp. 44--57, 2002.

\bibitem{bioucas2012hyperspectral}
J.~M. Bioucas-Dias, A.~Plaza, N.~Dobigeon \emph{et~al.}, ``Hyperspectral
  unmixing overview: Geometrical, statistical, and sparse regression-based
  approaches,'' \emph{IEEE J. Sel. Topics Applied Earth Observations Remote
  Sens.}, vol.~5, no.~2, pp. 354--379, 2012.

\bibitem{yuksel2015multiple}
S.~E. Yuksel, J.~Bolton, and P.~Gader, ``Multiple-instance hidden markov models
  with applications to landmine detection,'' \emph{IEEE Trans. Geosci. Remote
  Sens.}, vol.~53, no.~12, pp. 6766--6775, 2015.

\bibitem{4389068}
A.~Zare, J.~Bolton, P.~Gader, and M.~Schatten, ``Vegetation mapping for
  landmine detection using long-wave hyperspectral imagery,'' \emph{IEEE Trans.
  Geosci. Remote Sens.}, vol.~46, no.~1, pp. 172--178, Jan 2008.

\bibitem{mahajan2014using}
G.~Mahajan, R.~Sahoo, R.~Pandey, V.~Gupta, and D.~Kumar, ``Using hyperspectral
  remote sensing techniques to monitor nitrogen, phosphorus, sulphur and
  potassium in wheat (triticum aestivum l.),'' \emph{Precision Agriculture},
  vol.~15, no.~5, pp. 499--522, 2014.

\bibitem{wang2012mixture}
Z.~Wang, L.~Lan, and S.~Vucetic, ``Mixture model for multiple instance
  regression and applications in remote sensing,'' \emph{IIEEE Trans. Geosci.
  Remote Sens.}, vol.~50, no.~6, pp. 2226--2237, 2012.

\bibitem{pike2016minimum}
R.~Pike, G.~Lu, D.~Wang \emph{et~al.}, ``A minimum spanning forest-based method
  for noninvasive cancer detection with hyperspectral imaging,'' \emph{IEEE
  Trans. Biomed. Eng.}, vol.~63, no.~3, pp. 653--663, 2016.

\bibitem{pardo2017directional}
A.~Pardo, E.~Real, V.~Krishnaswamy \emph{et~al.}, ``Directional kernel density
  estimation for classification of breast tissue spectra,'' \emph{IEEE Trans.
  Med. Imag.}, vol.~36, no.~1, pp. 64--73, 2017.

\bibitem{eismann2009automated}
M.~T. Eismann, A.~D. Stocker, and N.~M. Nasrabadi, ``Automated hyperspectral
  cueing for civilian search and rescue,'' \emph{Proc. IEEE}, vol.~97, no.~6,
  pp. 1031--1055, 2009.

\bibitem{lara2013monitoring}
M.~Lara, L.~Lle{\'o}, B.~Diezma-Iglesias, J.-M. Roger, and M.~Ruiz-Altisent,
  ``Monitoring spinach shelf-life with hyperspectral image through packaging
  films,'' \emph{J. Food Eng.}, vol. 119, no.~2, pp. 353--361, 2013.

\bibitem{manolakis2002detection}
D.~Manolakis and G.~Shaw, ``Detection algorithms for hyperspectral imaging
  applications,'' \emph{IEEE Signal Process. Mag.}, vol.~19, no.~1, pp. 29--43,
  2002.

\bibitem{nasrabadi2014hyperspectral}
N.~M. Nasrabadi, ``Hyperspectral target detection: An overview of current and
  future challenges,'' \emph{IEEE Signal Process. Mag.}, vol.~31, no.~1, pp.
  34--44, 2014.

\bibitem{gader:2013}
P.~Gader, A.~Zare \emph{et~al.}, ``\uppercase{MUUFL} gulfport hyperspectral and
  lidar airborne data set,'' University of Florida, Gainesville, FL,
  REP-2013-570, Tech. Rep., Oct. 2013.

\bibitem{Dietterich:1997}
T.~G. Dietterich, R.~H. Lathrop, and T.~Lozano-P{\'e}rez, ``Solving the
  multiple instance problem with axis-parallel rectangles,'' \emph{Artificial
  Intell.}, vol.~89, no.~1, pp. 31--71, 1997.

\bibitem{Maron:1998}
O.~Maron and T.~Lozano-Perez, ``A framework for multiple-instance learning.''
  in \emph{Advances Neural Inf. Process. Syst. (NIPS)}, vol.~10, 1998, pp.
  570--576.

\bibitem{Zare:2015fumi}
C.~Jiao and A.~Zare, ``Functions of multiple instances for learning target
  signatures,'' \emph{IEEE Trans. Geosci. Remote Sens.}, vol.~53, no.~8, pp.
  4670 -- 4686, 2015.

\bibitem{jiao2016ICPR}
------, ``Multiple instance dictionary learning using functions of multiple
  instances,'' in \emph{Int. Conf. Pattern Recognition (ICPR)}, 2016, pp.
  2688--2693.

\bibitem{zare2016miace}
A.~Zare and C.~Jiao, ``Discriminative multiple instance hyperspectral target
  characterization,'' \emph{IEEE Trans. Pattern Anal. Mach. Intell.}, In Press.

\bibitem{jiao2017MIHE}
C.~Jiao and A.~Zare, ``Multiple instance hybrid estimator for learning target
  signatures,'' in \emph{Proc. IEEE Intl. Geosci. Remote Sens. Symp. (IGARSS)},
  2017, pp. 1--4.

\bibitem{andrews2002support}
S.~Andrews, I.~Tsochantaridis, and T.~Hofmann, ``Support vector machines for
  multiple-instance learning,'' in \emph{Advances Neural Inf. Process. Syst.
  (NIPS)}, 2002, pp. 561--568.

\bibitem{chen2006miles}
Y.~Chen, J.~Bi, and J.~Z. Wang, ``\protect{MILES}: Multiple-instance learning
  via embedded instance selection,'' \emph{IEEE Trans. Pattern Anal. Mach.
  Intell.}, vol.~28, no.~12, pp. 1931--1947, 2006.

\bibitem{zhu20041}
J.~Zhu, S.~Rosset, T.~Hastie, and R.~Tibshirani, ``1-norm support vector
  machines,'' in \emph{Advances Neural Inf. Process. Syst. (NIPS)}, vol.~16,
  no.~1, 2004, pp. 49--56.

\bibitem{Zhang:2002}
Q.~Zhang and S.~A. Goldman, ``{EM-DD: An improved multiple-instance learning
  technique},'' in \emph{Advances Neural Inf. Process. Syst. (NIPS)}, vol.~2,
  2002, pp. 1073--1080.

\bibitem{Zare:2014whispers}
A.~Zare and C.~Jiao, ``Extended functions of multiple instances for target
  characterization,'' in \emph{IEEE Workshop Hyperspectral Image Signal
  Process.: Evolution in Remote Sens. (WHISPERS)}, 2014, pp. 1--4.

\bibitem{Theiler:2006}
J.~Theiler and B.~R. Foy, ``Effect of signal contamination in matched-filter
  detection of the signal on a cluttered background,'' \emph{IEEE Geosci.
  Remote Sens. Lett}, vol.~3, no.~1, pp. 98--102, Jan 2006.

\bibitem{Nasrabadi:2008}
N.~M. Nasrabadi, ``Regularized spectral matched filter for target recognition
  in hyperspectral imagery,'' \emph{IEEE Signal Process. Lett}, vol.~15, pp.
  317--320, 2008.

\bibitem{Kraut:1999}
S.~Kraut and L.~Scharf, ``The {CFAR} adaptive subspace detector is a
  scale-invariant {GLRT},'' \emph{IEEE Trans. Signal Process.}, vol.~47, no.~9,
  pp. 2538 --2541, Sept. 1999.

\bibitem{kraut:2001}
S.~Kraut, L.~Scharf, and L.~McWhorter, ``Adaptive subspace detectors,''
  \emph{IEEE Trans. Signal Process.}, vol.~49, no.~1, pp. 1--16, 2001.

\bibitem{maron1998multiple}
O.~Maron and A.~L. Ratan, ``Multiple-instance learning for natural scene
  classification.'' in \emph{Int. Conf. Mach. Learning (ICML)}, vol.~98, 1998,
  pp. 341--349.

\bibitem{babenko2008simultaneous}
B.~Babenko, P.~Doll{\'a}r, Z.~Tu, and S.~Belongie, ``Simultaneous learning and
  alignment: Multi-instance and multi-pose learning,'' in \emph{Workshop on
  Faces in `Real-Life' Images: Detection, Alignment, and Recognition}, 2008.

\bibitem{bullen1988means}
P.~Bullen, D.~Mitrinovi{\'c}, and P.~Vasi{\'c}, ``Means and their inequalities,
  mathematics and its applications,'' 1988.

\bibitem{xu2012multiple}
Y.~Xu, J.-Y. Zhu, E.~Chang, and Z.~Tu, ``Multiple clustered instance learning
  for histopathology cancer image classification, segmentation and
  clustering,'' in \emph{Computer Vision and Pattern Recognition (CVPR), 2012
  IEEE Conference on}.\hskip 1em plus 0.5em minus 0.4em\relax IEEE, 2012, pp.
  964--971.

\bibitem{quellec2016multiple}
G.~Quellec, M.~Lamard, M.~Cozic, G.~Coatrieux, and G.~Cazuguel,
  ``Multiple-instance learning for anomaly detection in digital mammography,''
  \emph{IEEE Transactions on Medical Imaging}, vol.~35, no.~7, pp. 1604--1614,
  2016.

\bibitem{wu2014milcut}
J.~Wu, Y.~Zhao, J.-Y. Zhu, S.~Luo, and Z.~Tu, ``Milcut: A sweeping line
  multiple instance learning paradigm for interactive image segmentation,'' in
  \emph{Computer Vision and Pattern Recognition (CVPR), 2014 IEEE Conference
  on}, 2014, pp. 256--263.

\bibitem{kraus2016classifying}
O.~Z. Kraus, J.~L. Ba, and B.~J. Frey, ``Classifying and segmenting microscopy
  images with deep multiple instance learning,'' \emph{Bioinformatics},
  vol.~32, no.~12, pp. i52--i59, 2016.

\bibitem{wohlhart2011multiple}
P.~Wohlhart, M.~K{\"o}stinger, P.~M. Roth, and H.~Bischof, ``Multiple instance
  boosting for face recognition in videos,'' in \emph{Joint Pattern Recognition
  Symposium}.\hskip 1em plus 0.5em minus 0.4em\relax Springer, 2011, pp.
  132--141.

\bibitem{tibshirani1996regression}
R.~Tibshirani, ``Regression shrinkage and selection via the lasso,'' \emph{J.
  Royal Statistical Soc. Series B (Methodological)}, pp. 267--288, 1996.

\bibitem{chen2001atomic}
S.~S. Chen, D.~L. Donoho, and M.~A. Saunders, ``Atomic decomposition by basis
  pursuit,'' \emph{SIAM review}, vol.~43, no.~1, pp. 129--159, 2001.

\bibitem{mallat1999wavelet}
S.~Mallat, \emph{A wavelet tour of signal processing, Third Edition: The Sparse
  Way}.\hskip 1em plus 0.5em minus 0.4em\relax Academic Press, 2008.

\bibitem{bach2012optimization}
F.~Bach, R.~Jenatton, J.~Mairal, and G.~Obozinski, ``Optimization with
  sparsity-inducing penalties,'' \emph{Found. Trends Mach. Learning}, vol.~4,
  no.~1, pp. 1--106, 2012.

\bibitem{mairal2014sparse}
J.~Mairal, F.~Bach, and J.~Ponce, ``Sparse modeling for image and vision
  processing,'' \emph{Found. Trends Comput. Graphics Vision}, vol.~8, no. 2-3,
  pp. 85--283, 2014.

\bibitem{figueiredo2003algorithm}
M.~A. Figueiredo and R.~D. Nowak, ``An \protect{EM} algorithm for wavelet-based
  image restoration,'' \emph{IEEE Trans. Image Process.}, vol.~12, no.~8, pp.
  906--916, 2003.

\bibitem{daubechies2003iterative}
I.~Daubechies, M.~Defrise, and C.~De~Mol, ``An iterative thresholding algorithm
  for linear inverse problems with a sparsity constraint,'' \emph{Commun. Pure
  Applied Math.}, vol.~57, pp. 1413--1457, 2004.

\bibitem{broadwater2004hybrid}
J.~Broadwater, R.~Meth, and R.~Chellappa, ``A hybrid algorithm for subpixel
  detection in hyperspectral imagery,'' in \emph{Proc. IEEE Intl. Geosci.
  Remote Sens. Symp. (IGARSS)}, vol.~3, 2004, pp. 1601--1604.

\bibitem{Broadwater:2007}
J.~Broadwater and R.~Chellappa, ``Hybrid detectors for subpixel targets,''
  \emph{IEEE Trans. Pattern Anal. Mach. Intell.}, vol.~29, no.~11, pp.
  1891--1903, Nov. 2007.

\bibitem{ramirez2010classification}
I.~Ramirez, P.~Sprechmann, and G.~Sapiro, ``Classification and clustering via
  dictionary learning with structured incoherence and shared features,'' in
  \emph{IEEE Conf. Comput. Vision Pattern Recognition (CVPR)}, 2010, pp.
  3501--3508.

\bibitem{yang2011fisher}
M.~Yang, L.~Zhang, X.~Feng, and D.~Zhang, ``Fisher discrimination dictionary
  learning for sparse representation,'' in \emph{Int. Conf. Comput.
  Vision}.\hskip 1em plus 0.5em minus 0.4em\relax IEEE, 2011, pp. 543--550.

\bibitem{yang2014sparse}
------, ``Sparse representation based fisher discrimination dictionary learning
  for image classification,'' \emph{Int. J. Comput. Vision}, vol. 109, no.~3,
  pp. 209--232, 2014.

\bibitem{zare_MIHE_code:2018}
C.~Jiao and A.~Zare, ``\protect{GatorSense/MIHE}: Initial release (version
  0.1), \protect{Zenodo},'' July 2018,
  \url{http://doi.org/10.5281/zenodo.1320109}.

\bibitem{nascimento:2005}
J.~M. Nascimento and J.~M. Dias, ``Vertex component analysis: A fast algorithm
  to unmix hyperspectral data,'' \emph{IEEE Transactions on Geoscience and
  Remote Sensing}, vol.~43, no.~4, pp. 898--910, 2005.

\bibitem{aster:2009}
A.~Baldridge, S.~Hook, C.~Grove, and G.~Rivera, ``The \uppercase{ASTER}
  spectral library version 2.0,'' \emph{Remote Sensing of Environment}, vol.
  113, no.~4, pp. 711--715, 2009.

\bibitem{shrivastava2015gen}
A.~Shrivastava, V.~M. Patel, j.~K. Pillai, and R.~Chellappa, ``Generalized
  dictionaries for multiple instance learning,'' \emph{Int. J. of Comput.
  Vision}, vol. 114, no.~2, pp. 288--305, Septmber 2015.

\bibitem{shrivastava2014dictionary}
A.~Shrivastava, J.~K. Pillai, V.~M. Patel, and R.~Chellappa, ``Dictionary-based
  multiple instance learning,'' in \emph{IEEE Int. Conf. Image Process.}, 2014,
  pp. 160--164.

\bibitem{chang2011libsvm}
C.-C. Chang and C.-J. Lin, ``\protect{LIBSVM}: a library for support vector
  machines,'' \emph{ACM Trans. Intell. Syst. Technology}, vol.~2, no.~3, pp.
  1--27, 2011.

\bibitem{glenn_gulfport:2013}
T.~Glenn, A.~Zare, P.~Gader, and D.~Dranishnikov, ``Bullwinkle: Scoring code
  for sub-pixel targets (version 1.0) [software],'' 2013,
  \url{http://engineers.missouri.edu/zarea/code/}.

\bibitem{NEON_data}
{National Ecological Observatory Network}, 2016, \protect{Data} accessed on
  Jan., 2016. Available on-line \url{http://data.neonscience.org/} from
  Battelle, Boulder, CO, USA.

\bibitem{graves2016tree}
S.~J. Graves, G.~P. Asner, R.~E. Martin, C.~B. Anderson, M.~S. Colgan,
  L.~Kalantari, and S.~A. Bohlman, ``Tree species abundance predictions in a
  tropical agricultural landscape with a supervised classification model and
  imbalanced data,'' \emph{Remote Sensing}, vol.~8, no.~2, p. 161, 2016.

\bibitem{nia2015impact}
M.~S. Nia, D.~Z. Wang, S.~A. Bohlman, P.~Gader, S.~J. Graves, and M.~Petrovic,
  ``Impact of atmospheric correction and image filtering on hyperspectral
  classification of tree species using support vector machine,'' \emph{Journal
  of Applied Remote Sensing}, vol.~9, no.~1, pp. 095\,990--095\,990, 2015.

\bibitem{zare_gulfport:2018}
A.~Zare, T.~Glenn, and P.~Gader, ``Gatorsense/hsi\_toolkit (version v0.1),
  \protect{Zenodo},'' may 2018, \url{http://doi.org/10.5281/zenodo.1186417}.

\bibitem{aharon2006ksvd}
M.~Aharon, M.~Elad, and A.~Bruckstein, ``K-\protect{SVD}: An algorithm for
  designing overcomplete dictionaries for sparse representation,'' \emph{IEEE
  Trans. on Signal Process.}, vol.~54, no.~11, pp. 4311--4322, 2006.

\bibitem{facchinei2007finite}
F.~Facchinei and J.-S. Pang, \emph{Finite-dimensional variational inequalities
  and complementarity problems}.\hskip 1em plus 0.5em minus 0.4em\relax
  Springer Science \& Business Media, 2007.

\end{thebibliography}

\end{document}